%% file: main.tex
\documentclass[acmsmall]{acmart}
\AtBeginDocument{%
  \providecommand\BibTeX{{%
    \normalfont B\kern-0.5em{\scshape i\kern-0.25em b}\kern-0.8em\TeX}}}

\setcopyright{acmcopyright}
\copyrightyear{2018}
\acmYear{2018}
\acmDOI{XXXXXXX.XXXXXXX}

\acmJournal{JACM}
\acmVolume{37}
\acmNumber{4}
\acmArticle{111}
\acmMonth{8}

\usepackage{amsmath}
\usepackage{acronym}
\usepackage{algorithmic}
\usepackage{booktabs}
\usepackage{multirow}
\usepackage{amssymb} 
\usepackage{amsfonts}  
\usepackage{mathrsfs}  

\usepackage{enumitem}
\usepackage{forest}
\usepackage{array}

\usepackage{xspace}
\usepackage{cite}
\usepackage{hyperref}
\usepackage{natbib}
\usepackage{graphicx}
\usepackage{float}
\usepackage{subfigure}
\usepackage{forest}
\usepackage{url}
\usepackage{todonotes}
\usepackage{xcolor}
\usepackage{comment}
\usepackage{tikz}
\usepackage{pgfplots}
\usepackage{pgfplotstable}

\newcommand{\llms}{LLMs\xspace}
\begin{document}

\title{A Survey on Evaluation of Large Language Models }

\author{Yupeng Chang}
\authornote{Both authors contributed equally to this research.}
\email{ypchang_jluai@outlook.com}
\orcid{ }
\author{Xu Wang}
\authornotemark[1]
\email{xwang22@mails.jlu.edu.cn}
\affiliation{%
  \institution{School of Artificial Intelligence, Jilin University}
  \streetaddress{2699 Qianjin St}
  \city{Changchun}
  \country{China}
  \postcode{130012}
}

\author{Jindong Wang}
\authornote{Corresponding author.}
\affiliation{%
  \institution{Microsoft Research Asia}
  \city{Beijing}
  \country{China}}
\email{Jindong.wang@microsoft.com}

\author{Yuan Wu}
\authornotemark[2]
\affiliation{%
  \institution{School of Artificial Intelligence, Jilin University}
  \city{Changchun}
  \country{China}}
\email{yuanwu@jlu.edu.cn}

\author{Linyi Yang}
\affiliation{%
  \institution{Westlake University}
  \city{Hangzhou}
  \country{China}}

\author{Kaijie Zhu}
\affiliation{%
 \institution{Institute of Automation, Chinese Academy of Sciences}
 \city{Beijing}
 \country{China}}

\author{Hao Chen}
\affiliation{%
  \institution{Carnegie Mellon University}
  \state{Pennsylvania}
  \country{USA}}

\author{Xiaoyuan Yi}
\affiliation{%
  \institution{Microsoft Research Asia}
  \city{Beijing}
  \country{China}}

\author{Cunxiang Wang}
\affiliation{%
  \institution{Westlake University}
  \city{Hangzhou}
  \country{China}}

\author{Yidong Wang}
\affiliation{
  \institution{Peking University}
  \city{Beijing}
  \country{China}}

\author{Wei Ye}
\affiliation{
  \institution{Peking University}
  \city{Beijing}
  \country{China}}

\author{Yue Zhang}
\affiliation{%
  \institution{Westlake University}
  \city{Hangzhou}
  \country{China}}
  
\author{Yi Chang}
\affiliation{%
  \institution{School of Artificial Intelligence, Jilin University}
  \city{Changchun}
  \country{China}}
  
\author{Philip S. Yu}
\affiliation{
  \institution{University of Illinois at Chicago}
  \state{Illinois}
  \country{USA}}

\author{Qiang Yang}
\affiliation{
  \institution{Hong Kong University of Science and Technology}
  \city{Kowloon}
  \state{Hong Kong}
  \country{China}}

\author{Xing~Xie}
\affiliation{%
  \institution{Microsoft Research Asia}
  \city{Beijing}
  \country{China}}

\renewcommand{\shortauthors}{Chang et al.}

\begin{abstract}
\label{myabstract}
Large language models (\llms) are gaining increasing popularity in both academia and industry, owing to their unprecedented performance in various applications. As \llms continue to play a vital role in both research and daily use, their evaluation becomes increasingly critical, not only at the task level, but also at the society level for better understanding of their potential risks. Over the past years, significant efforts have been made to examine \llms from various perspectives. 
This paper presents a comprehensive review of these evaluation methods for \llms, focusing on three key dimensions: \emph{what to evaluate}, \emph{where to evaluate}, and \emph{how to evaluate}.
Firstly, we provide an overview from the perspective of evaluation tasks, encompassing general natural language processing tasks, reasoning, medical usage, ethics, education, natural and social sciences, agent applications, and other areas. 
Secondly, we answer the `where' and `how' questions by diving into the evaluation methods and benchmarks, which serve as crucial components in assessing the performance of \llms.
Then, we summarize the success and failure cases of \llms in different tasks.
Finally, we shed light on several future challenges that lie ahead in \llms evaluation. Our aim is to offer invaluable insights to researchers in the realm of \llms evaluation, thereby aiding the development of more proficient \llms. Our key point is that evaluation should be treated as an essential discipline to better assist the development of \llms. We consistently maintain the related open-source materials at: \url{https://github.com/MLGroupJLU/LLM-eval-survey}.
\end{abstract}




\received{20 February 2007}
\received[revised]{12 March 2009}
\received[accepted]{5 June 2009}

\maketitle

\section{Introduction}
\label{sec:introduction}

Understanding the essence of intelligence and establishing whether a machine embodies it poses a compelling question for scientists. 
It is generally agreed upon that authentic intelligence equips us with reasoning capabilities, enables us to test hypotheses, and prepares for future eventualities~\citep{khalfa1994intelligence}. 
In particular, Artificial Intelligence (AI) researchers focus on the development of machine-based intelligence, as opposed to biologically based intellect~\citep{mccarthy2007artificial}.
Proper measurement helps to understand intelligence.
For instance, measures for general intelligence in human individuals often encompass IQ tests~\citep{brody1999intelligence}.


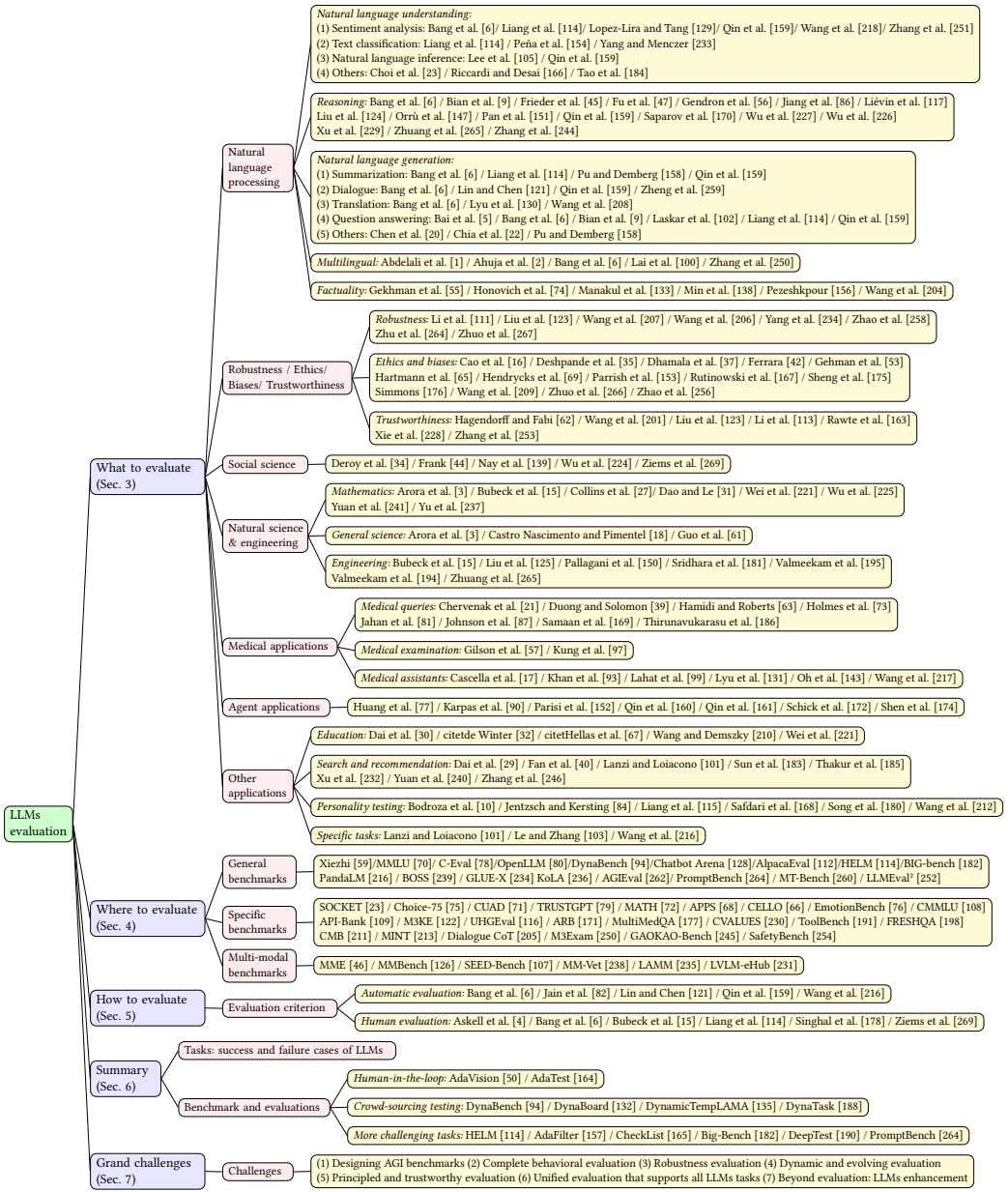
\begin{figure*}[t!]
	\centering
	\resizebox{\textwidth}{!}{
	\input{Figures/fig-tree}
	}
	\caption{Structure of this paper.}
	\label{fig-main}
\end{figure*}

Within the scope of AI, the Turing Test~\citep{turing2009computing}, a widely recognized test for assessing intelligence by discerning if responses are of human or machine origin, has been a longstanding objective in AI evolution. 
It is generally believed among researchers that a computing machine that successfully passes the Turing Test can be considered as intelligent. 
Consequently, when viewed from a wider lens, the chronicle of AI can be depicted as the timeline of creation and evaluation of intelligent models and algorithms. 
With each emergence of a novel AI model or algorithm, researchers invariably scrutinize its capabilities in real-world scenarios through evaluation using specific and challenging tasks. 
For instance, the Perceptron algorithm~\citep{gallant1990perceptron}, touted as an Artificial General Intelligence (AGI) approach in the 1950s, was later revealed as inadequate due to its inability to resolve the XOR problem. 
The subsequent rise and application of Support Vector Machines (SVMs)~\citep{cortes1995support} and deep learning~\citep{lecun2015deep} have marked both progress and setbacks in the AI landscape. 
A significant takeaway from previous attempts is the paramount importance of AI evaluation, which serves as a critical tool to identify current system limitations and inform the design of more powerful models.


Recently, large language models (\llms) have incited substantial interest across both academic and industrial domains~\citep{wei2022emergent,bommasani2021opportunities,zhao2023survey}.
As demonstrated by existing work~\citep{bubeck2023sparks}, the great performance of \llms has raised promise that they could be AGI in this era.
\llms possess the capabilities to solve diverse tasks, contrasting with prior models confined to solving specific tasks.
Due to its great performance in handling different applications such as general natural language tasks and domain-specific ones, 
\llms are increasingly used by individuals with critical information needs, such as students or patients.

Evaluation is of paramount prominence to the success of \llms due to several reasons.
First, evaluating \llms helps us better understand the strengths and weakness of \llms.
For instance, the PromptBench~\citep{zhu2023promptbench} benchmark illustrates that current \llms are sensitive to adversarial prompts, thus a careful prompt engineering is necessary for better performance.
Second, better evaluations can provide better guidance for human-\llms interaction, which could inspire future interaction design and implementation.
Third, the broad applicability of \llms underscores the paramount importance of ensuring their safety and reliability, particularly in safety-sensitive sectors such as financial institutions and healthcare facilities.
Finally, as \llms are becoming larger with more emergent abilities, existing evaluation protocols may not be enough to evaluate their capabilities and potential risks.
Therefore, we aim to raise awareness in the community of the importance to \llms evaluations by reviewing the current evaluation protocols and most importantly, shed light on future research about designing new \llms evaluation protocols.

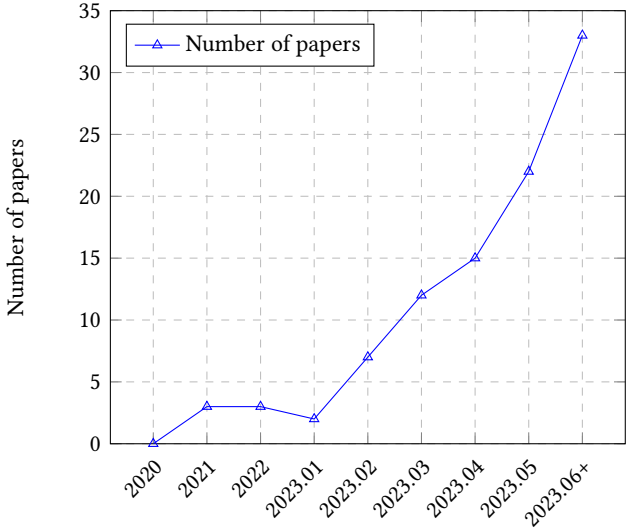
\begin{figure}[t!]
\centering
\input{Figures/fig-numpapers}
\caption{Trend of \llms evaluation papers over time (2020 - Jun. 2023, including Jul. 2023.).}
\label{fig-numbers}
\end{figure}

With the introduction of ChatGPT~\citep{chatgpt} and GPT-4~\citep{openai2023gpt4}, there have been a number of research efforts aiming at evaluating ChatGPT and other \llms from different aspects (\figurename~\ref{fig-numbers}),  encompassing a range of factors such as natural language tasks, reasoning, robustness, trustworthiness, medical applications, and ethical considerations.
Despite these efforts, a comprehensive overview capturing the entire gamut of evaluations is still lacking.
Furthermore, the ongoing evolution of \llms has also presented novel aspects for evaluation, thereby challenging existing evaluation protocols and reinforcing the need for thorough, multifaceted evaluation techniques.
While existing research such as \citet{bubeck2023sparks} claimed that GPT-4 can be seen as sparks of AGI, others contest this claim due to the human-crafted nature of its evaluation approach.


This paper serves as the first comprehensive survey on the evaluation of large language models.
As depicted in \figurename~\ref{fig-main}, we explore existing work in three dimensions: 1) What to evaluate, 2) Where to evaluate, and 3) How to evaluate. Specifically, ``what to evaluate" encapsulates existing evaluation tasks for \llms, ``where to evaluate" involves selecting appropriate datasets and benchmarks for evaluation, while ``how to evaluate" is concerned with the evaluation process given appropriate tasks and datasets. These three dimensions are integral to the evaluation of \llms. We subsequently discuss potential future challenges in the realm of \llms evaluation.

The contributions of this paper are as follows:
\begin{enumerate}
    \item We provide a comprehensive overview of \llms evaluations from three aspects: what to evaluate, where to evaluate, and how to evaluate. Our categorization is general and encompasses the entire life cycle of \llms evaluation.
    \item Regarding what to evaluate, we summarize existing tasks in various areas and obtain insightful conclusions on the success and failure case of \llms (Sec.~\ref{sec-summary}), providing experience for future research.
    \item As for where to evaluate, we summarize evaluation metrics, datasets, and benchmarks to provide a profound understanding of current \llms evaluations. In terms of how to evaluate, we explore current protocols and summarize novel evaluation approaches.
    \item We further discuss future challenges in evaluating \llms. We open-source and maintain the related materials of \llms evaluation at \url{https://github.com/MLGroupJLU/LLM-eval-survey} to foster a collaborative community for better evaluations.
\end{enumerate}

The paper is organized as follows.
In Sec.~\ref{sec-back}, we provide the basic information of \llms and AI model evaluation.
Then, Sec.~\ref{sec-what} reviews existing work from the aspects of ``what to evaluate''.
After that, Sec.~\ref{sec-where} is the ``where to evaluate'' part, which summarizes existing datasets and benchmarks.
Sec.~\ref{sec-how} discusses how to perform the evaluation.
In Sec.~\ref{sec-summary}, we summarize the key findings of this paper.
We discuss grand future challenges in Sec.~\ref{sec-challenge} and Sec.~\ref{sec-con} concludes the paper.

\section{Background}
\label{sec-back}
\input{sec-background}

\section{What to Evaluate}
\label{sec-what}

\input{sec-what-to-eval}
\section{Where to Evaluate: Datasets and Benchmarks}
\label{sec-where}

\input{sec-benchmark-dataset}

\section{How to Evaluate}
\label{sec-how}

\input{sec-how-to-eval}

\section{Summary}
\label{sec-summary}

\input{sec-summary}

\section{Grand Challenges and Opportunities for Future Research}
\label{sec-challenge}

\input{sec-challenge}

\section{Conclusion}
\label{sec-con}

Evaluation carries profound significance, becoming imperative in the advancement of AI models, especially within the context of large language models.
This paper presents the first survey to give a comprehensive overview of the evaluation on \llms from three aspects: what to evaluate, how to evaluate, and where to evaluate.
By encapsulating evaluation tasks, protocols, and benchmarks, our aim is to augment understanding of the current status of \llms, elucidate their strengths and limitations, and furnish insights for future \llms progression.

Our survey reveals that current \llms exhibit certain limitations in numerous tasks, notably reasoning and robustness tasks.
Concurrently, the need for contemporary evaluation systems to adapt and evolve remains evident, ensuring the accurate assessment of \llms' inherent capabilities and limitations. 
We identify several grand challenges that future research should address, with the aspiration that \llms can progressively enhance their service to humanity.

\section*{Acknowledgements}

This work is supported in part by NSF under grant III-2106758. 

\section*{Disclaimer}

The goal of this paper is mainly to summarize and discuss existing evaluation efforts on large language models.
Results and conclusions in each paper are original contributions of their corresponding authors, particularly for potential issues in ethics and biases.
This paper may discuss some side effects of \llms and the only intention is to foster a better understanding.

Due to the evolution of \llms especially online services such as Claude and ChatGPT, it is very likely that they become stronger and some of the limitations described in this paper are mitigated (and new limitations may arise).
We encourage interested readers to take this survey as a reference for future research and conduct real experiments in current systems when performing evaluations.

Finally, the evaluation of \llms is continuously developing, thus we may miss some new papers or benchmarks. 
We welcome all constructive feedback and suggestions.


\bibliographystyle{ACM-Reference-Format}
\bibliography{refs}


\end{document}

%% file: Figures/fig-tree.tex
\begin{forest}
  for tree={
  grow=east,
  reversed=true,
  anchor=base west,
  parent anchor=east,
  child anchor=west,
  base=left,
  font=\small,
  rectangle,
  draw,
  rounded corners,align=left,
  minimum width=2.5em,
  inner xsep=4pt,
  inner ysep=1pt,
  },
  where level=1{text width=5em,fill=blue!10}{},
  where level=2{text width=5em,font=\footnotesize,fill=pink!30}{},
  where level=3{font=\footnotesize,yshift=0.26pt,fill=yellow!20}{},
  [\llms \\ evaluation,fill=green!20
        [What to evaluate\\(Sec.~\ref{sec-what}),text width=7em
            [Natural\\language\\processing,text width=4em
              [\emph{Natural language understanding:} \\
              (1) Sentiment analysis: 
              \citet{bang2023multitask}/ \citet{liang2022holistic}/ \citet{lopez2023can}/ \citet{qin2023chatgpt}/ \citet{wang2023chatgpt1}/ \citet{zhang2023sentiment} \\ 
              (2) Text classification: 
              \citet{liang2022holistic} / \citet{pena2023leveraging} / \citet{yang2023large}   \\
              (3) Natural language inference: 
              \citet{lee2023can} / \citet{qin2023chatgpt}  \\
              (4) Others: 
              \citet{choi2023llms} / \citet{riccardi2023two} /  \citet{tao2023eveval} ]
              [\emph{Reasoning:} 
              \citet{bang2023multitask} / \citet{bian2023chatgpt} / \citet{frieder2023mathematical} / \citet{fu2023chain} / \citet{gendron2023large} / \citet{jiang2023structgpt} / \citet{lievin2022can} \\ \citet{liu2023evaluating} /   
              \citet{orru2023human} / \citet{pan2023unifying} / \citet{qin2023chatgpt} /  \citet{saparov2023testing} / \citet{wu2023reasoning} / \citet{wu2022autoformalization} \\ \citet{xu2023large} / \citet{zhuang2023efficiently} / \citet{zhang2023evaluating1} \\
              ]
              [\emph{Natural language generation:} \\
              (1) Summarization: 
              \citet{bang2023multitask} / \citet{liang2022holistic} / \citet{pu2023chatgpt} / \citet{qin2023chatgpt}\\
              (2) Dialogue: 
              \citet{bang2023multitask} / \citet{lin2023llm} / \citet{qin2023chatgpt} / \citet{zheng2023lmsys}  \\
              (3) Translation: 
              \citet{bang2023multitask} / \citet{lyu2023new} / \citet{wang2023document}  \\
              (4) Question answering: 
              \citet{bai2023benchmarking} / \citet{bang2023multitask} / \citet{bian2023chatgpt} / \citet{laskar2023systematic} / \citet{liang2022holistic} / \citet{qin2023chatgpt} \\
              (5) Others: 
              \citet{chen2023exploring} / \citet{chia2023instructeval} / \citet{pu2023chatgpt}
              ]
              [\emph{Multilingual:} 
              \citet{abdelali2023benchmarking} / \citet{ahuja2023mega} / \citet{bang2023multitask} / \citet{lai2023chatgpt} / \citet{zhang2023m3exam}]
              [\emph{Factuality:} 
              \citet{gekhman2023trueteacher} / \citet{honovich2022true} / \citet{manakul2023selfcheckgpt} / \citet{min2023factscore} / \citet{pezeshkpour2023measuring} / \citet{wang2023evaluating}]
            ]
            [Robustness / Ethics/ \\Biases/ Trustworthiness,text width=8em 
                [ \emph{Robustness:} 
                \citet{li2023survey} / \citet{liu2023mitigating} / \citet{wang2022generalizing} / \citet{wang2023robustness} / \citet{yang2022glue} / \citet{zhao2023evaluating} \\ \citet{zhu2023promptbench} / \citet{zhuo2023robustness}
                ]
                [ \emph{Ethics and biases:} 
                \citet{cao2023assessing} / \citet{deshpande2023toxicity} / \citet{dhamala2021bold} / \citet{ferrara2023should} /  
                \citet{gehman2020realtoxicityprompts} \\  \citet{hartmann2023political} / 
                \citet{hendrycks2020aligning} / 
                \citet{parrish2022bbq} / \citet{rutinowski2023self} / 
                \citet{sheng2021societal} \\ \citet{simmons2022moral} / \citet{wang2023large} / \citet{zhuo2023exploring} / \citet{zhao2023chbias} 
                ]
                [ \emph{Trustworthiness:} 
                \citet{hagendorff2023humanlike} /  \citet{wang2023decodingtrust} / \citet{liu2023mitigating} / \citet{li2023evaluating} / \citet{rawte2023survey} \\ 
                \citet{xie2023ask} / \citet{zhang2023sirens} 
                ]
            ]
            [Social science, text width=5em
                [\citet{deroy2023ready} / \citet{frank2023baby} / \citet{nay2023large} / \citet{wu2023large} / \citet{ziems2023can} ]
            ]    
            [Natural science\\\& engineering, text width=5em
                [\emph{Mathematics:} 
                \citet{arora2023have} / \citet{bubeck2023sparks} / \citet{collins2023evaluating}/  \citet{dao2023investigating} / \citet{wei2023cmath} / 
                \citet{wu2023empirical} \\ \citet{yuan2023well} / \citet{yu2023metamath}
                ]
                [\emph{General science:} 
                \citet{arora2023have} / \citet{castro2023large} / \citet{guo2023indeed}]
                [\emph{Engineering:} 
                \citet{bubeck2023sparks} / \citet{liu2023your} / \citet{pallagani2023understanding} / \citet{sridhara2023chatgpt} / \citet{valmeekam2022large}\\  
                \citet{valmeekam2023planning} / \citet{zhuang2023efficiently}
                ]
            ]
            [Medical applications, text width=7em 
                [\emph{Medical queries:} 
                \citet{chervenak2023promise} / \citet{duong2023analysis} / \citet{hamidi2023evaluation} / \citet{holmes2023evaluating} \\ \citet{jahan2023evaluation} /  
                \citet{johnson2023assessing} / \citet{samaan2023assessing} / \citet{thirunavukarasu2023trialling} ]
                [\emph{Medical examination:} 
                \citet{gilson2023does} / \citet{kung2023performance}]
                [\emph{Medical assistants:} 
                \citet{cascella2023evaluating} / \citet{khan2023covllm} / \citet{lahat2023evaluating} / 
                \citet{lyu2023translating} / \citet{oh2023chatgpt} / \citet{wang2023can}]
            ]
            [Agent applications, text width=6.5em
            [
            \citet{huang2023language} / \citet{karpas2022mrkl} / \citet{parisi2022talm} / \citet{qin2023tool} / \citet{qin2023toolllm} / \citet{schick2023toolformer} / \citet{shen2023hugginggpt}
            ]]
            [Other\\applications, text width=4em
                [\emph{Education:} 
                \citet{dai2023can} / citet\citet{de2023can} / citet\citet{hellas2023exploring} / \citet{wang2023chatgpt} / \citet{wei2023cmath}]
                [\emph{Search and recommendation:} 
                \citet{dai2023uncovering} / \citet{fan2023recommender} / \citet{lanzi2023chatgpt} / \citet{sun2023chatgpt} / \citet{thakur2021beir} \\ \citet{xu2023chatgpt} / \citet{yuan2023recommender} / \citet{zhang2023chatgpt}]
                [\emph{Personality testing:} 
                \citet{bodroza2023personality} / \citet{jentzsch2023chatgpt} / \citet{liang2023leveraging} / \citet{safdari2023personality} / \citet{song2023have} / \citet{wang2023emotional}] 
                [\emph{Specific tasks:} 
                \citet{lanzi2023chatgpt} / \citet{le2023evaluation} / \citet{wang2023pandalm}]
            ]
        ]
        [Where to evaluate\\(Sec.~\ref{sec-where}),text width=7em
          [General\\benchmarks,text width=4.2em
            [Xiezhi~\citep{gu2023xiezhi}/MMLU~\citep{hendrycks2020measuring}/ C-Eval~\citep{huang2023c}/OpenLLM~\citep{leaderboard}/DynaBench~\citep{kiela2021dynabench}/Chatbot Arena~\citep{chatbotarena}/AlpacaEval~\citep{alpaca_eval}/HELM~\citep{liang2022holistic}/BIG-bench~\citep{srivastava2022beyond}\\ PandaLM~\citep{wang2023pandalm} / BOSS~\citep{yuan2023revisiting} / GLUE-X~\citep{yang2022glue} 
            KoLA~\citep{yu2023kola} / AGIEval~\citep{zhong2023agieval}/ PromptBench~\citep{zhu2023promptbench} / MT-Bench~\citep{zheng2023judging} / LLMEval²~\citep{zhang2023wider}]
            ]
          [Specific\\ benchmarks,text width=4.2em
            [SOCKET~\citep{choi2023llms} / Choice-75~\citep{hou2023choice} / 
            CUAD~\citep{hendrycks2021cuad} / 
            TRUSTGPT~\citep{huang2023trustgpt} / MATH~\citep{hendrycks2021measuring-1} / APPS~\citep{hendrycks2021measuring-2} / CELLO~\citep{he2023can} / EmotionBench~\citep{huang2023emotionally} / CMMLU~\citep{li2023cmmlu} \\ API-Bank~\citep{li2023apibank} /   M3KE~\citep{liu2023m3ke} / UHGEval~\citep{liang2023uhgeval} / ARB~\citep{sawada2023arb} / MultiMedQA~\citep{singhal2022large} /  CVALUES~\citep{xu2023cvalues} / ToolBench~\citep{toolbench} / FRESHQA~\citep{vu2023freshllms} \\ CMB~\citep{wang2023cmb} / MINT~\citep{wang2023mint} / Dialogue CoT~\citep{wang2023chainofthought} / M3Exam~\citep{zhang2023m3exam} / GAOKAO-Bench~\citep{zhang2023evaluating} / SafetyBench~\citep{zhang2023safetybench}]
            ]
             [Multi-modal\\ benchmarks,text width=4.2em
            [MME~\citep{fu2023mme} / MMBench~\citep{liu2023mmbench} / SEED-Bench~\citep{li2023seed} / MM-Vet~\citep{yu2023mm} / LAMM~\citep{yin2023lamm} / LVLM-eHub~\citep{xu2023lvlmehub}]
            ]
        ]
        [How to evaluate\\(Sec.~\ref{sec-how}),text width=7em
          [Evaluation criterion, text width=7em
              [\emph{Automatic evaluation:} 
              \citet{bang2023multitask} / \citet{jain2023bring} / \citet{lin2023llm} / \citet{qin2023chatgpt} / \citet{wang2023pandalm} 
              ]
              [\emph{Human evaluation:} 
              \citet{askell2021general} / \citet{bang2023multitask} / \citet{bubeck2023sparks} / \citet{liang2022holistic} / \citet{singhal2023large} / \citet{ziems2023can}  
              ]
          ]
        ]
        [Summary\\(Sec.~\ref{sec-summary}),text width=4em
          [Tasks: success and failure cases of \llms,text width=14em
          ]
          [Benchmark and evaluations, text width=9.5em
            [\emph{Human-in-the-loop:} 
            AdaVision~\citep{gao2022adaptive} / AdaTest~\citep{ribeiro2022adaptive}
            ]
            [\emph{Crowd-sourcing testing:} 
            DynaBench~\citep{kiela2021dynabench} / DynaBoard~\citep{ma2021dynaboard} / DynamicTempLAMA~\citep{margatina2023dynamic} / 
            DynaTask~\citep{thrush2022dynatask}
            ]
            [\emph{More challenging tasks:} 
            HELM~\citep{liang2022holistic} / AdaFilter~\citep{phang2021adversarially} / CheckList~\citep{ribeiro2020beyond} / Big-Bench~\citep{srivastava2022beyond} / 
            DeepTest~\citep{tian2018deeptest} / PromptBench~\citep{zhu2023promptbench}
            ]
          ]
        ]
        [Grand challenges\\(Sec.~\ref{sec-challenge}),text width=7em
          [Challenges,text width=4em
            [(1) Designing AGI benchmarks (2) Complete behavioral evaluation (3) Robustness evaluation (4) Dynamic and evolving evaluation \\  (5) Principled and trustworthy evaluation (6) Unified evaluation that supports all \llms tasks (7) Beyond evaluation: \llms enhancement]
            ]
        ]
    ]
\end{forest}

%% file: Figures/fig-numpapers.tex
\begin{tikzpicture}
\pgfplotstableread{
Year Papers
2020 0
2021 3
2022 3
2023.01 2
2023.02 7
2023.03 12 
2023.04 15
2023.05 22
2023.06+ 33
}\datatable

\begin{axis}[
    ylabel={Number of papers},
    xtick={1,2,3,4,5,6,7,8,9},
    xticklabels={2020,2021,2022,2023.01,2023.02,2023.03,2023.04,2023.05,2023.06+},
    xticklabel style={rotate=45, anchor=north east},
    ymin=0,
    ymax=35,
    ytick distance=5,
    legend pos=north west,
    ymajorgrids=true,
    grid=both,
    grid style=dashed,
    clip mode=individual,
    font=\small,
]

\addplot[color=blue, mark=triangle, mark options={fill=blue}] table[x expr=\coordindex+1, y=Papers] {\datatable};

\legend{Number of papers}
\end{axis}
\end{tikzpicture}

%% file: sec-background.tex
\subsection{Large Language Models}

Language models (LMs) \citep{gao2004introduction,kombrink2011recurrent,devlin2018bert} are computational models that have the capability to understand and generate human language.
LMs have the transformative ability to predict the likelihood of word sequences or generate new text based on a given input.
N-gram models \citep{brown1992class}, the most common type of LM, estimate word probabilities based on the preceding context. 
However, LMs also face challenges, such as the issue of rare or unseen words, the problem of overfitting, and the difficulty in capturing complex linguistic phenomena. Researchers are continuously working on improving LM architectures and training methods to address these challenges. 

Large Language Models (LLMs) \citep{kasneci2023chatgpt,zhao2023survey,chen2021evaluating} are advanced language models with massive parameter sizes and exceptional learning capabilities.
The core module behind many LLMs such as GPT-3 \citep{floridi2020gpt}, InstructGPT \citep{ouyang2022training}, and GPT-4 \citep{openai2023gpt4} is the self-attention module in Transformer
\citep{vaswani2017attention} that serves as the fundamental building block for language modeling tasks. Transformers have revolutionized the field of NLP with their ability to handle sequential data efficiently, allowing for parallelization and capturing long-range dependencies in text.
One key feature of \llms is in-context learning \citep{brown2020language}, where the model is trained to generate text based on a given context or prompt. This enables LLMs to generate more coherent and contextually relevant responses, making them suitable for interactive and conversational applications.
Reinforcement Learning from Human Feedback (RLHF) \citep{ziegler2019fine,christiano2017deep} is another crucial aspect of \llms. This technique involves fine-tuning the model using human-generated responses as rewards, allowing the model to learn from its mistakes and improve its performance over time.

In an autoregressive language model, such as GPT-3 and PaLM \citep{chowdhery2022palm}, given a context sequence \(X\), the LM tasks aim to predict the next token \(y\). The model is trained by maximizing the probability of the given token sequence conditioned on the context, i.e., $P(y | X) = P(y | x_1, x_2, ..., x_{t-1})$, where \(x_1, x_2, ..., x_{t-1}\) are the tokens in the context sequence, and \(t\) is the current position. By using the chain rule, the conditional probability can be decomposed into a product of probabilities at each position:
\begin{equation*}
P(y | X) = \prod_{t=1}^{T} P(y_t | x_1, x_2, ..., x_{t-1}),
\label{eq:autoregressive}
\end{equation*}
where \(T\) is sequence length. In this way, the model predicts each token at each position in an autoregressive manner, generating a complete text sequence.

One common approach to interacting with \llms is prompt engineering \citep{zhou2022large,white2023prompt,clavie2023large}, where users design and provide specific prompt texts to guide LLMs in generating desired responses or completing specific tasks. This is widely adopted in existing evaluation efforts.
People can also engage in question-and-answer interactions \citep{jansson2021online}, where they pose questions to the model and receive answers, or engage in dialogue interactions, having natural language conversations with LLMs.
In conclusion, \llms, with their Transformer architecture, in-context learning, and RLHF capabilities, have revolutionized NLP and hold promise in various applications.
\tablename~\ref{tb-compare} provides a brief comparison of traditional ML, deep learning, and \llms.


\input{tables/tb-related}

\subsection{AI Model Evaluation}

AI model evaluation is an essential step in assessing the performance of a model.
There are some standard model evaluation protocols, including \(k\)-fold cross-validation, holdout validation, leave one out cross-validation (LOOCV), bootstrap, and reduced set \citep{kohavi1995study,berrar2019cross}.
For instance, \(k\)-fold cross-validation divides the dataset into \(k\) parts, with one part used as a test set and the rest as training sets, which can reduce training data loss and obtain relatively more accurate model performance evaluation~\citep{fushiki2011estimation}; Holdout validation divides the dataset into training and test sets, with a smaller calculation amount but potentially more significant bias; LOOCV is a unique \(k\)-fold cross-validation method where only one data point is used as the test set~\citep{wong2015performance}; Reduced set trains the model with one dataset and tests it with the remaining data, which is computationally simple, but the applicability is limited.
The appropriate evaluation method should be chosen according to the specific problem and data characteristics for more reliable performance indicators.

\begin{figure}[t!]
    \centering
    \includegraphics[width=.7\textwidth]{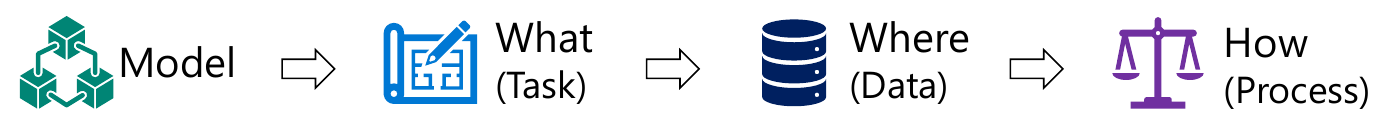}
    \caption{The evaluation process of AI models.}
    \label{fig-eval-proc}
\end{figure}

\figurename~\ref{fig-eval-proc} illustrates the evaluation process of AI models, including \llms.
Some evaluation protocols may not be feasible to evaluate deep learning models due to the extensive training size.
Thus, evaluation on a static validation set has long been the standard choice for deep learning models.
For instance, computer vision models leverage static test sets such as ImageNet~\citep{deng2009imagenet} and MS COCO~\citep{lin2014microsoft} for evaluation.
\llms also use GLUE~\citep{wang2018glue} or SuperGLUE~\citep{wang2019superglue} as the common test sets.

As \llms are becoming more popular with even poorer interpretability, existing evaluation protocols may not be enough to evaluate the true capabilities of \llms thoroughly.
We will introduce recent evaluations of \llms in Sec.~\ref{sec-how}.

%% file: tables/tb-related.tex
\begin{table}[h!]
\caption{Comparison of Traditional ML, Deep Learning, and \llms}
\label{tb-compare}
\centering
\resizebox{.63\textwidth}{!}{
\begin{tabular}{|l|c|c|c|}
\hline
\textbf{Comparison} & \textbf{Traditional ML} &  \textbf{Deep Learning} & \textbf{\llms}\\
\hline\hline
Training Data Size & Large& Large& Very large\\\hline
Feature Engineering & Manual & Automatic & Automatic\\\hline
Model Complexity & Limited & Complex & Very Complex\\\hline
Interpretability & Good & Poor & Poorer\\\hline
Performance & Moderate & High & Highest \\\hline
Hardware Requirements & Low & High & Very High \\\hline

\end{tabular}}
\end{table}

%% file: sec-what-to-eval.tex
What tasks should we evaluate \llms to show their performance?
On what tasks can we claim the strengths and weaknesses of \llms?
In this section, we divide existing tasks into the following categories: natural language processing, robustness, ethics, biases and trustworthiness, social sciences, natural science and engineering, medical applications, agent applications (using \llms as agents), and other applications.\footnote{Note that \llms are evaluated in various tasks and the categorization in this paper is only one possible way for classification of these works. There are certainly other taxonomies.}

\subsection{Natural Language Processing Tasks}

The initial objective behind the development of language models, particularly large language models, was to enhance performance on natural language processing tasks, encompassing both understanding and generation.
Consequently, the majority of evaluation research has been primarily focused on natural language tasks.
\tablename~\ref{tb-what-nlptask} summarizes the evaluation aspects of existing research, and we mainly highlight their conclusions in the following.\footnote{Several NLP areas have intersections and thus our categorization of these areas is only one possible way to categorize.}

\input{tables/tb-nlptask}

\subsubsection{Natural language understanding}
Natural language understanding represents a wide spectrum of tasks that aims to obtain a better understanding of the input sequence.
We summarize recent efforts in \llms evaluation from several aspects.

\textbf{Sentiment analysis} is a task that analyzes and interprets the text to determine the emotional inclination.
It is typically a binary (positive and negative) or triple (positive, neutral, and negative) class classification problem.
Evaluating sentiment analysis tasks is a popular direction.
\citet{liang2022holistic} and \citet{zeng2022glm} showed that the performance of the models on this task is usually high.
ChatGPT's sentiment analysis prediction performance is superior to traditional sentiment analysis methods \citep{lopez2023can} and comes close to that of GPT-3.5 \citep{qin2023chatgpt}.
In fine-grained sentiment and emotion cause analysis, ChatGPT also exhibits exceptional performance \citep{wang2023chatgpt1}.
In low-resource learning environments, \llms exhibit significant advantages over small language models \citep{zhang2023sentiment}, but the ability of ChatGPT to understand low-resource languages is limited \citep{bang2023multitask}.
In conclusion, \llms have demonstrated commendable performance in sentiment analysis tasks. Future work should focus on enhancing their capability to understand emotions in under-resourced languages.

\textbf{Text classification} and sentiment analysis are related fields, text classification not only focuses on sentiment, but also includes the processing of all texts and tasks.
The work of \citet{liang2022holistic} showed that GLM-130B was the best-performed model, with an overall accuracy of 85.8\% for miscellaneous text classification.
\citet{yang2023large} found that ChatGPT can produce credibility ratings for a wide range of news outlets, and these ratings have a moderate correlation with those from human experts.
Furthermore, ChatGPT achieves acceptable accuracy in a binary classification scenario (AUC=0.89).
\citet{pena2023leveraging} discussed the problem of topic classification for public affairs documents and showed that using an LLM backbone in combination with SVM classifiers is a useful strategy to conduct the multi-label topic classification task in the domain of public affairs with accuracies over 85\%.
Overall, \llms perform well on text classification and can even handle text classification tasks in unconventional problem settings as well.

\textbf{Natural language inference (NLI)} is the task of determining whether the given ``hypothesis'' logically follows from the ``premise''.
\citet{qin2023chatgpt} showed that ChatGPT outperforms GPT-3.5 for NLI tasks.
They also found that ChatGPT excels in handling factual input that could be attributed to its RLHF training process in favoring human feedback. 
However, \citet{lee2023can} observed \llms perform poorly in the scope of NLI and further fail in representing human disagreement, which indicates that \llms still have a large room for improvement in this field.

\textbf{Semantic understanding} refers to the meaning or understanding of language and its associated concepts. It involves the interpretation and comprehension of words, phrases, sentences, and the relationships between them. Semantic processing goes beyond the surface level and focuses on understanding the underlying meaning and intent.
\citet{tao2023eveval} comprehensively evaluated the event semantic processing abilities of \llms covering understanding, reasoning, and prediction about the event semantics. Results indicated that \llms possess an understanding of individual events, but their capacity to perceive the semantic similarity among events is constrained. In reasoning tasks, \llms exhibit robust reasoning abilities in causal and intentional relations, yet their performance in other relation types is comparatively weaker. In prediction tasks, \llms exhibit enhanced predictive capabilities for future events with increased contextual information.
\citet{riccardi2023two} explored the semantic proficiency of \llms and showed that these models perform poorly in evaluating basic phrases.
Furthermore, GPT-3.5 and Bard cannot distinguish between meaningful and nonsense phrases, consistently classifying highly nonsense phrases as meaningful.
GPT-4 shows significant improvements, but its performance is still significantly lower than that of humans. 
In summary, the performance of \llms in semantic understanding tasks is poor. In the future, we can start from this aspect and focus on improving its performance on this application.

In \textbf{social knowledge understanding}, \citet{choi2023llms} evaluated how well models perform at learning and recognizing concepts of social knowledge and the results revealed that despite being much smaller in the number of parameters, finetuning supervised models such as BERT lead to much better performance than zero-shot models using state-of-the-art \llms, such as GPT \citep{radford2018improving}, GPT-J-6B \citep{wang2021gpt} and so on.
This statement demonstrates that supervised models significantly outperform zero-shot models in terms of performance, highlighting that an increase in parameters does not necessarily guarantee a higher level of social knowledge in this particular scenario.

\subsubsection{Reasoning}\label{sec:reasoning}
The task of reasoning poses significant challenges for an intelligent AI model.
To effectively tackle reasoning tasks, the models need to not only comprehend the provided information but also utilize reasoning and inference to deduce answers when explicit responses are absent.
\tablename~\ref{tb-what-nlptask} reveals that there is a growing interest in evaluating the reasoning ability of \llms, as evidenced by the increasing number of articles focusing on exploring this aspect. 
Currently, the evaluation of reasoning tasks can be broadly categorized into mathematical reasoning, commonsense reasoning, logical reasoning, and domain-specific reasoning.

ChatGPT exhibits a strong capability for arithmetic reasoning by outperforming GPT-3.5 in the majority of tasks \citep{qin2023chatgpt}. 
However, its proficiency in mathematical reasoning still requires improvement \citep{bang2023multitask,frieder2023mathematical, zhuang2023efficiently}.
On symbolic reasoning tasks, ChatGPT is mostly worse than GPT-3.5, which may be because ChatGPT is prone to uncertain responses, leading to poor performance \citep{bang2023multitask}. 
Through the poor performance of \llms on task variants of counterfactual conditions, \citet{wu2023reasoning} showed that the current \llms have certain limitations in abstract reasoning ability.
On abstract reasoning, \citet{gendron2023large} found that existing \llms have very limited ability.
In logical reasoning, \citet{liu2023evaluating} indicated that ChatGPT and GPT-4 outperform traditional fine-tuning methods on most benchmarks, demonstrating their superiority in logical reasoning. 
However, both models face challenges when handling new and out-of-distribution data.
ChatGPT does not perform as well as other \llms, including GPT-3.5 and BARD \citep{xu2023large,qin2023chatgpt}.
This is because ChatGPT is designed explicitly for chatting, so it does an excellent job of maintaining rationality.
FLAN-T5, LLaMA, GPT-3.5, and PaLM perform well in general deductive reasoning tasks \citep{saparov2023testing}.
GPT-3.5 is not good at keeping oriented for reasoning in the inductive setting \citep{xu2023large}.
For multi-step reasoning, \citet{fu2023chain} showed PaLM and Claude2 are the only two model families that achieve similar performance (but still worse than the GPT model family).
Moreover, LLaMA-65B is the most robust open-source \llms to date, which performs closely to code-davinci-002.
Some papers separately evaluate the performance of ChatGPT on some reasoning tasks: ChatGPT generally performs poorly on commonsense reasoning tasks, but relatively better than non-text semantic reasoning \citep{bang2023multitask}.
Meanwhile, ChatGPT also lacks spatial reasoning ability, but exhibits better temporal reasoning.
Finally, while the performance of ChatGPT is acceptable on causal and analogical reasoning, it performs poorly on multi-hop reasoning ability, which is similar to the weakness of other \llms on complex reasoning \citep{ott2023thoughtsource}. 
In professional domain reasoning tasks, zero-shot InstructGPT and Codex are capable of complex medical reasoning tasks, but still need to be further improved \citep{lievin2022can}.
In terms of language insight issues, \citet{orru2023human} demonstrated the potential of ChatGPT for solving verbal insight problems, as ChatGPT's performance was comparable to that of human participants. 
It should be noted that most of the above conclusions are obtained for specific data sets.
In contrast, more complex tasks have become the mainstream benchmarks for assessing the capabilities of \llms. These include tasks such as mathematical reasoning \citep{zhang2023evaluating1,wu2022autoformalization,yu2023metamath} and structured data inference \citep{pan2023unifying,jiang2023structgpt}.
Overall, \llms show great potential in reasoning and show a continuous improvement trend, but still face many challenges and limitations, requiring more in-depth research and optimization.

\subsubsection{Natural language generation}

NLG evaluates the capabilities of \llms in generating specific texts, which consists of several tasks, including summarization, dialogue generation, machine translation, question answering, and other open-ended generation tasks.

\textbf{Summarization} is a generation task that aims to learn a concise abstract for the given sentence.
In this evaluation, \citet{liang2022holistic} found that TNLG v2 (530B) \citep{smith2022using} achieved the highest score in both scenarios, followed by OPT (175B) \citep{zhang2022opt} in second place.
The fine-tuned Bart \citep{lewis2019bart} is still better than zero-shot ChatGPT.
Specifically, ChatGPT demonstrates comparable zero-shot performance to the text-davinci-002 \citep{bang2023multitask}, but performs worse than GPT-3.5 \citep{qin2023chatgpt}.
These findings indicate that \llms, particularly ChatGPT, have a general performance in summarization tasks. 

Evaluating the performance of \llms on \textbf{dialogue} tasks is crucial to the development of dialogue systems and improving human-computer interaction.
Through such evaluation, the natural language processing ability, context understanding ability and generation ability of the model can be improved, so as to realize a more intelligent and more natural dialogue system.
Both Claude and ChatGPT generally achieve better performance across all dimensions when compared to GPT-3.5 \citep{qin2023chatgpt,lin2023llm}.
When comparing the Claude and ChatGPT models, both models demonstrate competitive performance across different evaluation dimensions, with Claude slightly outperforming ChatGPT in specific configurations.
Research by \citet{bang2023multitask} underscores that fully fine-tuned models tailored for specific tasks surpass ChatGPT in both task-oriented and knowledge-based dialogue contexts. 
Additionally, \citet{zheng2023lmsys} have curated a comprehensive LLMs conversation dataset, LMSYS-Chat-1M, encompassing up to one million samples. This dataset serves as a valuable resource for evaluating and advancing dialogue systems.

While \llms are not explicitly trained for \textbf{translation} tasks, they can still demonstrate strong performance.
\citet{wang2023document} demonstrated that ChatGPT and GPT-4 exhibit superior performance in comparison to commercial machine translation (MT) systems, as evaluated by humans. 
Additionally, they outperform most document-level NMT methods in terms of sacreBLEU scores.
During contrastive testing, ChatGPT shows lower accuracy in comparison to traditional translation models. 
However, GPT-4 demonstrates a robust capability in explaining discourse knowledge, even though it may occasionally select incorrect translation candidates.
The findings from \citet{bang2023multitask} indicated that ChatGPT performs X $\to$ Eng translation well, but it still lacks the ability to perform Eng $\to$ X translation.
\citet{lyu2023new} investigated several research directions in MT utilizing \llms. 
This study significantly contributes to the advancement of MT research and highlights the potential of LLMs in enhancing translation capabilities.
In summary, while \llms perform satisfactorily in several translation tasks, there is still room for improvement, e.g., enhancing the translation capability from English to non-English languages.

\textbf{Question answering} is a crucial technology in the field of human-computer interaction, and it has found wide application in scenarios like search engines, intelligent customer service, and QA systems.
The measurement of accuracy and efficiency in QA models will have significant implications for these applications.
According to \citet{liang2022holistic}, among all the evaluated models, InstructGPT davinci v2 (175B) exhibited the highest performance in terms of accuracy, robustness, and fairness across the 9 QA scenarios.
Both GPT-3.5 and ChatGPT demonstrate significant advancements compared to GPT-3 in their ability to answer general knowledge questions. 
In most domains, ChatGPT surpasses GPT-3.5 by more than 2\% in terms of performance \citep{bian2023chatgpt,qin2023chatgpt}.
However, ChatGPT performs slightly weaker than GPT-3.5 on the CommonsenseQA and Social IQA benchmarks. 
This can be attributed to ChatGPT’s cautious nature, as it tends to decline to provide an answer when there is insufficient information available.
Fine-tuned models, such as Vícuna and ChatGPT, exhibit exceptional performance with near-perfect scores, surpassing models that lack supervised fine-tuning by a significant margin \citep{bang2023multitask,bai2023benchmarking}.
\citet{laskar2023systematic} evaluated the effectiveness of ChatGPT on a range of academic datasets, including various tasks such as answering questions, summarizing text, generating code, reasoning with commonsense, solving math problems, translating languages, detecting bias, and addressing ethical issues.
Overall, \llms showcase flawless performance on QA tasks and hold the potential for further enhancing their proficiency in social, event, and temporal commonsense knowledge in the future.

There are also other generation tasks to explore.
In the field of \textbf{sentence style transfer}, \citet{pu2023chatgpt} demonstrated that ChatGPT surpasses the previous SOTA supervised model through training on the same subset for few-shot learning, as evident from the higher BLEU score. 
However, when it comes to controlling the formality of sentence style, ChatGPT’s performance still differs significantly from human behavior.
In \textbf{writing tasks}, \citet{chia2023instructeval} discovered that \llms exhibit consistent performance across various categories such as informative, professional, argumentative, and creative writing. 
This finding implies that \llms possess a general proficiency in writing capabilities.
In \textbf{text generation} quality, \citet{chen2023exploring} revealed that ChatGPT excels in assessing text quality from multiple angles, even in the absence of reference texts, surpassing the performance of most existing automated metrics. 
Employing ChatGPT to generate numerical scores for text quality emerged as the most reliable and effective approach among the various testing methods studied.

\subsubsection{Multilingual tasks}
While English is the predominant language, many \llms are trained on mixed-language training data. 
The combination of multilingual data indeed helps \llms gain the ability to process inputs and generate responses in different languages, making them widely adopted and accepted across the globe. 
However, due to the relatively recent emergence of this technology, \llms are primarily evaluated on English data, leading to a potential oversight of evaluating their multilingual performance. 
To address this, several articles have provided comprehensive, open, and independent evaluations of \llms' performance on various NLP tasks in different non-English languages. 
These evaluations offer valuable insights for future research and applications.

\citet{abdelali2023benchmarking} evaluated the performance of ChatGPT in standard Arabic NLP tasks and observed that ChatGPT exhibits lower performance compared to SOTA models in the zero-shot setting for most tasks.
\citet{bang2023multitask,zhang2023m3exam,lai2023chatgpt, ahuja2023mega} utilized a greater number of languages across multiple datasets, encompassing a wider range of tasks, and conducted a more comprehensive evaluation of \llms, including BLOOM, Vicuna, Claude, ChatGPT, and GPT-4. 
The results indicated that these \llms perform poorly when it came to non-Latin languages and languages with limited resources.
Despite translating the input to English and using it as the query, generative \llms still displays subpar performance across tasks and languages compared to SOTA models \citep{ahuja2023mega}.
Furthermore, \citet{bang2023multitask} highlighted that ChatGPT still faces a limitation in translating sentences written in non-Latin script languages with rich linguistic resources.
The aforementioned demonstrates that there are numerous challenges and ample opportunities for enhancement in multilingual tasks for \llms.
Future research should prioritize achieving multilingual balance and addressing the challenges faced by non-Latin languages and low-resource languages, with the aim of better supporting users worldwide. 
At the same time, attention should be paid to the impartiality and neutrality of the language in order to mitigate any potential biases, including English bias or other biases, that could impact multilingual applications.

\subsubsection{Factuality}

Factuality in the context of \llms refers to the extent to which the information or answers provided by the model align with real-world truths and verifiable facts. Factuality in \llms significantly impacts a variety of tasks and downstream applications, such as QA systems, information extraction, text summarization, dialogue systems, and automated fact-checking, where incorrect or inconsistent information could lead to substantial misunderstandings and misinterpretations. Evaluating factuality is of great importance in order to trust and efficiently use these models. This includes the ability of these models to maintain consistency with known facts, avoid generating misleading or false information (known as ``factual hallucination"), and effectively learn and recall factual knowledge. A range of methodologies have been proposed to measure and improve the factuality of \llms. 

\citet{wang2023evaluating} assessed the internal knowledge capabilities of several large models, namely InstructGPT, ChatGPT-3.5, GPT-4, and BingChat \citep{BingChat}, by examining their ability to answer open questions based on the Natural Questions \citep{NQ} and TriviaQA \citep{TQ} datasets. The evaluation process involved human assessment. The results of the study indicated that while GPT-4 and BingChat can provide correct answers for more than 80\% of the questions, there is still a remaining gap of over 15\% to achieve complete accuracy.
In the work of \citet{honovich2022true}, they conducted a review of current factual consistency evaluation methods and highlighted the absence of a unified comparison framework and the limited reference value of related scores compared to binary labels. To address this, they transformed existing fact consistency tasks into binary labels, specifically considering only whether there is a factual conflict with the input text, without factoring in external knowledge. The research discovered that fact evaluation methods founded on natural language inference and question generation answering exhibit superior performance and can complement each other.
\citet{pezeshkpour2023measuring} proposed a novel metric, based on information theory, to assess the inclusion of specific knowledge in \llms. The metric utilized the concept of uncertainty in knowledge to measure factualness, calculated by \llms filling in prompts and examining the probability distribution of the answer. The paper discussed two methods for injecting knowledge into \llms: explicit inclusion of knowledge in the prompts and implicit fine-tuning of the \llms using knowledge-related data. The study demonstrated that this approach surpasses traditional ranking methods by achieving an accuracy improvement of over 30\%.
\citet{gekhman2023trueteacher} improved the method for evaluating fact consistency in summarization tasks. It proposed a novel approach that involved training student NLI models using summaries generated by multiple models and annotated by \llms to ensure fact consistency. The trained student model was then used for summarization fact consistency evaluation.
\citet{manakul2023selfcheckgpt} operated on two hypotheses regarding how \llms generate factual or hallucinated responses. It proposed the use of three formulas (BERTScore \citep{zhang2019bertscore}, MQAG \citep{manakul2023mqag} and n-gram) to evaluate factuality and employed alternative \llms to gather token probabilities for black-box language models. The study discovered that simply computing sentence likelihood or entropy helped validate the factuality of the responses.
\citet{min2023factscore} broke down text generated by \llms into individual ``atomic" facts, which were then evaluated for their correctness. The FActScore is used to measure the performance of estimators through the calculation of F1 scores. The paper tested various estimators and revealed that current estimators still have some way to go in effectively addressing the task.
\citet{lin2021truthfulqa} introduced the TruthfulQA dataset, designed to cause models to make mistakes. Multiple language models were tested by providing factual answers. The findings from these experiments suggest that simply scaling up model sizes may not necessarily improve their truthfulness, and recommendations are provided for the training approach. This dataset has become widely used for evaluating the factuality of \llms \citep{Kadavath2022LanguageM,touvron2023llama,openai2023gpt4,Wei2022EmergentAO}.

\subsection{Robustness, Ethic, Bias, and Trustworthiness}
\label{Robustness, Ethics, Biases, and Trustworthiness}

\input{tables/tb-robutness}

The evaluation encompasses crucial aspects of robustness, ethics, biases, and trustworthiness. 
These factors have gained increasing importance in assessing the performance of \llms comprehensively. 

\subsubsection{Robustness}
Robustness studies the stability of a system when facing unexpected inputs.
Specifically, out-of-distribution (OOD)~\citep{wang2022generalizing} and adversarial robustness are two popular research topics for robustness.
\citet{wang2023robustness} is an early work that evaluated ChatGPT and other \llms from both the adversarial and OOD perspectives using existing benchmarks such as AdvGLUE~\citep{wang2021adversarial}, ANLI~\citep{nie2019adversarial}, and DDXPlus~\citep{fansi2022ddxplus} datasets.
\citet{zhuo2023robustness} evaluated the robustness of semantic parsing.
\citet{yang2022glue} evaluated OOD robustness by extending the GLUE \citep{wang2018glue} dataset.
The results of this study emphasize the potential risks to the overall system security when manipulating visual input.
For vision-language models, \citet{zhao2023evaluating} evaluated \llms on visual input and transferred them to other visual-linguistic models, revealing the vulnerability of visual input.
\citet{li2023survey} provided an overview of OOD evaluation for language models: adversarial robustness, domain generalization, and dataset biases.
Bridging these lines of research, the authors conducted a comparative analysis, unifying the three approaches. They succinctly outlined the data-generation processes and evaluation protocols for each line of study, all while emphasizing the prevailing challenges and future research prospects.
Additionally, \citet{liu2023mitigating} introduced a large-scale robust visual instruction dataset to enhance the performance of large-scale multi-modal models in handling relevant images and human instructions.

For adversarial robustness, \citet{zhu2023promptbench} evaluated the robustness of \llms to prompts by proposing a unified benchmark called PromptBench.
They comprehensively evaluated adversarial text attacks at multiple levels (character, word, sentence, and semantics). The results showed that contemporary \llms are vulnerable to adversarial prompts, highlighting the importance of the models' robustness when facing adversarial inputs.
As for new adversarial datasets, \citet{wang2023decodingtrust} introduced AdvGLUE++ benchmark data for assessing adversarial robustness and implemented a new evaluation protocol to scrutinize machine ethics via jailbreaking system prompts.

\subsubsection{Ethic and bias}

LLMs have been found to internalize, spread, and potentially magnify harmful information existing in the crawled training corpora, usually, toxic languages, like offensiveness, hate speech, and insults~\citep{gehman2020realtoxicityprompts}, as well as social biases like stereotypes towards people with a particular demographic identity (\textit{e.g.}, gender, race, religion, occupation, and ideology)~\citep{sheng2021societal}. More recently, \citet{zhuo2023exploring} used conventional testing sets and metrics~\citep{gehman2020realtoxicityprompts,dhamala2021bold,parrish2022bbq} to perform a systematic evaluation of ChatGPT's toxicity and social bias, finding that it still exhibits noxious content to some extend. Taking a further step, \citet{deshpande2023toxicity} introduced role-playing into the model and observed an increase in generated toxicity up to 6x. Furthermore, such role-playing also caused biased toxicity towards specific entities. Different from simply measuring social biases, \citet{ferrara2023should} investigated the sources, underlying mechanisms, and corresponding ethical consequences of these biases potentially produced by ChatGPT. Beyond social biases, LLMs have also been assessed by political tendency and personality traits~\citep{rutinowski2023self,hartmann2023political} based questionnaires like the Political Compass Test and MBTI test, demonstrating a propensity for progressive views and an ENFJ personality type. In addition, LLMs like GPT-3 were found to have moral biases~\citep{simmons2022moral} in terms of the Moral Foundation theory~\citep{graham2013moral};
The study conducted by~\citep{hendrycks2020aligning}~reveals that existing LMs have potential in ethical judgment, but still need improvement. 
\citep{zhao2023chbias} proposes a Chinese conversational bias evaluation dataset CHBias, discovers bias risks in pre-trained models, and explores debiasing methods.
Moreover, in the assessment of GPT-4 alignment, ~\citep{wang2023large} discovered a systematic bias.
ChatGPT is also observed to exhibit somewhat bias on cultural values~\citep{cao2023assessing}.
\citet{wang2023decodingtrust} also incorporated an evaluation dataset specifically aimed at gauging stereotype bias, using both targeted and untargeted system prompts.
All these ethical issues might elicit serious risks, impeding the deployment of LLMs and having a profound negative impact on society.

\subsubsection{Trustworthiness}\label{sec:trustworthiness}

Some work focuses on other trustworthiness problems in addition to robustness and ethics.\footnote{The term `trustworthiness' in this section refers to other work that contains more than robustness and ethics.}
In their 2023 study, DecodingTrust, \citet{wang2023decodingtrust} offered a multifaceted exploration of trustworthiness vulnerabilities in the GPT models, especially GPT-3.5 and GPT-4.
Their evaluation expanded beyond the typical trustworthiness concerns to include eight critical aspects: toxicity, stereotype bias, adversarial and out-of-distribution robustness, robustness to adversarial demonstrations, privacy, machine ethics, and fairness. DecodingTrust's investigation employs an array of newly constructed scenarios, tasks, and metrics.
They revealed that while GPT-4 often showcases improved trustworthiness over GPT-3.5 in standard evaluations, it is simultaneously more susceptible to attacks.

In another study by \citet{hagendorff2023humanlike}, \llms with enhanced cognitive abilities were evaluated.
They found that these models can avoid common human intuitions and cognitive errors, demonstrating super-rational performance. By utilizing cognitive reflection tests and semantic illusion experiments, the researchers gained insights into the psychological aspects of \llms.
This method offers new perspectives for evaluating model biases and ethical issues that may not have been previously identified.
Furthermore, a study by \citep{xie2023ask} brings attention to a significant concern: the consistency of judgment in \llms diminishes notably when faced with disruptions such as questioning, negation, or misleading cues, even if their initial judgments were accurate. The research delves into various prompting methods designed to mitigate this issue and successfully demonstrates their efficacy.

\llms are capable of generating coherent and seemingly factual text. However, the information generated can include factual inaccuracies or statements ungrounded in reality, a phenomenon known as \textbf{hallucination} \citep{rawte2023survey,zhang2023sirens}. 
Evaluating these issues helps improve the training methods of \llms to reduce the occurrence of hallucinations. 
For the evaluation of illusions in large-scale visual models, 
\citet{liu2023mitigating} introduced a comprehensive and robust large-scale visual instruction dataset: LRV-Instruction. Through the GAVIE method, they fine-tuned the evaluation visual instructions, and experimental results demonstrated that LRV-Instruction effectively alleviates illusions in LLMs. 
In addition, 
\citet{li2023evaluating} conducted an assessment of illusions in large-scale visual language models, revealing through experiments that the distribution of objects in visual instructions significantly impacts object illusions in LVLMs. To enhance the assessment of object illusions in LVLMs, they introduced a polling-based query method, known as POPE. This method provides an improved evaluation of object illusions in LVLMs.

\subsection{Social Science}

Social science involves the study of human society and individual behavior, including economics, sociology, political science, law, and other disciplines.
Evaluating the performance of \llms in social science is important for academic research, policy formulation, and social problem-solving.
Such evaluations can help improve the applicability and quality of models in the social sciences, increasing understanding of human societies and promoting social progress.

\citet{wu2023large} evaluated the potential use of \llms in addressing scaling and measurement issues in social science and found that \llms can generate meaningful responses regarding political ideology and significantly improve text-as-data methods in social science.

In computational social science (CSS) tasks, \citet{ziems2023can} presented a comprehensive evaluation of \llms on several CSS tasks.
During classification tasks, \llms exhibit the lowest absolute performance on event argument extraction, character tropes, implicit hate, and empathy classification, achieving accuracy below 40\%. 
These tasks either involve complex structures (event arguments) or subjective expert taxonomies with semantics that differ from those learned during LLM pretraining. 
Conversely, \llms achieve the best performance on misinformation, stance, and emotion classification. 
When it comes to generation tasks, \llms often produce explanations that surpass the quality of gold references provided by crowd workers. 
In summary, while \llms can greatly enhance the traditional CSS research pipeline, they cannot completely replace it.

Some articles also evaluate \llms on legal tasks. 
The zero-shot performance of \llms is mediocre in legal case judgment summarization.
\llms have several problems, including incomplete sentences and words, meaningless sentences merge, and more serious errors such as inconsistent and hallucinated information \citep{deroy2023ready}.
The results showed that further improvement is necessary for \llms to be useful for case judgment summarization by legal experts.
\citet{nay2023large} indicated that \llms, particularly when combined with prompting enhancements and the correct legal texts, could perform better but not yet at expert tax lawyer levels. 

Lastly, within the realm of psychology,
\citet{frank2023baby} adopted an interdisciplinary approach and drew insights from developmental psychology and comparative psychology to explore alternative methods for evaluating the capabilities of \llms. By integrating different perspectives, researchers can deepen their understanding of the essence of cognition and effectively leverage the potential of advanced technologies such as large language models, while mitigating potential risks.

In conclusion, the utilization of \llms has significantly benefited individuals in addressing social science-related tasks, leading to improved work efficiency. The outputs produced by \llms serve as valuable resources for enhancing productivity. However, it is crucial to acknowledge that existing \llms cannot completely replace human professionals in this domain.

\subsection{Natural Science and Engineering}
Evaluating the performance of \llms in natural science and engineering can help guide applications and development in scientific research, technology development, and engineering studies.

\input{tables/tb-nature}

\subsubsection{Mathematics}

For fundamental mathematical problems, most large language models (\llms) demonstrate proficiency in addition and subtraction, and possess some capability in multiplication. However, they face challenges when it comes to division, exponentiation, trigonometry functions, and logarithm functions. On the other hand, \llms exhibit competence in handling decimal numbers, negative numbers, and irrational numbers \citep{yuan2023well}. 
In terms of performance, ChatGPT and GPT-4 outperform other models significantly, showcasing their superiority in solving mathematical tasks \citep{wei2023cmath}. 
These two models have a distinct advantage in dealing with large numbers (greater than 1e12) and complex, lengthy mathematical queries.
GPT-4 outperforms ChatGPT by achieving a significant increase in accuracy of 10 percentage points and a reduction in relative error by 50\%, due to its superior division and trigonometry abilities, proper understanding of irrational numbers, and consistent step-by-step calculation of long expressions.

When confronted with complex and challenging mathematical problems, \llms exhibit subpar performance. 
Specifically, GPT-3 demonstrates nearly random performance, while GPT-3.5 shows improvement, and GPT-4 performs the best \citep{arora2023have}. 
Despite the advancements made in the new models, it is important to note that the peak performance remains relatively low compared to that of experts and these models lack the capability to engage in mathematical research~\citep{bubeck2023sparks}.
The specific tasks of algebraic manipulation and calculation continue to pose challenges for GPTs \citep{collins2023evaluating, bubeck2023sparks}. 
The primary reasons behind GPT-4's low performance in these tasks are errors in algebraic manipulation and difficulties in retrieving pertinent domain-specific concepts.
\citet{wu2023empirical} evaluated the use of GPT-4 on difficult high school competition problems and GPT-4 reached 60\% accuracy on half of the categories. 
Intermediate algebra and precalculus can only be solved with a low accuracy rate of around 20\%.
ChatGPT is not good at answering questions on topics including derivatives and applications, Oxyz spatial calculus, and spatial geometry \citep{dao2023investigating}.
\citet{dao2023investigating, wei2023cmath} showed that ChatGPT's performance worsens as task difficulty increases: it correctly answered 83\% of the questions at the recognition level, 62\% at the comprehension level, 27\% at the application level, and only 10\% at the highest cognitive complexity level.
Given those problems at higher knowledge levels tend to be more complex, requiring in-depth understanding and problem-solving skills, such results are to be expected.

These results indicate that the effectiveness of \llms is highly influenced by the complexity of problems they encounter. 
This finding holds significant implications for the design and development of optimized artificial intelligence systems capable of successfully handling these challenging tasks.

\subsubsection{General science}
Further improvements are needed in the application of \llms in the field of chemistry.
\citet{castro2023large} presented five straightforward tasks from various subareas of chemistry to assess ChatGPT’s comprehension of the subject, with accuracy ranging from 25\% to 100\%.
\citet{guo2023indeed} created a comprehensive benchmark that encompasses 8 practical chemistry tasks, which is designed to assess the performance of \llms (including GPT-4, GPT-3.5, and Davinci-003) for each chemistry task. Based on the experiment results, GPT-4 demonstrates superior performance compared to the other two models.
\citep{arora2023have} showed that \llms perform worse on physics problems than chemistry problems, probably because chemistry problems have lower inference complexity than physics problems in this setting.
There are limited evaluation studies on \llms in the field of general science, and the current findings indicate that further improvement is needed in the performance of \llms within this domain.

\subsubsection{Engineering} 
Within engineering, the tasks can be organized in ascending order of difficulty, including code generation, software engineering, and commonsense planning.

In code generation tasks, the smaller \llms trained for the tasks are competitive in performance, and CodeGen-16B \citep{nijkamp2022codegen} is comparable in performance to ChatGPT using a larger parameter setting, reaching about a 78\% match \citep{liu2023your}.
Despite facing challenges in mastering and comprehending certain fundamental concepts in programming languages, ChatGPT showcases a commendable level of coding level \citep{zhuang2023efficiently}.
Specifically, ChatGPT has developed superior skills in dynamic programming, greedy algorithm, and search, surpassing highly capable college students, but it struggles in data structure, tree, and graph theory.
GPT-4 demonstrates an advanced ability to generate code based on given instructions, comprehend existing code, reason about code execution, simulate the impact of instructions, articulate outcomes in natural language, and execute pseudocode effectively \citep{bubeck2023sparks}.

In software engineering tasks, ChatGPT generally performs well and provides detailed responses, often surpassing both human expert output and SOTA output. However, for certain tasks such as code vulnerability detection and information retrieval-based test prioritization, the current version of ChatGPT fails to provide accurate answers, rendering it unsuitable for these specific tasks \citep{sridhara2023chatgpt}.

In commonsense planning tasks, \llms may not perform well, even in simple planning tasks where humans excel \citep{valmeekam2022large,valmeekam2023planning}. \citet{pallagani2023understanding} demonstrated that the fine-tuned CodeT5 \citep{wang2021codet5} performs the best across all considered domains, with the shortest inference time. Moreover, it explored the capability of \llms for plan generalization and found that their generalization capabilities appear to be limited. It turns out that \llms can handle simple engineering tasks, but they perform poorly on complex engineering tasks.

\subsection{Medical Applications}
The application of \llms in the medical field has recently received significant attention. 
As a result, this section aims to provide a comprehensive review of the ongoing efforts dedicated to implementing \llms in medical applications. 
We have categorized these applications into three aspects as shown in \tablename~\ref{table_3.2}: medical query, medical examination, and medical assistants. 
A detailed examination of these categories will enhance our understanding of the potential impact and advantages that \llms can bring to the medical domain.

\subsubsection{Medical queries}


The significance of evaluating \llms on medical queries lies in providing accurate and reliable medical answers to meet the needs of healthcare professionals and patients for high-quality medical information. As shown in \tablename~\ref{table_3.2}, the majority of \llms evaluations in the medical field concentrate on medical queries.
ChatGPT generated relatively accurate information for various medical queries, including genetics~\citep{duong2023analysis}, radiation oncology physics~\citep{holmes2023evaluating}, biomedicine~\citep{jahan2023evaluation}, and many other medical disciplines~\citep{samaan2023assessing,johnson2023assessing,hamidi2023evaluation}, demonstrating its effectiveness in the field of medical queries to a certain extent.
As for the limitations, \citet{thirunavukarasu2023trialling} assessed ChatGPT's performance in primary care and found that its average score in the student comprehensive assessment falls below the passing score, indicating room for improvement. 
\citet{chervenak2023promise} highlighted that while ChatGPT can generate responses similar to existing sources in fertility-related clinical prompts, its limitations in reliably citing sources and potential for fabricating information restrict its clinical utility. 
\input{tables/tb-medical}
\subsubsection{Medical examination}
The studies by~\citet{gilson2023does,kung2023performance} have evaluated the performance of \llms in medical examination assessment through the United States Medical Licensing Examination (USMLE)~\footnote{\url{https://www.usmle.org/}}.
In the study of \citep{gilson2023does}, ChatGPT's performance in answering USMLE Step 1 and Step 2 exam questions was assessed using novel multiple-choice question sets. The results indicated that ChatGPT achieves varying accuracies across different datasets. However, the presence of out-of-context information was found to be lower compared to the correct answer in the NBME-Free-Step1 and NBME-Free-Step2 datasets.
\citet{kung2023performance} showed that ChatGPT achieves or approaches the passing threshold in these exams with no tailored training.
The model demonstrates high consistency and insight, indicating its potential to assist in medical education and clinical decision-making. ChatGPT can be used as a tool to answer medical questions, provide explanations, and support decision-making processes. This offers additional resources and support for medical students and clinicians in their educational and clinical practices.
Moreover, \citet{sharma2023performance} found that answers generated by ChatGPT are more context-aware with better deductive reasoning abilities compared to Google search results.

\subsubsection{Medical assistants}

In the field of medical assistance, \llms demonstrate potential applications, including research on identifying gastrointestinal diseases \citep{lahat2023evaluating}, dementia diagnosis \citep{wang2023can}, accelerating the evaluation of COVID-19 literature \citep{khan2023covllm}, and their overall potential in healthcare~\citep{cascella2023evaluating}. However, there are also limitations and challenges, such as lack of originality, high input requirements, resource constraints, uncertainty in answers, and potential risks related to misdiagnosis and patient privacy issues.

Moreover, several studies have evaluated the performance and feasibility of ChatGPT in the medical education field.
In the study by \citet{oh2023chatgpt}, ChatGPT, specifically GPT-3.5 and GPT-4 models, were evaluated in terms of their understanding of surgical clinical information and their potential impact on surgical education and training. The results indicate an overall accuracy of 46.8\% for GPT-3.5 and 76.4\% for GPT-4, demonstrating a significant performance difference between the two models. Notably, GPT-4 consistently performs well across different subspecialties, suggesting its capability to comprehend complex clinical information and enhance surgical education and training.
Another study by \citet{lyu2023translating} explores the feasibility of utilizing ChatGPT in clinical education, particularly in translating radiology reports into easily understandable language.
The findings demonstrate that ChatGPT effectively translates radiology reports into accessible language and provides general recommendations.
Furthermore, the quality of ChatGPT has shown improvement compared to GPT-4.
These findings suggest that employing \llms in clinical education is feasible, although further efforts are needed to address limitations and unlock their full potential.

\subsection{Agent Applications}

Instead of focusing solely on general language tasks, \llms can be utilized as powerful tools in various domains. Equipping LLMs with external tools can greatly expand the capabilities of the model~\citep{qin2023tool}.
ToolLLM~\citep{qin2023toolllm} provides a comprehensive framework to equip open-source large language models with tool use capabilities.
\citet{huang2023language} introduced KOSMOS-1, which is capable of understanding general patterns, following instructions, and learning based on context.
The study by MRKL~\citet{karpas2022mrkl} emphasized the importance of understanding when and how to utilize external symbolic tools, as this knowledge is dependent on the capabilities of \llms, particularly when these tools can reliably perform functions.
Additionally, two other studies, Toolformer~\citep{schick2023toolformer} and TALM~\citep{parisi2022talm}, explored the utilization of tools to enhance language models. Toolformer employs a training approach to determine the optimal usage of specific APIs and integrates the obtained results into subsequent token predictions. On the other hand, TALM combines indistinguishable tools with text-based methods to augment language models and employs an iterative technique known as ``self-play", guided by minimal tool demonstrations.
Furthermore, \citet{shen2023hugginggpt} proposed the HuggingGPT framework, which leverages \llms to connect various AI models within the machine learning community (such as Hugging Face), aiming to address AI tasks.


\subsection{Other Applications}

In addition to above areas, there have been evaluations in various other domains, including education, search and recommendation, personality testing, and specific applications.

\subsubsection{Education} 
\llms have shown promise in revolutionizing the field of education. They have the potential to make significant contributions in several areas, such as assisting students in improving their writing skills, facilitating better comprehension of complex concepts, expediting the delivery of information, and providing personalized feedback to enhance student engagement. These applications aim to create more efficient and interactive learning experiences, offering students a broader range of educational opportunities. However, to fully harness the potential of \llms in education, extensive research, and ongoing refinement are necessary.


The evaluation of \llms for \textbf{educational assistance} aims to investigate and assess their potential contributions to the field of education. Such evaluations can be conducted from various perspectives.
According to \citet{dai2023can}, ChatGPT demonstrates the ability to generate detailed, fluent, and coherent feedback that surpasses that of human teachers. It can accurately assess student assignments and provide feedback on task completion, thereby assisting in the development of student skills.
However, ChatGPT's responses may lack novelty or insightful perspectives regarding teaching improvement \citep{wang2023chatgpt}.
Additionally, the study conducted by \citet{hellas2023exploring} revealed that \llms can successfully identify at least one actual problem in student code, although instances of misjudgment are also observed.
In conclusion, the utilization of \llms shows promise in addressing program logic issues, although challenges remain in achieving proficiency in output formatting. It is important to note that while these models can provide valuable insights, they may still generate errors similar to those made by students.



In \textbf{educational exams}, researchers aim to evaluate the application effectiveness of \llms, including automatic scoring, question generation, and learning guidance.
\citet{de2023can} showed that ChatGPT achieves an average of 71.8\% correctness, which is comparable to the average score of all participating students.
Subsequently, the evaluation was conducted using GPT-4, and it achieved a score of 8.33. Furthermore, this evaluation showed the effectiveness of leveraging bootstrapping that combines randomness via the ``temperature'' parameter in diagnosing incorrect answers.
\citet{zhang2023exploring} claimed that GPT-3.5 can solve MIT math and EECS exams with GPT-4 achieving better performance.
However, it turned out to be not fair since they accidentally included the correct answers into the prompts.

\input{tables/tb-others}

\subsubsection{Search and recommendation}
The assessment of \llms in search and recommendation can be broadly categorized into two areas.
Firstly, in the realm of \textbf{information retrieval}, \citet{sun2023chatgpt} investigated the effectiveness of generative ranking algorithms, such as ChatGPT and GPT-4, for information retrieval tasks. Experimental results demonstrate that guided ChatGPT and GPT-4 exhibit competitive performance on popular benchmark tests, even outperforming supervised methods. Additionally, the extraction of ChatGPT's ranking functionality into a specialized model shows superior performance when trained on 10K ChatGPT-generated data compared to training on 400K annotated MS MARCO data in the BEIR dataset \citep{thakur2021beir}.
Furthermore, \citet{xu2023chatgpt} conducted a randomized online experiment to investigate the behavioral differences of users when performing information retrieval tasks using search engines and chatbot tools.
Participants were divided into two groups: one using tools similar to ChatGPT and the other using tools similar to Google Search. The results show that the ChatGPT group spent less time on all tasks and the difference between these two groups is not significant.

Secondly, moving to the domain of \textbf{recommendation systems}, 
\llms have emerged as essential components that leverage their natural language processing capabilities to comprehend user preferences, item descriptions, and contextual information \citep{fan2023recommender}. By incorporating LLMs into recommendation pipelines, these systems can offer more accurate and personalized recommendations, thereby improving user experience and overall recommendation quality.
However, it is crucial to address the potential risks associated with using LLMs for recommendations. Recent research by \citet{zhang2023chatgpt} has highlighted the issue of unfair recommendations generated by ChatGPT. This emphasizes the importance of evaluating fairness when employing LLMs in recommendation scenarios.
\citet{dai2023uncovering} suggest that ChatGPT exhibits strong performance in recommender systems. The use of listwise ranking is found to strike the best balance between cost and performance. Furthermore, ChatGPT shows promise in addressing the cold-start problem and providing interpretable recommendations. 
Moreover, the research by \citet{yuan2023recommender} and \citet{li2023exploring} demonstrated the promising potential of the modality-based recommendation model (MoRec) and text-based collaborative filtering (TCF) in recommendation systems.

\subsubsection{Personality testing} 
Personality testing aims to measure individuals' personality traits and behavioral tendencies, and \llms as powerful natural language processing models have been widely applied in such tasks. 

Research conducted by \citet{bodroza2023personality} investigated the personality features of using Davinci-003 as a chatbot and found variations in the consistency of its answers, despite exhibiting prosocial characteristics.
However, there remains uncertainty regarding whether the chatbot's responses are driven by conscious self-reflection or algorithmic processes.
\citet{song2023have} examined the manifestation of personality in language models and discovered that many models perform unreliably in self-assessment tests and exhibit inherent biases.
Therefore, it is necessary to develop specific machine personality measurement tools to enhance reliability.
These studies offer vital insights to better understand \llms in personality testing.
\citet{safdari2023personality} proposed a comprehensive approach to conduct effective psychometric testing for the personality traits in the text generated by \llms.
In order to evaluate the emotional intelligence of \llms, \citet{wang2023emotional} developed a new psychometric assessment method. By referencing a framework constructed from over 500 adults, the authors tested various mainstream \llms. The results showed that most \llms achieve above-average scores in emotional quotient (EQ), with GPT-4 scoring 117, surpassing 89\% of human participants. However, a multivariate pattern analysis indicated that certain \llms achieve human-level performance without relying on mechanisms resembling those found in humans. This is evident from the distinct differences in the quality of their representational patterns, as compared to humans.
\citet{liang2023leveraging} employed the word guessing game to evaluate \llms' language and theory of mind intelligences, a more engaging and interactive assessment method.
\citet{jentzsch2023chatgpt} discussed the challenges of incorporating humor into \llms, particularly ChatGPT.
They found that while ChatGPT demonstrates impressive capabilities in NLP tasks, it falls short in generating humorous responses.
This study emphasizes the importance of humor in human communication and the difficulties that \llms face in capturing the subtleties and context-dependent nature of humor.
It discusses the limitations of current approaches and highlights the need for further research on more sophisticated models that can effectively understand and generate humor. 

\subsubsection{Specific applications}
Moreover, various research endeavors have been conducted to explore the application and evaluation of \llms across a wide spectrum of tasks, such as \textbf{game design} \citep{lanzi2023chatgpt}, \textbf{model performance assessment} \citep{wang2023pandalm}, and \textbf{log parsing} \citep{le2023evaluation}.
Collectively, these findings enhance our comprehension of the practical implications associated with the utilization of \llms across diverse tasks. They shed light on the potential and limitations of these models while providing valuable insights for performance improvement.

%% file: tables/tb-nlptask.tex
\begin{table*}
\centering
\caption{Summary of evaluation on \textbf{natural language processing} tasks: NLU (Natural Language Understanding, including SA (Sentiment Analysis), TC (Text Classification), NLI (Natural Language Inference) and other NLU tasks), Reasoning, NLG (Natural Language Generation, including Summ. (Summarization), Dlg. (Dialogue), Tran (Translation), QA (Question Answering) and other NLG tasks), and Multilingual tasks (ordered by the name of the first author).}
\label{tb-what-nlptask}
\centering
\resizebox{\textwidth}{!}{
\begin{tabular}{ |l| c| c| c |c| c| c |c |c| c |c |c|}
\hline
\multicolumn{1}{|l|}{\multirow{2}{*}{\textbf{Reference}}} & \multicolumn{4}{c|}{\textbf{NLU}}                                                     & \multicolumn{1}{c|}{\multirow{2}{*}{\textbf{RNG.}}} & \multicolumn{5}{c|}{\textbf{NLG}}                                                                                      & \multicolumn{1}{c|}{\multirow{2}{*}{\textbf{Mul.}}} \\ \cline{2-5} \cline{7-11}
\multicolumn{1}{|c|}{}                                    & \multicolumn{1}{l|}{SA} & \multicolumn{1}{l|}{TC} & \multicolumn{1}{l|}{NLI} & Others & \multicolumn{1}{c|}{}                               & \multicolumn{1}{l|}{Summ.} & \multicolumn{1}{l|}{Dlg.} & \multicolumn{1}{l|}{Tran.} & \multicolumn{1}{l|}{QA} & Others & \multicolumn{1}{c|}{}                               \\ 
\hline \hline
\citet{abdelali2023benchmarking} & & & & & & & & & & &\checkmark\\\hline
\citet{ahuja2023mega} & & & & & & & & & & &\checkmark\\\hline
\citet{bian2023chatgpt}  & & & & &\checkmark & & & &\checkmark & & \\ \hline
\citet{bang2023multitask} &\checkmark & & & &\checkmark &\checkmark &\checkmark &\checkmark &\checkmark & &\checkmark\\\hline
\citet{bai2023benchmarking} & & & & & & & & &\checkmark & &\\\hline
\citet{chen2023exploring} & & & & & & & & & &\checkmark &  \\ \hline
\citet{choi2023llms} & & & &\checkmark & & & & & & & \\\hline
\citet{chia2023instructeval} & & & & & & & & & &\checkmark &  \\\hline
\citet{frieder2023mathematical} & & & & &\checkmark & & & & & & \\\hline
\citet{fu2023chain}  & & & & &\checkmark & & & & & &\\\hline
\citet{gekhman2023trueteacher} & & & & & &\checkmark & & & & &\\\hline
\citet{gendron2023large} & & & & &\checkmark & & & & & &\\\hline
\citet{honovich2022true} & & &\checkmark & & &\checkmark &\checkmark & & &\checkmark &\\\hline
\citet{jiang2023structgpt} & & & & &\checkmark & & & & & &\\\hline
\citet{lai2023chatgpt} & & & & & & & & & & &\checkmark\\ \hline
\citet{laskar2023systematic} &\checkmark & & \checkmark& &\checkmark &\checkmark & &\checkmark &\checkmark &\checkmark &\checkmark \\\hline
\citet{lopez2023can} &\checkmark & & & & & & & & & & \\ \hline
\citet{liang2022holistic} &\checkmark &\checkmark & & & &\checkmark & & &\checkmark & & \\\hline
\citet{lee2023can} & & &\checkmark & & & & & & & &\\\hline
\citet{lin2023llm} & & & & & & &\checkmark & & & &\\\hline
\citet{lievin2022can} & & & & &\checkmark & & & & & &\\\hline
\citet{liu2023evaluating} & & & & &\checkmark & & & & & &\\\hline
\citet{lyu2023new} & & & & & & & & & \checkmark & &\\\hline
\citet{manakul2023selfcheckgpt} & & & & & & & & &\checkmark &\checkmark &\\\hline
\citet{min2023factscore} & & & & & & & & & &\checkmark &\\\hline
\citet{orru2023human} & & & & &\checkmark & & & & & &\\\hline
\citet{pan2023unifying} & & & & &\checkmark & & & & & &\\\hline
\citet{pena2023leveraging} & &\checkmark & & & & & & & & & \\\hline
\citet{pu2023chatgpt} & & & & & &\checkmark & & & &\checkmark & \\\hline
\citet{pezeshkpour2023measuring} & & & & & & & & & &\checkmark &\\\hline
\citet{qin2023chatgpt}  &\checkmark & &\checkmark & &\checkmark &\checkmark &\checkmark & &\checkmark & &  \\ \hline
\citet{riccardi2023two} & & & &\checkmark & & & & & & & \\\hline
\citet{saparov2023testing} & & & & &\checkmark & & & & & &\\\hline
\citet{tao2023eveval} & & & &\checkmark & & & & & & & \\\hline
\citet{wang2023document} & & & & & & & &\checkmark & & & \\ \hline
\citet{wang2023chatgpt1} &\checkmark & & & & & & & & & &\\\hline
\citet{wang2023evaluating} & & &\checkmark & & & & & & \checkmark & &\\\hline
\citet{wu2023reasoning} & & & & &\checkmark & & & & & & \\\hline
\citet{wu2022autoformalization} & & & & &\checkmark & & & & & & \\\hline
\citet{xu2023large} & & & & &\checkmark & & & & & &\\\hline
\citet{yang2023large}  & &\checkmark & & & & & & & & & \\ \hline
\citet{zheng2023lmsys}& & & & & & & \checkmark& & & & \\\hline
\citet{zhang2023sentiment} &\checkmark & & & & & & & & & & \\\hline
\citet{zhang2023m3exam} & & & & & & & & & & &\checkmark \\\hline
\citet{zhuang2023efficiently} & & & & &\checkmark & & & & & & \\\hline
\citet{zhang2023evaluating1} & & & & &\checkmark & & & & & & \\\hline

\end{tabular}}
\end{table*}

%% file: tables/tb-robutness.tex
\begin{table*}
\centering
\caption{Summary of \llms evaluation on \textbf{robustness, ethics, biases, and trustworthiness} (ordered by the name of the first author).}
\label{table_3.3}
\centering
\resizebox{.7\textwidth}{!}{
\begin{tabular}{| l |c| c| c| }
\hline
\textbf{Reference} & \textbf{Robustness} & \textbf{Ethics and biases} & \textbf{Trustworthiness} \\
\hline\hline

\citet{cao2023assessing} & &\checkmark & \\\hline
\citet{dhamala2021bold} & &\checkmark & \\\hline
\citet{deshpande2023toxicity} & &\checkmark & \\\hline
\citet{ferrara2023should} & &\checkmark & \\\hline
\citet{gehman2020realtoxicityprompts} & &\checkmark & \\\hline
\citet{hartmann2023political} & &\checkmark & \\\hline
\citet{hendrycks2020aligning} & &\checkmark & \\\hline
\citet{hagendorff2023humanlike} & & &\checkmark \\\hline
\citet{li2023survey} &\checkmark & & \\\hline
\citet{liu2023mitigating} &\checkmark & & \\\hline
\citet{liu2023mitigating}& & & \checkmark\\
\hline
\citet{li2023evaluating}& & &\checkmark \\
\hline
\citet{parrish2022bbq} & &\checkmark & \\\hline
\citet{rutinowski2023self} & &\checkmark & \\\hline
\citet{rawte2023survey}& & &\checkmark \\
\hline
\citet{sheng2021societal} & &\checkmark & \\\hline
\citet{simmons2022moral} & &\checkmark & \\\hline
\citet{wang2022generalizing} &\checkmark & & \\\hline
\citet{wang2023robustness} &\checkmark & & \\\hline
\citet{wang2023decodingtrust} & \checkmark & \checkmark &\checkmark \\\hline
\citet{wang2023large} & &\checkmark & \\\hline
\citet{xie2023ask} & & &\checkmark \\\hline
\citet{yang2022glue} &\checkmark & & \\\hline
\citet{zhao2023evaluating} &\checkmark & & \\  \hline
\citet{zhuo2023robustness} &\checkmark & & \\\hline
\citet{zhu2023promptbench} &\checkmark & & \\\hline
\citet{zhuo2023exploring} & &\checkmark & \\\hline
\citet{zhang2023sirens}& & & \checkmark\\\hline

\end{tabular}}
\end{table*}

%% file: tables/tb-nature.tex
\begin{table}
\centering
\caption{Summary of evaluations on \textbf{natural science and engineering tasks} based on three aspects: Mathematics, General science and Engineering (ordered by the name of the first author).}
\label{table_3.4}
\centering
\resizebox{.76\textwidth}{!}{
\begin{tabular}{ |l| c |c| c| }
\hline
\textbf{Reference} & \textbf{Mathematics} & \textbf{General science} & \textbf{Engineering} \\
\hline\hline
\citet{arora2023have} &\checkmark &\checkmark & \\ \hline 
\citet{bubeck2023sparks} &\checkmark & &\checkmark \\\hline
\citet{castro2023large} & &\checkmark & \\  \hline
\citet{collins2023evaluating} &\checkmark & & \\ \hline 
\citet{dao2023investigating} &\checkmark & &\\ \hline 
\citet{guo2023indeed} & &\checkmark &\\\hline
\citet{liu2023your} & & &\checkmark \\ \hline 
\citet{pallagani2023understanding} & & &\checkmark \\\hline  
\citet{sridhara2023chatgpt} & & &\checkmark \\\hline  
\citet{valmeekam2023planning} & & &\checkmark \\\hline  
\citet{valmeekam2022large} & & &\checkmark \\ \hline 
\citet{wei2023cmath} &\checkmark & & \\\hline
\citet{wu2023empirical} &\checkmark & &\\ \hline 
\citet{yuan2023well} &\checkmark & & \\ \hline 
\citet{yu2023metamath} &\checkmark & & \\\hline
\citet{zhuang2023efficiently} & & &\checkmark \\\hline

\end{tabular}}
\end{table}

%% file: tables/tb-medical.tex
\begin{table}[t!]
\centering
\caption{Summary of evaluations on \textbf{medical applications} based on the three aspects: Medical queries, Medical assistants, and Medical examination (ordered by the name of the first author).}
\label{table_3.2}
\centering
\resizebox{.82\textwidth}{!}{
\begin{tabular}{ |l |c |c| c |}
\hline
\textbf{Reference} & \textbf{Medical queries} & \textbf{Medical examination} & \textbf{Medical assistants} \\
\hline\hline
\citet{cascella2023evaluating} & & &\checkmark   \\\hline
\citet{chervenak2023promise} &\checkmark & &  \\  \hline
\citet{duong2023analysis} &\checkmark & &  \\ \hline
\citet{gilson2023does} & &\checkmark&   \\ \hline
\citet{hamidi2023evaluation} &\checkmark & &  \\ \hline
\citet{holmes2023evaluating} &\checkmark & &  \\ \hline
\citet{jahan2023evaluation} &\checkmark & &  \\\hline
\citet{johnson2023assessing} &\checkmark & &  \\ \hline
\citet{khan2023covllm} & & &\checkmark  \\ \hline
\citet{kung2023performance} & &\checkmark&   \\ \hline
\citet{lahat2023evaluating} & & &\checkmark  \\ \hline
\citet{lyu2023translating} & & &\checkmark  \\ \hline
\citet{oh2023chatgpt} & & &\checkmark  \\ \hline
\citet{samaan2023assessing} &\checkmark & &  \\  \hline
\citet{thirunavukarasu2023trialling} &\checkmark & &  \\ \hline
\citet{wang2023can} & & &\checkmark  \\ \hline

\end{tabular}}
\end{table}

%% file: tables/tb-others.tex
\begin{table}[t!]
\centering
\caption{Summary of evaluations on \textbf{other applications} based on the four aspects: Education, Search and recommendation, Personality testing and Specific applications (ordered by the name of the first author).}
\label{table_3.4}
\centering
\resizebox{.98\textwidth}{!}{
\begin{tabular}{ |l |c |c |c |c| }
\hline
\textbf{Reference} & \textbf{Education} & \textbf{Search and recommendation} & \textbf{Personality testing} & \textbf{Specific applications}\\
\hline\hline
\citet{bodroza2023personality} & & &\checkmark &\\ \hline
\citet{dai2023can} &\checkmark & && \\\hline
\citet{de2023can} &\checkmark & & &\\\hline
\citet{dai2023uncovering} & &\checkmark & &\\ \hline
\citet{fan2023recommender} & &\checkmark & &\\\hline
\citet{hellas2023exploring} &\checkmark & & & \\ \hline
\citet{jentzsch2023chatgpt} & & &\checkmark &\\ \hline
\citet{lanzi2023chatgpt} & & & & \checkmark \\ \hline
\citet{le2023evaluation} & & & & \checkmark \\ \hline
\citet{li2023exploring} & &\checkmark & &\\ \hline
\citet{liang2023leveraging} & & &\checkmark &\\ \hline
\citet{sun2023chatgpt} & &\checkmark & &\\ \hline
\citet{song2023have} & & &\checkmark &\\ \hline
\citet{safdari2023personality} & & &\checkmark &\\\hline
\citet{thakur2021beir}& &\checkmark & &\\ \hline
\citet{wang2023chatgpt} &\checkmark & & & \\ \hline
\citet{wang2023emotional} & & &\checkmark & \\\hline
\citet{wang2023pandalm} & & & & \checkmark\\ \hline
\citet{xu2023chatgpt} & &\checkmark & &\\ \hline
\citet{yuan2023recommender} & &\checkmark & &\\ \hline
\citet{zhang2023chatgpt} & &\checkmark & &\\ \hline

\end{tabular}}
\end{table}

%% file: sec-benchmark-dataset.tex
\input{tables/tb-benchmarks}

\llms evaluation datasets are used to test and compare the performance of different language models on various tasks, as depicted in Sec.~\ref{sec-what}.
These datasets, such as GLUE~\citep{wang2018glue} and SuperGLUE~\citep{wang2019superglue}, aim to simulate real-world language processing scenarios and cover diverse tasks such as text classification, machine translation, reading comprehension, and dialogue generation.
This section will not discuss any single dataset for language models but benchmarks for \llms.

A variety of benchmarks have emerged to evaluate their performance. 
In this study, we compile a selection of 46 popular benchmarks, as shown in \tablename~\ref{tb-benchmarks}.\footnote{Note that as the evaluation of \llms is a hot research area, it is very likely that we cannot cover all benchmarks. We welcome suggestions and comments to make this list perfect.}
Each benchmark focuses on different aspects and evaluation criteria, providing valuable contributions to their respective domains.
For a better summarization, we divide these benchmarks into three categories: benchmarks for general language tasks, benchmarks for specific downstream tasks, and benchmarks for multi-modal tasks.










\subsection{Benchmarks for General Tasks}
\llms are designed to solve a vast majority of tasks.
To this end, existing benchmarks tend to evaluate the performance in different tasks.

Chatbot Arena~\citep{chatbotarena} and MT-Bench~\citep{zheng2023judging} are two significant benchmarks that contribute to the evaluation and advancement of chatbot models and \llms in different contexts.
Chatbot Arena provides a platform to assess and compare diverse chatbot models through user engagement and voting.
Users can engage with anonymous models and express their preferences via voting.
The platform gathers a significant volume of votes, facilitating the evaluation of models' performance in realistic scenarios.
Chatbot Arena provides valuable insights into the strengths and limitations of chatbot models, thereby contributing to the progress of chatbot research and advancement.

Meanwhile, MT-Bench evaluates \llms on multi-turn dialogues using comprehensive questions tailored to handling conversations.
It provides a comprehensive set of questions specifically designed for assessing the capabilities of models in handling multi-turn dialogues.
MT-Bench possesses several distinguishing features that differentiate it from conventional evaluation methodologies.
Notably, it excels in simulating dialogue scenarios representative of real-world settings, thereby facilitating a more precise evaluation of a model's practical performance.
Moreover, MT-Bench effectively overcomes the limitations in traditional evaluation approaches, particularly in gauging a model's competence in handling intricate multi-turn dialogue inquiries.

Instead of focusing on specific tasks and evaluation metrics, HELM \citep{liang2022holistic} provides a comprehensive assessment of \llms. It evaluates language models across various aspects such as language understanding, generation, coherence, context sensitivity, common-sense reasoning, and domain-specific knowledge. HELM aims to holistically evaluate the performance of language models across different tasks and domains.
For \llms Evaluator, \citet{zhang2023wider} introduces LLMEval², which encompasses a wide range of capability evaluations.
In addition, Xiezhi~\citep{gu2023xiezhi} presents a comprehensive suite for assessing the knowledge level of large-scale language models in different subject areas.
The evaluation conducted through Xiezhi enables researchers to comprehend the notable limitations inherent in these models and facilitates a deeper comprehension of their capabilities in diverse fields.
For evaluating language models beyond their existing capacities, BIG-bench \citep{srivastava2022beyond} introduces a diverse collection of 204 challenging tasks contributed by 450 authors from 132 institutions.
These tasks cover various domains such as math, childhood development, linguistics, biology, common-sense reasoning, social bias, physics, software development, etc.


Recent work has led to the development of benchmarks for evaluating language models' knowledge and reasoning abilities. The Knowledge-Oriented Language Model Evaluation KoLA \citep{yu2023kola} focuses on assessing language models' comprehension and utilization of semantic knowledge for inference. As such, KoLA serves as an important benchmark for evaluating the depth of language understanding and reasoning in language models, thereby driving progress in language comprehension. 
To enable crowd-sourced evaluations of language tasks, DynaBench~\citep{kiela2021dynabench} supports dynamic benchmark testing. DynaBench explores new research directions including the effects of closed-loop integration, distributional shift characteristics, annotator efficiency, influence of expert annotators, and model robustness to adversarial attacks in interactive settings. 
Furthermore, to evaluate language models' ability to learn and apply multidisciplinary knowledge across educational levels, the Multidisciplinary Knowledge Evaluation M3KE ~\citep{liu2023m3ke} was recently introduced. M3KE assesses knowledge application within the Chinese education system.


The development of standardized benchmarks for evaluating \llms on diverse tasks has been an important research focus. MMLU~\citep{hendrycks2020measuring} provides a comprehensive suite of tests for assessing text models in multi-task contexts.
AlpacaEval~\citep{alpaca_eval} stands as an automated evaluation benchmark, which places its focus on assessing the performance of \llms across various natural language processing tasks. It provides a range of metrics, robustness measures, and diversity evaluations to gauge the capabilities of LLMs. AlpacaEval has significantly contributed to advancing LLMs in diverse domains and promoting a deeper understanding of their performance.
Furthermore, AGIEval \citep{zhong2023agieval}, serves as a dedicated evaluation framework for assessing the performance of foundation models in the domain of human-centric standardized exams. 
Moreover, OpenLLM~\citep{leaderboard} functions as an evaluation benchmark by offering a public competition platform for comparing and assessing different LLM models' performance on various tasks. It encourages researchers to submit their models and compete on different tasks, driving progress and competition in LLM research.

As for tasks beyond standard performance, there are benchmarks designed for OOD, adversarial robustness, and fine-tuning.
GLUE-X \citep{yang2022glue} is a novel attempt to create a unified benchmark aimed at evaluating the robustness of NLP models in OOD scenarios. This benchmark emphasizes the significance of robustness in NLP and provides insights into measuring and enhancing the robustness of models.
In addition, \citet{yuan2023revisiting} presents BOSS, a benchmark collection for assessing out-of-distribution robustness in natural language processing tasks.
PromptBench \citep{zhu2023promptbench} centers on the importance of prompt engineering in fine-tuning \llms. It provides a standardized evaluation framework to compare different prompt engineering techniques and assess their impact on model performance. PromptBench facilitates the enhancement and optimization of fine-tuning methods for \llms.
To ensure impartial and equitable evaluation, PandaLM \citep{wang2023pandalm} is introduced as a discriminative large-scale language model specifically designed to differentiate among multiple high-proficiency LLMs through training. In contrast to conventional evaluation datasets that predominantly emphasize objective correctness, PandaLM incorporates crucial subjective elements, including relative conciseness, clarity, adherence to instructions, comprehensiveness, and formality.
 
\subsection{Benchmarks for Specific Downstream Tasks}
Other than benchmarks for general tasks, there exist benchmarks specifically designed for certain downstream tasks.

Question-answering benchmarks have become a fundamental component in the assessment of \llms and their overall performance. 
MultiMedQA \citep{singhal2022large} is a medical QA benchmark that focuses on medical examinations, medical research, and consumer healthcare questions. It consists of seven datasets related to medical QA, including six existing datasets and one new dataset. The goal of this benchmark is to evaluate the performance of \llms in terms of clinical knowledge and QA abilities.
To assess the ability of LLMs in dynamic QA about current world knowledge, \citet{vu2023freshllms} introduced FRESHQA. By incorporating relevant and current information retrieved from search engines into prompts, there is a significant enhancement in the performance of LLMs on FRESHQA.
To effectively assess in-depth dialogue, \citet{wang2023chainofthought} introduced the Dialogue CoT, incorporating two efficient dialogue strategies: Explicit CoT and CoT.

The assessment of \llms in diverse and demanding tasks has garnered substantial attention in recent research. To this end, a range of specialized benchmarks have been introduced to evaluate \llms' capabilities in specific domains and applications. Among these, ARB, as presented by \citet{sawada2023arb}, focuses on probing the performance of \llms in advanced reasoning tasks spanning multiple domains.
Additionally, ethical considerations in \llms have become an area of paramount importance. TRUSTGPT, as tailored by \citet{huang2023trustgpt}, addresses critical ethical dimensions, including toxicity, bias, and value alignment, within the context of \llms.
Furthermore, the simulation of human emotional reactions by \llms remains an area with significant potential for improvement, as highlighted by the EmotionBench benchmark by \citet{huang2023emotionally}.
In terms of security evaluation, \citet{zhang2023safetybench} have introduced SafetyBench, a benchmark specifically designed to test the security performance of a range of popular Chinese and English \llms. The results of this evaluation reveal substantial security flaws in current LLMs.
To evaluate the daily decision-making capabilities of intelligent systems, \citet{hou2023choice} introduced Choice-75.
Additionally, to assess \llms' aptitude in understanding complex instructions, \citet{he2023can} have introduced CELLO. This benchmark encompasses the design of eight distinctive features, the development of a comprehensive evaluation dataset, and the establishment of four evaluation criteria alongside their respective measurement standards.

Other specific benchmarks such as C-Eval \citep{huang2023c}, which is the first extensive benchmark to assess the advanced knowledge and reasoning capabilities of foundation models in Chinese.
Additionally, \citet{li2023cmmlu} introduces CMMLU as a comprehensive Chinese proficiency standard and evaluates the performance of 18 LLMs across various academic disciplines. The findings reveal that the majority of LLMs demonstrate suboptimal performance in Chinese language environments, highlighting areas for improvement.
M3Exam \citep{zhang2023m3exam} provides a unique and comprehensive evaluation framework that incorporates multiple languages, modalities, and levels to test the general capabilities of \llms in diverse contexts.
Additionally, GAOKAO-Bench~\citep{zhang2023evaluating} provides a comprehensive evaluation benchmark for gauging the proficiency of large language models in intricate and context-specific tasks, utilizing questions sourced from the Chinese Gaokao examination.
On the other hand, SOCKET \citep{choi2023llms} serves as an NLP benchmark designed to evaluate the performance of \llms in learning and recognizing social knowledge concepts.
It consists of several tasks and case studies to assess the limitations of \llms in social capabilities.
MATH~\citep{hendrycks2021measuring-1} concentrates on assessing reasoning and problem-solving proficiencies of AI models within the domain of mathematics.
APPS~\citep{hendrycks2021measuring-2} is a more comprehensive and rigorous benchmark for evaluating code generation, measuring the ability of language models to generate python code according to natural language specifications.
CUAD~\citep{hendrycks2021cuad} is an expert-annotated, domain-specific legal contract review dataset that presents a challenging research benchmark and potential for enhancing deep learning models' performance in contract understanding tasks.
CVALUES~\citep{xu2023cvalues} introduces a humanistic evaluation benchmark to assess the alignment of LLMs with safety and responsibility standards.
In the realm of comprehensive Chinese medicine, \citet{wang2023cmb} introduced CMB, a medical evaluation benchmark rooted in the Chinese language and culture. It addresses the potential inconsistency in the local context that may arise from relying solely on English-based medical assessments.
In the realm of hallucination assessment, \citep{liang2023uhgeval} has developed UHGEval, a benchmark specifically designed to evaluate the performance of Chinese LLMs in text generation without being constrained by hallucination-related limitations.

In addition to existing evaluation benchmarks, there is a research gap in assessing the effectiveness of utilizing tools for \llms. To address this gap,
the API-Bank benchmark \citep{li2023apibank} is introduced as the first benchmark explicitly designed for tool-augmented LLMs. It comprises a comprehensive Tool-Augmented LLM workflow, encompassing 53 commonly used API tools and 264 annotated dialogues, encompassing a total of 568 API calls.
Furthermore, the ToolBench project \citep{toolbench} aims to empower the development of large language models that effectively leverage the capabilities of general-purpose tools. By providing a platform for creating optimized instruction datasets, the ToolBench project seeks to drive progress in language models and enhance their practical applications.
To evaluate \llms in multi-turn interactions, \citet{wang2023mint} proposed MINT, which utilizes tools and natural language feedback.

\subsection{Benchmarks for Multi-modal task}\label{sec:benm}

For the evaluation of Multimodal Large Language Models (MLLMs), MME~\citep{fu2023mme} serves as an extensive evaluative benchmark, aiming to assess their perceptual and cognitive aptitudes. It employs meticulously crafted instruction-answer pairs alongside succinct instruction design, thereby guaranteeing equitable evaluation conditions. 
To robustly evaluate large-scale vision-language models, \citet{liu2023mmbench} introduced MMBench, which comprises a comprehensive dataset and employs a CircularEval assessment method.
Additionally, MMICL~\citep{zhao2023mmicl} enhances visual language models for multimodal inputs and excels in tasks such as MME and MMBench.
Furthermore, LAMM~\citep{yin2023lamm} extends its research to encompass multimodal point clouds.
LVLM-eHub~\citep{xu2023lvlmehub} undertakes an exhaustive evaluation of LVLMs using an online competitive platform and quantitative capacity assessments.
To comprehensively assess the generative and understanding capabilities of Multi-modal Large Language Models (MLLMs), \citet{li2023seed} introduced a novel benchmark named SEED-Bench.
This benchmark consists of 19,000 multiple-choice questions that have been annotated by human assessors. Additionally, the evaluation covers 12 different aspects, including the models' proficiency in understanding patterns within images and videos.
In summary, recent works have developed robust benchmarks and improved models that advance the study of multimodal languages.

%% file: tables/tb-benchmarks.tex
\begin{table*}
\centering
\caption{Summary of existing \llms evaluation benchmarks (ordered by the name of the first author).}
\label{tb-benchmarks}
\resizebox{\textwidth}{!}{
\begin{tabular}{|l|c|c|c|}
\hline
\textbf{Benchmark} & \textbf{Focus} & \textbf{Domain} & \textbf{Evaluation Criteria} \\
\hline\hline
SOCKET~\citep{choi2023llms} & Social knowledge & Specific downstream task &  Social language understanding \\ \hline

MME~\citep{fu2023mme} & Multimodal \llms & Multi-modal task & Ability of perception and cognition \\ \hline
Xiezhi~\citep{gu2023xiezhi} & Comprehensive domain knowledge & General language task & Overall performance across multiple benchmarks \\ \hline
Choice-75~\citep{hou2023choice}& Script learning&Specific downstream task &Overall performance of \llms  \\ \hline
CUAD~\citep{hendrycks2021cuad} &Legal contract review &Specific downstream task & Legal contract understanding \\ \hline
TRUSTGPT~\citep{huang2023trustgpt} &Ethic &Specific downstream task & Toxicity, bias, and value-alignment \\ \hline
MMLU~\citep{hendrycks2020measuring} & Text models & General language task & Multitask accuracy \\ \hline
MATH~\citep{hendrycks2021measuring-1} & Mathematical problem  & Specific downstream task & Mathematical ability \\ \hline
APPS~\citep{hendrycks2021measuring-2} & Coding challenge competence & Specific downstream task & Code generation ability \\ \hline
CELLO~\citep{he2023can}& Complex
instructions & Specific downstream task  & Four designated evaluation criteria \\ \hline
C-Eval~\citep{huang2023c} & Chinese evaluation & General language task  & 52 Exams in a Chinese context \\ \hline
EmotionBench~\citep{huang2023emotionally} & Empathy ability & Specific downstream task & Emotional changes \\ \hline
OpenLLM~\citep{leaderboard} &  Chatbots & General language task & Leaderboard rankings \\ \hline
DynaBench~\citep{kiela2021dynabench} & Dynamic evaluation & General language task & NLI, QA, sentiment, and hate speech \\ \hline
Chatbot Arena~\citep{chatbotarena} & Chat assistants & General language task & Crowdsourcing and Elo rating system \\ \hline
AlpacaEval~\citep{alpaca_eval} & Automated evaluation & General language task & Metrics, robustness, and diversity \\ \hline
CMMLU~\citep{li2023cmmlu} &Chinese multi-tasking &Specific downstream task & Multi-task language understanding capabilities \\\hline
HELM~\citep{liang2022holistic} & Holistic evaluation & General language task & Multi-metric \\\hline
API-Bank~\citep{li2023apibank} &  Tool utilization & Specific downstream task & API call, retrieval, and planning \\\hline
M3KE~\citep{liu2023m3ke} &  Multi-task & Specific downstream task & Multi-task accuracy\\\hline
MMBench~\citep{liu2023mmbench} &  Large vision-language models(LVLMs) &  Multi-modal task& Multifaceted capabilities
of VLMs\\\hline
SEED-Bench~\citep{li2023seed}& Multimodal Large Language Models &  Multi-modal task& Generative understanding of MLLMs
\\\hline
UHGEval~\citep{liang2023uhgeval} & Hallucination of Chinese \llms  & Specific downstream task &Form, metric and granularity  \\\hline
ARB~\citep{sawada2023arb} & Advanced reasoning ability & Specific downstream task & Multidomain advanced reasoning ability\\\hline
BIG-bench~\citep{srivastava2022beyond} & Capabilities and limitations of LMs & General language task & Model performance and calibration \\\hline
MultiMedQA~\citep{singhal2022large} & Medical QA & Specific downstream task & Accuracy and human evaluation \\\hline
CVALUES~\citep{xu2023cvalues} & Safety and responsibility & Specific downstream task & Alignment ability
of \llms \\\hline
LVLM-eHub~\citep{xu2023lvlmehub}& LVLMs & Multi-modal task &  Multimodal capabilities of LVLMs\\\hline
ToolBench~\citep{toolbench} & Software tools & Specific downstream task & Execution success rate \\\hline
FRESHQA~\citep{vu2023freshllms}& Dynamic QA & Specific downstream task & Correctness and hallucination \\\hline
CMB~\citep{wang2023cmb}& Chinese comprehensive medicine & Specific downstream task & Expert evaluation and automatic evaluation \\\hline
PandaLM~\citep{wang2023pandalm} & Instruction tuning & General language task & Winrate judged by PandaLM \\\hline
MINT~\citep{wang2023mint} & Multi-turn interaction & Specific downstream task & Success rate with \textit{k}-turn
 budget $SR_k$ \\\hline
Dialogue CoT~\citep{wang2023chainofthought}& In-depth dialogue & Specific downstream task & Helpfulness and acceptness of \llms \\\hline
BOSS~\citep{yuan2023revisiting}& OOD robustness in NLP& General language task & OOD robustness \\\hline
MM-Vet~\citep{yu2023mm}& Complicated multi-modal tasks & Multi-modal task & Integrated
vision-language capabilities \\\hline
LAMM~\citep{yin2023lamm}& Multi-modal point clouds & Multi-modal task & Task-specific metrics \\\hline
GLUE-X~\citep{yang2022glue} & OOD robustness for NLP tasks & General language task & OOD robustness \\\hline
KoLA~ \citep{yu2023kola} & Knowledge-oriented evaluation & General language task & Self-contrast metrics \\\hline
AGIEval~\citep{zhong2023agieval} & Human-centered foundational models & General language task & General \\\hline
PromptBench~\citep{zhu2023promptbench} & Adversarial prompt resilience & General language task & Adversarial robustness \\\hline
MT-Bench~\citep{zheng2023judging} & Multi-turn conversation & General language task & Winrate judged by GPT-4\\\hline
M3Exam~\citep{zhang2023m3exam} & Multilingual, multimodal and multilevel & Specific downstream task & Task-specific metrics \\\hline
GAOKAO-Bench~\citep{zhang2023evaluating} & Chinese Gaokao examination & Specific downstream task & Accuracy and scoring rate \\\hline
SafetyBench~\citep{zhang2023safetybench} & Safety  & Specific downstream task &Safety abilities of \llms  \\\hline
LLMEval²~\citep{zhang2023wider} & LLM Evaluator  & General language task &Acc, macro-f1 and kappa correlation coefficient  \\\hline


\end{tabular}
}
\end{table*}

%% file: sec-how-to-eval.tex
In this section, we introduce two common evaluation methods: automatic evaluation and human evaluation.
Our categorization is based on whether or not the evaluation criterion can be automatically computed.
If it can be automatically calculated, we categorize it into \emph{automatic} evaluation; otherwise, it falls into \emph{human} evaluation.

\input{tables/tb-llm-eval}

\subsection{Automatic Evaluation} 
Automated evaluation is a common, and perhaps the most popular, evaluation method that typically uses standard metrics and evaluation tools to evaluate model performance. 
Compared with human evaluation, automatic evaluation does not require intensive human participation, which not only saves time, but also reduces the impact of human subjective factors and makes the evaluation process more standardized.
For example, both \citet{qin2023chatgpt} and \citet{bang2023multitask} use automated evaluation methods to evaluate a large number of tasks.
Recently, with the development of \llms, some advanced automatic evaluation techniques are also designed to help evaluate.
\citet{lin2023llm} proposed LLM-EVAL, a unified multidimensional automatic evaluation method for open-domain conversations with \llms.
PandaLM~\citep{wang2023pandalm} can achieve reproducible and automated language model assessment by training an LLM that serves as the ``judge'' to evaluate different models.
Proposing a self-supervised evaluation framework, \citet{jain2023bring} enabled a more efficient form of evaluating models in real-world deployment by eliminating the need for laborious labeling of new data.
In addition, many benchmarks also apply automatic evaluation, such as MMLU~\citep{hendrycks2020measuring}, HELM\citep{liang2022holistic}, C-Eval~\citep{huang2023c}, AGIEval~\citep{zhong2023agieval}, AlpacaFarm~\citep{dubois2023alpacafarm}, Chatbot Arena~\citep{chatbotarena}, etc.

\input{tables/tb-automatic}

Based on the literature that adopted automatic evaluation, we summarized the main metrics in automatic evaluation in \tablename~\ref{tb-automatic}.
The key metrics include the following four aspects:
\begin{enumerate}[leftmargin=2em]
\setlength\itemsep{0em}
    \item \textbf{Accuracy} is a measure of how correct a model is on a given task. The concept of accuracy may vary in different scenarios and is dependent on the specific task and problem definition. It can be measured using various metrics such as Exact Match, F1 score, and ROUGE score.
    \begin{itemize}
        \item Exact Match (EM) is a metric used to evaluate whether the model's output in text generation tasks precisely matches the reference answer. In question answering tasks, if the model's generated answer is an exact match with the manually provided answer, the EM is 1; otherwise, it is 0.
        \item The F1 score is a metric for evaluating the performance of binary classification models, combining the model's precision and recall. The formula for calculation is as follows: \(F1=\frac{2\times Precision \times Recall }{Precision + Recall} \).
        \item ROUGE is primarily employed to assess the performance of tasks such as text summarization and machine translation, involving considerations of overlap and matching between texts.
    \end{itemize}
    
    \item \textbf{Calibrations} pertains to the degree of agreement between the confidence level of the model output and the actual prediction accuracy. 
    \begin{itemize}
        \item Expected Calibration Error (ECE) is one of the commonly used metrics to evaluate model calibration performance~\citep{guo2017calibration}. \citet{tian2023just} utilized ECE to study the calibration of RLHF-LMs, including ChatGPT, GPT-4, Claude 1, Claude 2 and Llama2. For the calculation of ECE, they categorize model predictions based on confidence and measure the average accuracy of the predictions within each confidence interval. 
         \item Area Under the Curve of selective accuracy and coverage (AUC)~\citep{geifman2017selective} is another commonly used metric. 
    \end{itemize}
    
    \item \textbf{Fairness} refers to whether the model treats different groups consistently, that is, whether the model's performance is equal across different groups. This can include attributes such as gender, race, age, and more. DecodingTrust~\citep{wang2023decodingtrust} employs the following two metrics for measuring fairness:
    \begin{itemize}
        \item Demographic Parity Pifference (DPD) measures whether the model's predictions are distributed equally across different population groups. If predictions differ significantly between groups, the DPD is high, indicating that the model may be unfairly biased against different groups. The calculation of DPD involves the prediction of the model and the true label, and the following formula can be used: \(DPD=P(\hat{y}|Z=1)-P(\hat{y}|Z=0)\), where \(\hat{y}\) is the binary classification prediction of the model, Z is the identifier of the population group (usually binary, indicating two different groups, such as men and women), \(P(\hat{y}|Z=1)\) and \(P(\hat{y}|Z=0)\) respectively represent the probabilities of predicting the positive class in population \(Z=1\) and \(Z=0\).
        \item Equalized Odds Difference (EOD) aims to ensure that the model provides equal error rates across different populations, that is, the model's prediction error probability distribution is similar for different populations. The calculation of EOD involves probabilities related to true positive (TP), true negative (TN), false positive (FP), and false negative (FN) predictions. The formula for EOD is as follows:
        \(max \{ P(\hat{y}=1|Y=1,Z=1) - P(\hat{y}=1|Y=1,Z=0), P(\hat{y}=1|Y=0,Z=1) - P(\hat{y}=1|Y=0,Z=0)\} \)
        where \(\hat{y}\) is the binary classification prediction of the model, \(Y\) is the true label, \(Z\) is the demographic group identifier (typically binary, representing two different groups), and \(P(\hat{y}=1|Y=1,Z=1)\) denotes the probability of the model predicting a positive class when the true label is positive and belongs to group \(Z=1\).
    \end{itemize}
    
    \item \textbf{Robustness} evaluates the performance of a model in the face of various challenging inputs, including adversarial attacks, changes in data distribution, noise, etc. 
    \begin{itemize}
        \item Attack Success Rate (ASR) serves as a metric for evaluating the adversarial robustness of \llms~\citep{wang2023robustness}.
        Specifically, consider a dataset \(\mathcal{D} =\left \{ (x_{i},y_{i} ) \right \}_{i=1}^{N} \) containing \(N\) pairs of samples \(x_{i}\) and ground truth \(y_{i}\). For an adversarial attack method \(\mathcal{A}\), given an input \(x\), this method can produce adversarial examples \(\mathcal{A}(x)\)  to  attack surrogate model \(f\), with the success rate is calculated as: \(ASR=\sum_{(x,y \in D )} \frac{\mathcal{I}\left [ f(\mathcal{A}(x) )\ne y \right ]  }{\mathcal{I} \left [ f(x)=y \right ] } \), where \(\mathcal{I}\) is the indicator function ~\citep{wang2021adversarial}.
        \item Performance Drop Rate (PDR), a new unified metric, effectively assesses the robustness of prompt in \llms~\citep{zhu2023promptbench}. PDR quantifies the relative performance degradation after a prompt attack, and the formula is as follows: \(PDR=1-\frac{\sum_{(x,y)\in D }\mathcal{M}\left [  f(\left [ A(P),x \right ] ),y\right ] }{\sum_{(x,y)\in D }\mathcal{M}\left [ f(\left [ P,x \right ] ),y\right ] } \), where \(A\) represents the adversarial attack applied to prompt \(P\), and \(M\) denotes the evaluation function, which varies across different tasks~\citep{zhu2023promptbench}.
    \end{itemize}
\end{enumerate}



\subsection{Human Evaluation}
The increasingly strengthened capabilities of \llms have certainly gone beyond standard evaluation metrics on general natural language tasks.
Therefore, human evaluation becomes a natural choice in some non-standard cases where automatic evaluation is not suitable.
For instance, in open-generation tasks where embedded similarity metrics (such as BERTScore) are not enough, human evaluation is more reliable \citep{novikova2017we}.
While some generation tasks can adopt certain automatic evaluation protocols, human evaluation in these tasks is more favorable as generation can always go better than standard answers.

Human evaluation is a way to evaluate the quality and accuracy of model-generated results through human participation. 
Compared with automatic evaluation, manual evaluation is closer to the actual application scenario and can provide more comprehensive and accurate feedback.
In the manual evaluation of \llms, evaluators (such as experts, researchers, or ordinary users) are usually invited to evaluate the results generated by the model. For example, \citet{ziems2023can} used the annotations from experts for generation.
By human evaluation, \citet{liang2022holistic} assessed on summarization and disinformation scenarios on 6 models and \citet{bang2023multitask} evaluated analogical reasoning tasks.
\citet{bubeck2023sparks} did a series of human-crafted tests using GPT-4 and they found that GPT-4 performs close to or even exceeds human performance on multiple tasks.
This evaluation requires human evaluators to actually test and compare the performance of the models, not just evaluate the models through automated evaluation metrics.
Note that even human evaluations can have high variance and instability, which could be due to cultural and individual differences \citep{peng1997validity}.
In practical applications, these two evaluation methods are considered and weighed in combination with the actual situation.

Exploring the human evaluation methods of \llms requires thoughtful attention to various crucial factors to guarantee the dependability and precision of assessments \citep{singhal2023large}. Table \ref{tb-human} provides a concise overview of the essential aspects of human evaluation, including the number of evaluators, evaluation criteria, and evaluator's expertise level.
Primarily, the number of evaluators emerges as a crucial factor intricately intertwined with adequate representation and statistical significance. A judiciously chosen number of evaluators contributes to a more nuanced and comprehensive understanding of the \llms under scrutiny, enabling a more reliable extrapolation of the results to a broader context. 

Furthermore, evaluation criteria are fundamental components of the human assessment process. Expanding upon the principles of the 3H rule (Helpfulness, Honesty, and Harmlessness) \citep{askell2021general}, we have elaborated them into the following 6 human assessment criteria. 
These criteria include accuracy, relevance, fluency, transparency, safety, and human alignment. Through the application of these standards, a thorough analysis of LLMs' performance in syntax, semantics, and context is achieved, allowing for a more comprehensive evaluation of the quality of generated text.

\input{tables/tb-human}

\begin{enumerate}[leftmargin=2em]
\setlength\itemsep{0em}
    \item \textbf{Accuracy} \citep{singhal2023large} stands out as a pivotal criterion that assesses the precision and correctness of the generated text. It involves scrutinizing the extent to which the language model produces information that aligns with factual knowledge, avoiding errors and inaccuracies.
    \item \textbf{Relevance} \citep{zhong2022towards} focuses on the appropriateness and significance of the generated content. This criterion examines how well the text addresses the given context or query, ensuring that the information provided is pertinent and directly applicable.
    \item \textbf{Fluency} \citep{van2019best} assesses the language model's ability to produce content that flows smoothly, maintaining a consistent tone and style. A fluent text is not only grammatically correct but also ensures readability and a seamless user experience. Analysts evaluate how well the model avoids awkward expressions and abrupt shifts in language or topic, contributing to effective communication with users.

    \item \textbf{Transparency} delves into the clarity and openness of the language model's decision-making process. It involves assessing how well the model communicates its thought processes, enabling users to understand how and why certain responses are generated. A transparent model provides insights into its inner workings.

    \item \textbf{Safety} \citep{ji2023beavertails} emerges as a critical criterion concerned with the potential harm or unintended consequences arising from the generated text. It examines the language model's ability to avoid producing content that may be inappropriate, offensive, or harmful, ensuring the well-being of users and avoiding misinformation.

    \item \textbf{Human alignment} assesses the degree to which the language model's output aligns with human values, preferences, and expectations. It considers the ethical implications of the generated content, ensuring that the language model produces text that respects societal norms and user expectations, promoting a positive interaction with human users.
\end{enumerate}

Lastly, the expertise level of evaluators is a critical consideration, encompassing relevant domain knowledge, task familiarity, and methodological training. Delineating the requisite expertise level for evaluators ensures that they possess the necessary background knowledge to accurately comprehend and assess the domain-specific text generated by \llms. This strategy adds a layer of rigor to the evaluation process, reinforcing the credibility and validity of the findings.



%% file: tables/tb-llm-eval.tex
\begin{table*}[t!]
\centering
\caption{Summary of new \llms evaluation protocols.}
\label{tb-llm-eval}
\resizebox{.96\textwidth}{!}{
\begin{tabular}{|l|l|}
\hline
\textbf{Method } & \textbf{References }\\ \hline\hline
Human-in-the-loop      &  AdaVision~\citep{gao2022adaptive}, AdaTest~\citep{ribeiro2022adaptive}          \\ \hline
Crowd-sourcing testing &   DynaBench~\citep{kiela2021dynabench}, DynaBoard~\citep{ma2021dynaboard}, DynamicTempLAMA~\citep{margatina2023dynamic}, DynaTask~\citep{thrush2022dynatask} \\ \hline
More challenging tests &   HELM~\citep{liang2022holistic}, AdaFilter~\citep{phang2021adversarially}, CheckList~\citep{ribeiro2020beyond}, Big-Bench~\citep{srivastava2022beyond}, DeepTest~\citep{tian2018deeptest}         \\ \hline
\end{tabular}
}
\end{table*}

%% file: tables/tb-automatic.tex
\begin{table*}[t!]
  \centering
  \caption{Key metrics of automatic evaluation.}
  \label{tb-automatic}
  \resizebox{.9\textwidth}{!}{
    \begin{tabular}{|p{0.2\linewidth}|m{0.9\linewidth}|}
      \hline
      \textbf{General metrics} & \textbf{Metrics} \\ \hline\hline
      Accuracy & Exact match, Quasi-exact match, F1 score, ROUGE score~\citep{lin-2004-rouge}  \\ \hline
      Calibrations &  Expected calibration error \citep{guo2017calibration}, Area under the curve~\citep{geifman2017selective}\\ \hline
      Fairness & Demographic parity difference~\citep{zemel2013learning}, Equalized odds difference~\citep{hardt2016equality} \\ \hline
      Robustness &  Attack success rate~\citep{wang2021adversarial}, Performance drop rate~\citep{zhu2023promptbench} \\\hline
    \end{tabular}
  }
\end{table*}


%% file: tables/tb-human.tex




\begin{table*}[t!]
  \centering
  \caption{Summary of key factors in human evaluation}
  \label{tb-human}
  \resizebox{\textwidth}{!}{
    \begin{tabular}{|p{0.28\linewidth}|m{0.85\linewidth}|}
      \hline
      \textbf{Evaluation Criteria} & \textbf{Key Factor} \\ \hline\hline
      Number of evaluators & Adequate representation \citep{belz2006comparing}, Statistical significance \\ \hline

      Evaluation rubrics  & Accuracy \citep{singhal2023large}, Relevance \citep{zhong2022towards}, Fluency \citep{van2019best}, Transparency, Safety \citep{ji2023beavertails}, Human alignment \\ \hline

      Evaluator's expertise level & Relevant domain expertise \citep{olney2023generating}, Task familiarity, Methodological training \\ \hline
    \end{tabular}
  }
\end{table*}



%% file: sec-summary.tex
In this section, we summarize the key findings based on our review in sections~\ref{sec-what}, \ref{sec-where}, and \ref{sec-how}.

First of all, we would like to highlight that despite all the efforts spent on summarizing existing works on evaluation, there is \emph{no} evidence to explicitly show that one certain evaluation protocol or benchmark is the most useful and successful, but with \textbf{different characteristics and focuses}.
This also demonstrates that not a single model can perform best in all kinds of tasks.
The purpose of this survey is to go beyond simply determining the ``best'' benchmark or evaluation protocol.
By summarizing and analyzing existing efforts on \llms evaluation, we may identify the current success and failure cases of \llms, derive new trends for evaluation protocols, and most importantly, propose new challenges and opportunities for future research.

\subsection{Task: Success and Failure Cases of \llms}\label{sec:s}

We now summarize the success and failure cases of \llms in different tasks.
Note that all the following conclusions are made based on existing evaluation efforts and the results are only dependent on specific datasets.
\subsubsection{What can \llms do well?}
\begin{itemize}
    \item \llms demonstrate proficiency in generating text \citep{brown2020language, bommasani2021opportunities, chowdhery2022palm} by producing fluent and precise linguistic expressions.
    \item \llms obtain impressive performance in tasks involving language understanding, including sentiment analysis \citep{gao2020making, lopez2023can, 
 qin2023chatgpt}, text classification \citep{liang2022holistic, yang2023large, pena2023leveraging}, as well as the handling of factual input \citep{qin2023chatgpt}.

    \item \llms demonstrate robust arithmetic reasoning capabilities \citep{qin2023chatgpt} and excel in logical reasoning \citep{liu2023evaluating}. Moreover, they exhibit noteworthy proficiency in temporal reasoning \citep{bang2023multitask}. Furthermore, more intricate tasks such as mathematical reasoning \citep{zhang2023evaluating1,wu2022autoformalization,yu2023metamath} and structured data inference \citep{pan2023unifying,jiang2023structgpt} have emerged as the prevailing benchmarks for evaluation.
    \item \llms exhibit robust contextual comprehension, enabling them to generate coherent responses that align with the given input \citep{thoppilan2022lamda}.
    \item \llms also achieve satisfying performance across several natural language processing tasks, including machine translation \citep{wang2023document, bang2023multitask, lyu2023new}, text generation \citep{chen2023exploring}, and question answering \citep{liang2022holistic, laskar2023systematic}.
\end{itemize}

\subsubsection{When can \llms fail?}
\begin{itemize}
\item Within the realm of NLI, \llms exhibit subpar performance and encounter challenges in accurately representing human disagreements \citep{lee2023can}. 
\item \llms exhibit restricted proficiency in discerning semantic similarity between events \citep{tao2023eveval} and demonstrate substandard performance in evaluating fundamental phrases \citep{riccardi2023two}.
\item \llms have limited abilities on abstract reasoning \citep{gendron2023large}, and are prone to confusion or errors in complex contexts \citep{ott2023thoughtsource}.
\item In linguistic contexts featuring non-Latin scripts and limited resources, \llms manifest suboptimal performance \citep{bang2023multitask,zhang2023m3exam,lai2023chatgpt, ahuja2023mega}. Furthermore, generative \llms consistently display proficiency levels below the expected standards across various tasks and languages \citep{ahuja2023mega}.
\item \llms demonstrate susceptibility when processing visual modal information \citep{zhao2023evaluating}. Furthermore, they have the capacity to assimilate, disseminate, and potentially magnify detrimental content found within the acquired training datasets, frequently encompassing toxic linguistic elements, including offensive, hostile, and derogatory language \citep{gehman2020realtoxicityprompts}.
\item \llms may exhibit social biases and toxicity \citep{gehman2020realtoxicityprompts,dhamala2021bold,parrish2022bbq} during the generation process, resulting in the production of biased outputs.
\item \llms may manifest credibility deficits \citep{wang2023decodingtrust}, potentially giving rise to fabricated information or erroneous facts within dialogues \citep{rawte2023survey,zhang2023sirens}.

\item \llms have limitations in incorporating real-time or dynamic information \citep{liu2023summary}, making them less suitable for tasks that require up-to-date knowledge or rapid adaptation to changing contexts.
\item \llms is sensitive to prompts, especially adversarial prompts \citep{zhu2023promptbench}, which trigger new evaluations and algorithms to improve its robustness.
\end{itemize}


\subsection{Benchmark and Evaluation Protocol}

With the rapid development and widespread use of \llms, the importance of evaluating them in practical applications and research has become crucial. This evaluation process should include not only task-level evaluation but also a deep understanding of the potential risks they pose from a societal perspective.
In this section, we summarize existing benchmarks and protocols in \tablename~\ref{tb-llm-eval}.

First, a shift from objective calculation to human-in-the-loop testing, allowing for greater human feedback during the evaluation process.
AdaVision \citep{gao2022adaptive}, an interactive process for testing vision models, enables users to label a small amount of data for model correctness, which helps users identify and fix coherent failure modes.
In AdaTest \citep{ribeiro2022adaptive}, the user filters test samples by only selecting high-quality tests and organizing them into semantically related topics. 

Second, a move from static to crowd-sourcing test sets is becoming more common.
Tools like DynaBench \citep{kiela2021dynabench}, DynaBoard \citep{ma2021dynaboard}, and DynaTask \citep{thrush2022dynatask} rely on crowdworkers to create and test hard samples.
Additionally, DynamicTempLAMA \citep{margatina2023dynamic} allows for dynamically constructed time-related tests.

Third, a shift from a unified to a challenging setting in evaluating machine learning models.
While unified settings involve a test set with no preference for any specific task, challenging settings create test sets for specific tasks.
Tools like DeepTest \citep{tian2018deeptest} use seeds to generate input transformations for testing, CheckList \citep{ribeiro2020beyond} builds test sets based on templates, and AdaFilter \citep{phang2021adversarially} adversarially constructs tests.
However, it is worth noting that AdaFilter may not be entirely fair as it relies on adversarial examples.
HELM \citep{liang2022holistic} evaluates LLMs from different aspects, while the Big-Bench \citep{srivastava2022beyond} platform is used to design hard tasks for machine learning models to tackle.
PromptBench~\citep{zhu2023promptbench} aims to evaluate the adversarial robustness of \llms by creating adversarial prompts, which is more challenging and the results demonstrated that current \llms are not robust to adversarial prompts.

%% file: sec-challenge.tex
\textbf{Evaluation as a new discipline:}
Our summarization inspires us to redesign a wide spectrum of aspects related to evaluation in the era of \llms.
In this section, we present several grand challenges.
Our key point is that \textbf{evaluation should be treated as an essential discipline to drive the success of \llms and other AI models}.
Existing protocols are not enough to thoroughly evaluate the true capabilities of \llms, which poses grand challenges and triggers new opportunities for future research on \llms evaluation.

\subsection{Designing AGI Benchmarks}

As we discussed earlier, while all tasks can potentially serve as evaluation tools for \llms, the question remains as to which can truly measure AGI capabilities. 
As we expect \llms to demonstrate AGI abilities, a comprehensive understanding of the differences between human and AGI capacities becomes crucial in the creation of AGI benchmarks.
The prevailing trend seems to conceptualize AGI as a superhuman entity, thereby utilizing cross-disciplinary knowledge from fields such as education, psychology, and social sciences to design innovative benchmarks.
Nonetheless, there remains a plethora of unresolved issues. For instance, does it make sense to use human values as a starting point for test construction, or should alternative perspectives be considered? 
Developing suitable AGI benchmarks presents many open questions demanding further exploration.

\subsection{Complete Behavioral Evaluation}

An ideal AGI evaluation should contain not only standard benchmarks on common tasks, but also evaluations on open tasks such as complete behavioral tests.
By behavioral test, we mean that AGI models should also be evaluated in an open environment.
For instance, by treating \llms as the central controller, we can construct evaluations on a robot manipulated by \llms to test its behaviors in real situations.
By treating \llms as a completely intelligent machine, the evaluations of its multi-modal dimensions should also be considered.
In fact, complete behavioral evaluations are complementary to standard AGI benchmarks and they should work together for better testing.

\subsection{Robustness Evaluation}

Beyond general tasks, it is crucial for \llms to maintain robustness against a wide variety of inputs in order to perform optimally for end-users, given their extensive integration into daily life.
For instance, the same prompts but with different grammars and expressions could lead ChatGPT and other \llms to generate diverse results, indicating that current \llms are not robust to the inputs.
While there are some prior work on robustness evaluation~\citep{wang2023robustness,zhu2023promptbench}, there are much room for advancement, such as including more diverse evaluation sets, examining more evaluation aspects, and developing more efficient evaluations to generate robustness tasks.
Concurrently, the concept and definition of robustness are constantly evolving. It is thus vital to consider updating the evaluation system to better align with emerging requirements related to ethics and bias.

\subsection{Dynamic and Evolving Evaluation}

Existing evaluation protocols for most AI tasks rely on static and public benchmarks, i.e., the evaluation datasets and protocols are often publicly available.
While this facilitates rapid and convenient evaluation within the community, it is unable to accurately assess the evolving abilities of \llms, given their rapid rate of development.
The capabilities of \llms may enhance over time which cannot be consistently evaluated by existing static benchmarks.
On the other hand, as \llms grow increasingly powerful with larger model sizes and training set sizes, static and public benchmarks are likely to be memorized by \llms, resulting in potential training data contamination.
Therefore, developing dynamic and evolving evaluation systems is the key to providing a fair evaluation of \llms.

\subsection{Principled and Trustworthy Evaluation}

When introducing an evaluation system, it is crucial to ascertain its integrity and trustworthiness.
Therefore, the necessity for trustworthy computing extends to the requirement for reliable evaluation systems as well.
This poses a challenging research question that intertwines with measurement theory, probability, and numerous other domains.
For instance, how can we ensure that dynamic testing truly generates out-of-distribution examples?
There is a scarcity of research in this domain, and it is hoped that future work will aim to scrutinize not only the algorithms but the evaluation system itself.

\subsection{Unified Evaluation that Supports All \llms Tasks}

There are many other research areas of \llms and we need to develop evaluation systems that can support all kinds of tasks such as value alignment, safety, verification, interdisciplinary research, fine-tuning, and others.
For instance, PandaLM~\citep{wang2023pandalm} is an evaluation system that assists \llms fine-tuning by providing an open-source evaluation model, which can automatically assess the performance of fine-tuning.
We expect that more evaluation systems are becoming more general and can be used as assistance in certain \llms tasks.

\subsection{Beyond Evaluation: \llms Enhancement}

Ultimately, evaluation is not the end goal but rather the starting point. 
Following the evaluation, there are undoubtedly conclusions to be drawn regarding performance, robustness, stability, and other factors. 
A proficient evaluation system should not only offer benchmark results but should also deliver an insightful analysis, recommendations, and guidance for future research and development. 
For instance, PromptBench~\citep{zhu2023promptbench} provides not only robustness evaluation results on adversarial prompts but also a comprehensive analysis through attention visualization, elucidating how adversarial texts can result in erroneous responses. 
The system further offers a word frequency analysis to identify robust and non-robust words in the test sets, thus providing prompt engineering guidance for end users. 
Subsequent research can leverage these findings to enhance \llms.
Another example is that \citet{wang2023exploring} first explored the performance of large vision-language models on imbalanced (long-tailed) tasks, which demonstrates the limitation of current large models.
Then, they explored different methodologies to enhance the performance on these tasks.
In summary, enhancement after evaluation helps to build better \llms and much can be done in the future.

%% file: main.bbl

\begin{thebibliography}{269}


\ifx \showCODEN    \undefined \def \showCODEN     #1{\unskip}     \fi
\ifx \showDOI      \undefined \def \showDOI       #1{#1}\fi
\ifx \showISBNx    \undefined \def \showISBNx     #1{\unskip}     \fi
\ifx \showISBNxiii \undefined \def \showISBNxiii  #1{\unskip}     \fi
\ifx \showISSN     \undefined \def \showISSN      #1{\unskip}     \fi
\ifx \showLCCN     \undefined \def \showLCCN      #1{\unskip}     \fi
\ifx \shownote     \undefined \def \shownote      #1{#1}          \fi
\ifx \showarticletitle \undefined \def \showarticletitle #1{#1}   \fi
\ifx \showURL      \undefined \def \showURL       {\relax}        \fi
\providecommand\bibfield[2]{#2}
\providecommand\bibinfo[2]{#2}
\providecommand\natexlab[1]{#1}
\providecommand\showeprint[2][]{arXiv:#2}

\bibitem[Abdelali et~al\mbox{.}(2023)]%
        {abdelali2023benchmarking}
\bibfield{author}{\bibinfo{person}{Ahmed Abdelali}, \bibinfo{person}{Hamdy
  Mubarak}, \bibinfo{person}{Shammur~Absar Chowdhury}, \bibinfo{person}{Maram
  Hasanain}, \bibinfo{person}{Basel Mousi}, \bibinfo{person}{Sabri Boughorbel},
  \bibinfo{person}{Yassine~El Kheir}, \bibinfo{person}{Daniel Izham},
  \bibinfo{person}{Fahim Dalvi}, \bibinfo{person}{Majd Hawasly},
  {et~al\mbox{.}}} \bibinfo{year}{2023}\natexlab{}.
\newblock \showarticletitle{Benchmarking Arabic AI with Large Language Models}.
\newblock \bibinfo{journal}{\emph{arXiv preprint arXiv:2305.14982}}
  (\bibinfo{year}{2023}).
\newblock


\bibitem[Ahuja et~al\mbox{.}(2023)]%
        {ahuja2023mega}
\bibfield{author}{\bibinfo{person}{Kabir Ahuja}, \bibinfo{person}{Rishav Hada},
  \bibinfo{person}{Millicent Ochieng}, \bibinfo{person}{Prachi Jain},
  \bibinfo{person}{Harshita Diddee}, \bibinfo{person}{Samuel Maina},
  \bibinfo{person}{Tanuja Ganu}, \bibinfo{person}{Sameer Segal},
  \bibinfo{person}{Maxamed Axmed}, \bibinfo{person}{Kalika Bali},
  {et~al\mbox{.}}} \bibinfo{year}{2023}\natexlab{}.
\newblock \showarticletitle{Mega: Multilingual evaluation of generative ai}.
\newblock \bibinfo{journal}{\emph{arXiv preprint arXiv:2303.12528}}
  (\bibinfo{year}{2023}).
\newblock


\bibitem[Arora et~al\mbox{.}(2023)]%
        {arora2023have}
\bibfield{author}{\bibinfo{person}{Daman Arora},
  \bibinfo{person}{Himanshu~Gaurav Singh}, {et~al\mbox{.}}}
  \bibinfo{year}{2023}\natexlab{}.
\newblock \showarticletitle{Have LLMs Advanced Enough? A Challenging Problem
  Solving Benchmark For Large Language Models}.
\newblock \bibinfo{journal}{\emph{arXiv preprint arXiv:2305.15074}}
  (\bibinfo{year}{2023}).
\newblock


\bibitem[Askell et~al\mbox{.}(2021)]%
        {askell2021general}
\bibfield{author}{\bibinfo{person}{Amanda Askell}, \bibinfo{person}{Yuntao
  Bai}, \bibinfo{person}{Anna Chen}, \bibinfo{person}{Dawn Drain},
  \bibinfo{person}{Deep Ganguli}, \bibinfo{person}{Tom Henighan},
  \bibinfo{person}{Andy Jones}, \bibinfo{person}{Nicholas Joseph},
  \bibinfo{person}{Ben Mann}, \bibinfo{person}{Nova DasSarma}, {et~al\mbox{.}}}
  \bibinfo{year}{2021}\natexlab{}.
\newblock \showarticletitle{A general language assistant as a laboratory for
  alignment}.
\newblock \bibinfo{journal}{\emph{arXiv preprint arXiv:2112.00861}}
  (\bibinfo{year}{2021}).
\newblock


\bibitem[Bai et~al\mbox{.}(2023)]%
        {bai2023benchmarking}
\bibfield{author}{\bibinfo{person}{Yushi Bai}, \bibinfo{person}{Jiahao Ying},
  \bibinfo{person}{Yixin Cao}, \bibinfo{person}{Xin Lv}, \bibinfo{person}{Yuze
  He}, \bibinfo{person}{Xiaozhi Wang}, \bibinfo{person}{Jifan Yu},
  \bibinfo{person}{Kaisheng Zeng}, \bibinfo{person}{Yijia Xiao},
  \bibinfo{person}{Haozhe Lyu}, {et~al\mbox{.}}}
  \bibinfo{year}{2023}\natexlab{}.
\newblock \showarticletitle{Benchmarking Foundation Models with
  Language-Model-as-an-Examiner}.
\newblock \bibinfo{journal}{\emph{arXiv preprint arXiv:2306.04181}}
  (\bibinfo{year}{2023}).
\newblock


\bibitem[Bang et~al\mbox{.}(2023)]%
        {bang2023multitask}
\bibfield{author}{\bibinfo{person}{Yejin Bang}, \bibinfo{person}{Samuel
  Cahyawijaya}, \bibinfo{person}{Nayeon Lee}, \bibinfo{person}{Wenliang Dai},
  \bibinfo{person}{Dan Su}, \bibinfo{person}{Bryan Wilie},
  \bibinfo{person}{Holy Lovenia}, \bibinfo{person}{Ziwei Ji},
  \bibinfo{person}{Tiezheng Yu}, \bibinfo{person}{Willy Chung},
  {et~al\mbox{.}}} \bibinfo{year}{2023}\natexlab{}.
\newblock \showarticletitle{A multitask, multilingual, multimodal evaluation of
  chatgpt on reasoning, hallucination, and interactivity}.
\newblock \bibinfo{journal}{\emph{arXiv preprint arXiv:2302.04023}}
  (\bibinfo{year}{2023}).
\newblock


\bibitem[Belz and Reiter(2006)]%
        {belz2006comparing}
\bibfield{author}{\bibinfo{person}{Anja Belz} {and} \bibinfo{person}{Ehud
  Reiter}.} \bibinfo{year}{2006}\natexlab{}.
\newblock \showarticletitle{Comparing automatic and human evaluation of NLG
  systems}. In \bibinfo{booktitle}{\emph{11th conference of the european
  chapter of the association for computational linguistics}}.
  \bibinfo{pages}{313--320}.
\newblock


\bibitem[Berrar(2019)]%
        {berrar2019cross}
\bibfield{author}{\bibinfo{person}{Daniel Berrar}.}
  \bibinfo{year}{2019}\natexlab{}.
\newblock \bibinfo{title}{Cross-Validation.}
\newblock
\newblock


\bibitem[Bian et~al\mbox{.}(2023)]%
        {bian2023chatgpt}
\bibfield{author}{\bibinfo{person}{Ning Bian}, \bibinfo{person}{Xianpei Han},
  \bibinfo{person}{Le Sun}, \bibinfo{person}{Hongyu Lin},
  \bibinfo{person}{Yaojie Lu}, {and} \bibinfo{person}{Ben He}.}
  \bibinfo{year}{2023}\natexlab{}.
\newblock \showarticletitle{Chatgpt is a knowledgeable but inexperienced
  solver: An investigation of commonsense problem in large language models}.
\newblock \bibinfo{journal}{\emph{arXiv preprint arXiv:2303.16421}}
  (\bibinfo{year}{2023}).
\newblock


\bibitem[Bodroza et~al\mbox{.}(2023)]%
        {bodroza2023personality}
\bibfield{author}{\bibinfo{person}{Bojana Bodroza}, \bibinfo{person}{Bojana~M
  Dinic}, {and} \bibinfo{person}{Ljubisa Bojic}.}
  \bibinfo{year}{2023}\natexlab{}.
\newblock \showarticletitle{Personality testing of GPT-3: Limited temporal
  reliability, but highlighted social desirability of GPT-3's personality
  instruments results}.
\newblock \bibinfo{journal}{\emph{arXiv preprint arXiv:2306.04308}}
  (\bibinfo{year}{2023}).
\newblock


\bibitem[Bommasani et~al\mbox{.}(2021)]%
        {bommasani2021opportunities}
\bibfield{author}{\bibinfo{person}{Rishi Bommasani}, \bibinfo{person}{Drew~A
  Hudson}, \bibinfo{person}{Ehsan Adeli}, \bibinfo{person}{Russ Altman},
  \bibinfo{person}{Simran Arora}, \bibinfo{person}{Sydney von Arx},
  \bibinfo{person}{Michael~S Bernstein}, \bibinfo{person}{Jeannette Bohg},
  \bibinfo{person}{Antoine Bosselut}, \bibinfo{person}{Emma Brunskill},
  {et~al\mbox{.}}} \bibinfo{year}{2021}\natexlab{}.
\newblock \showarticletitle{On the opportunities and risks of foundation
  models}.
\newblock \bibinfo{journal}{\emph{arXiv preprint arXiv:2108.07258}}
  (\bibinfo{year}{2021}).
\newblock


\bibitem[Brody(1999)]%
        {brody1999intelligence}
\bibfield{author}{\bibinfo{person}{Nathan Brody}.}
  \bibinfo{year}{1999}\natexlab{}.
\newblock \showarticletitle{What is intelligence?}
\newblock \bibinfo{journal}{\emph{International Review of Psychiatry}}
  \bibinfo{volume}{11}, \bibinfo{number}{1} (\bibinfo{year}{1999}),
  \bibinfo{pages}{19--25}.
\newblock


\bibitem[Brown et~al\mbox{.}(1992)]%
        {brown1992class}
\bibfield{author}{\bibinfo{person}{Peter~F Brown}, \bibinfo{person}{Vincent~J
  Della~Pietra}, \bibinfo{person}{Peter~V Desouza}, \bibinfo{person}{Jennifer~C
  Lai}, {and} \bibinfo{person}{Robert~L Mercer}.}
  \bibinfo{year}{1992}\natexlab{}.
\newblock \showarticletitle{Class-based n-gram models of natural language}.
\newblock \bibinfo{journal}{\emph{Computational linguistics}}
  \bibinfo{volume}{18}, \bibinfo{number}{4} (\bibinfo{year}{1992}),
  \bibinfo{pages}{467--480}.
\newblock


\bibitem[Brown et~al\mbox{.}(2020)]%
        {brown2020language}
\bibfield{author}{\bibinfo{person}{Tom Brown}, \bibinfo{person}{Benjamin Mann},
  \bibinfo{person}{Nick Ryder}, \bibinfo{person}{Melanie Subbiah},
  \bibinfo{person}{Jared~D Kaplan}, \bibinfo{person}{Prafulla Dhariwal},
  \bibinfo{person}{Arvind Neelakantan}, \bibinfo{person}{Pranav Shyam},
  \bibinfo{person}{Girish Sastry}, \bibinfo{person}{Amanda Askell},
  {et~al\mbox{.}}} \bibinfo{year}{2020}\natexlab{}.
\newblock \showarticletitle{Language models are few-shot learners}.
\newblock \bibinfo{journal}{\emph{Advances in neural information processing
  systems}}  \bibinfo{volume}{33} (\bibinfo{year}{2020}),
  \bibinfo{pages}{1877--1901}.
\newblock


\bibitem[Bubeck et~al\mbox{.}(2023)]%
        {bubeck2023sparks}
\bibfield{author}{\bibinfo{person}{S{\'e}bastien Bubeck},
  \bibinfo{person}{Varun Chandrasekaran}, \bibinfo{person}{Ronen Eldan},
  \bibinfo{person}{Johannes Gehrke}, \bibinfo{person}{Eric Horvitz},
  \bibinfo{person}{Ece Kamar}, \bibinfo{person}{Peter Lee},
  \bibinfo{person}{Yin~Tat Lee}, \bibinfo{person}{Yuanzhi Li},
  \bibinfo{person}{Scott Lundberg}, {et~al\mbox{.}}}
  \bibinfo{year}{2023}\natexlab{}.
\newblock \showarticletitle{Sparks of artificial general intelligence: Early
  experiments with gpt-4}.
\newblock \bibinfo{journal}{\emph{arXiv preprint arXiv:2303.12712}}
  (\bibinfo{year}{2023}).
\newblock


\bibitem[Cao et~al\mbox{.}(2023)]%
        {cao2023assessing}
\bibfield{author}{\bibinfo{person}{Yong Cao}, \bibinfo{person}{Li Zhou},
  \bibinfo{person}{Seolhwa Lee}, \bibinfo{person}{Laura Cabello},
  \bibinfo{person}{Min Chen}, {and} \bibinfo{person}{Daniel Hershcovich}.}
  \bibinfo{year}{2023}\natexlab{}.
\newblock \showarticletitle{Assessing Cross-Cultural Alignment between ChatGPT
  and Human Societies: An Empirical Study}. In
  \bibinfo{booktitle}{\emph{Proceedings of the First Workshop on Cross-Cultural
  Considerations in NLP (C3NLP)}}. \bibinfo{pages}{53--67}.
\newblock


\bibitem[Cascella et~al\mbox{.}(2023)]%
        {cascella2023evaluating}
\bibfield{author}{\bibinfo{person}{Marco Cascella}, \bibinfo{person}{Jonathan
  Montomoli}, \bibinfo{person}{Valentina Bellini}, {and} \bibinfo{person}{Elena
  Bignami}.} \bibinfo{year}{2023}\natexlab{}.
\newblock \showarticletitle{Evaluating the feasibility of ChatGPT in
  healthcare: an analysis of multiple clinical and research scenarios}.
\newblock \bibinfo{journal}{\emph{Journal of Medical Systems}}
  \bibinfo{volume}{47}, \bibinfo{number}{1} (\bibinfo{year}{2023}),
  \bibinfo{pages}{33}.
\newblock


\bibitem[Castro~Nascimento and Pimentel(2023)]%
        {castro2023large}
\bibfield{author}{\bibinfo{person}{Cayque~Monteiro Castro~Nascimento} {and}
  \bibinfo{person}{Andr{\'e}~Silva Pimentel}.} \bibinfo{year}{2023}\natexlab{}.
\newblock \showarticletitle{Do Large Language Models Understand Chemistry? A
  Conversation with ChatGPT}.
\newblock \bibinfo{journal}{\emph{Journal of Chemical Information and
  Modeling}} \bibinfo{volume}{63}, \bibinfo{number}{6} (\bibinfo{year}{2023}),
  \bibinfo{pages}{1649--1655}.
\newblock


\bibitem[Chen et~al\mbox{.}(2021)]%
        {chen2021evaluating}
\bibfield{author}{\bibinfo{person}{Mark Chen}, \bibinfo{person}{Jerry Tworek},
  \bibinfo{person}{Heewoo Jun}, \bibinfo{person}{Qiming Yuan},
  \bibinfo{person}{Henrique Ponde de~Oliveira Pinto}, \bibinfo{person}{Jared
  Kaplan}, \bibinfo{person}{Harri Edwards}, \bibinfo{person}{Yuri Burda},
  \bibinfo{person}{Nicholas Joseph}, \bibinfo{person}{Greg Brockman},
  {et~al\mbox{.}}} \bibinfo{year}{2021}\natexlab{}.
\newblock \showarticletitle{Evaluating large language models trained on code}.
\newblock \bibinfo{journal}{\emph{arXiv preprint arXiv:2107.03374}}
  (\bibinfo{year}{2021}).
\newblock


\bibitem[Chen et~al\mbox{.}(2023)]%
        {chen2023exploring}
\bibfield{author}{\bibinfo{person}{Yi Chen}, \bibinfo{person}{Rui Wang},
  \bibinfo{person}{Haiyun Jiang}, \bibinfo{person}{Shuming Shi}, {and}
  \bibinfo{person}{Ruifeng Xu}.} \bibinfo{year}{2023}\natexlab{}.
\newblock \showarticletitle{Exploring the use of large language models for
  reference-free text quality evaluation: A preliminary empirical study}.
\newblock \bibinfo{journal}{\emph{arXiv preprint arXiv:2304.00723}}
  (\bibinfo{year}{2023}).
\newblock


\bibitem[Chervenak et~al\mbox{.}(2023)]%
        {chervenak2023promise}
\bibfield{author}{\bibinfo{person}{Joseph Chervenak}, \bibinfo{person}{Harry
  Lieman}, \bibinfo{person}{Miranda Blanco-Breindel}, {and}
  \bibinfo{person}{Sangita Jindal}.} \bibinfo{year}{2023}\natexlab{}.
\newblock \showarticletitle{The promise and peril of using a large language
  model to obtain clinical information: ChatGPT performs strongly as a
  fertility counseling tool with limitations}.
\newblock \bibinfo{journal}{\emph{Fertility and Sterility}}
  (\bibinfo{year}{2023}).
\newblock


\bibitem[Chia et~al\mbox{.}(2023)]%
        {chia2023instructeval}
\bibfield{author}{\bibinfo{person}{Yew~Ken Chia}, \bibinfo{person}{Pengfei
  Hong}, \bibinfo{person}{Lidong Bing}, {and} \bibinfo{person}{Soujanya
  Poria}.} \bibinfo{year}{2023}\natexlab{}.
\newblock \showarticletitle{INSTRUCTEVAL: Towards Holistic Evaluation of
  Instruction-Tuned Large Language Models}.
\newblock \bibinfo{journal}{\emph{arXiv preprint arXiv:2306.04757}}
  (\bibinfo{year}{2023}).
\newblock


\bibitem[Choi et~al\mbox{.}(2023)]%
        {choi2023llms}
\bibfield{author}{\bibinfo{person}{Minje Choi}, \bibinfo{person}{Jiaxin Pei},
  \bibinfo{person}{Sagar Kumar}, \bibinfo{person}{Chang Shu}, {and}
  \bibinfo{person}{David Jurgens}.} \bibinfo{year}{2023}\natexlab{}.
\newblock \showarticletitle{Do LLMs Understand Social Knowledge? Evaluating the
  Sociability of Large Language Models with SocKET Benchmark}.
\newblock \bibinfo{journal}{\emph{arXiv preprint arXiv:2305.14938}}
  (\bibinfo{year}{2023}).
\newblock


\bibitem[Chowdhery et~al\mbox{.}(2022)]%
        {chowdhery2022palm}
\bibfield{author}{\bibinfo{person}{Aakanksha Chowdhery},
  \bibinfo{person}{Sharan Narang}, \bibinfo{person}{Jacob Devlin},
  \bibinfo{person}{Maarten Bosma}, \bibinfo{person}{Gaurav Mishra},
  \bibinfo{person}{Adam Roberts}, \bibinfo{person}{Paul Barham},
  \bibinfo{person}{Hyung~Won Chung}, \bibinfo{person}{Charles Sutton},
  \bibinfo{person}{Sebastian Gehrmann}, {et~al\mbox{.}}}
  \bibinfo{year}{2022}\natexlab{}.
\newblock \showarticletitle{Palm: Scaling language modeling with pathways}.
\newblock \bibinfo{journal}{\emph{arXiv preprint arXiv:2204.02311}}
  (\bibinfo{year}{2022}).
\newblock


\bibitem[Christiano et~al\mbox{.}(2017)]%
        {christiano2017deep}
\bibfield{author}{\bibinfo{person}{Paul~F Christiano}, \bibinfo{person}{Jan
  Leike}, \bibinfo{person}{Tom Brown}, \bibinfo{person}{Miljan Martic},
  \bibinfo{person}{Shane Legg}, {and} \bibinfo{person}{Dario Amodei}.}
  \bibinfo{year}{2017}\natexlab{}.
\newblock \showarticletitle{Deep reinforcement learning from human
  preferences}.
\newblock \bibinfo{journal}{\emph{Advances in neural information processing
  systems}}  \bibinfo{volume}{30} (\bibinfo{year}{2017}).
\newblock


\bibitem[Clavi{\'e} et~al\mbox{.}(2023)]%
        {clavie2023large}
\bibfield{author}{\bibinfo{person}{Benjamin Clavi{\'e}},
  \bibinfo{person}{Alexandru Ciceu}, \bibinfo{person}{Frederick Naylor},
  \bibinfo{person}{Guillaume Souli{\'e}}, {and} \bibinfo{person}{Thomas
  Brightwell}.} \bibinfo{year}{2023}\natexlab{}.
\newblock \showarticletitle{Large Language Models in the Workplace: A Case
  Study on Prompt Engineering for Job Type Classification}. In
  \bibinfo{booktitle}{\emph{International Conference on Applications of Natural
  Language to Information Systems}}. Springer, \bibinfo{pages}{3--17}.
\newblock


\bibitem[Collins et~al\mbox{.}(2023)]%
        {collins2023evaluating}
\bibfield{author}{\bibinfo{person}{Katherine~M Collins},
  \bibinfo{person}{Albert~Q Jiang}, \bibinfo{person}{Simon Frieder},
  \bibinfo{person}{Lionel Wong}, \bibinfo{person}{Miri Zilka},
  \bibinfo{person}{Umang Bhatt}, \bibinfo{person}{Thomas Lukasiewicz},
  \bibinfo{person}{Yuhuai Wu}, \bibinfo{person}{Joshua~B Tenenbaum},
  \bibinfo{person}{William Hart}, {et~al\mbox{.}}}
  \bibinfo{year}{2023}\natexlab{}.
\newblock \showarticletitle{Evaluating Language Models for Mathematics through
  Interactions}.
\newblock \bibinfo{journal}{\emph{arXiv preprint arXiv:2306.01694}}
  (\bibinfo{year}{2023}).
\newblock


\bibitem[Cortes and Vapnik(1995)]%
        {cortes1995support}
\bibfield{author}{\bibinfo{person}{Corinna Cortes} {and}
  \bibinfo{person}{Vladimir Vapnik}.} \bibinfo{year}{1995}\natexlab{}.
\newblock \showarticletitle{Support-vector networks}.
\newblock \bibinfo{journal}{\emph{Machine learning}}  \bibinfo{volume}{20}
  (\bibinfo{year}{1995}), \bibinfo{pages}{273--297}.
\newblock


\bibitem[Dai et~al\mbox{.}(2023b)]%
        {dai2023uncovering}
\bibfield{author}{\bibinfo{person}{Sunhao Dai}, \bibinfo{person}{Ninglu Shao},
  \bibinfo{person}{Haiyuan Zhao}, \bibinfo{person}{Weijie Yu},
  \bibinfo{person}{Zihua Si}, \bibinfo{person}{Chen Xu},
  \bibinfo{person}{Zhongxiang Sun}, \bibinfo{person}{Xiao Zhang}, {and}
  \bibinfo{person}{Jun Xu}.} \bibinfo{year}{2023}\natexlab{b}.
\newblock \showarticletitle{Uncovering ChatGPT's Capabilities in Recommender
  Systems}.
\newblock \bibinfo{journal}{\emph{arXiv preprint arXiv:2305.02182}}
  (\bibinfo{year}{2023}).
\newblock


\bibitem[Dai et~al\mbox{.}(2023a)]%
        {dai2023can}
\bibfield{author}{\bibinfo{person}{Wei Dai}, \bibinfo{person}{Jionghao Lin},
  \bibinfo{person}{Flora Jin}, \bibinfo{person}{Tongguang Li},
  \bibinfo{person}{Yi-Shan Tsai}, \bibinfo{person}{Dragan Gasevic}, {and}
  \bibinfo{person}{Guanliang Chen}.} \bibinfo{year}{2023}\natexlab{a}.
\newblock \showarticletitle{Can large language models provide feedback to
  students? a case study on chatgpt}.
\newblock  (\bibinfo{year}{2023}).
\newblock


\bibitem[Dao and Le(2023)]%
        {dao2023investigating}
\bibfield{author}{\bibinfo{person}{Xuan-Quy Dao} {and}
  \bibinfo{person}{Ngoc-Bich Le}.} \bibinfo{year}{2023}\natexlab{}.
\newblock \showarticletitle{Investigating the Effectiveness of ChatGPT in
  Mathematical Reasoning and Problem Solving: Evidence from the Vietnamese
  National High School Graduation Examination}.
\newblock \bibinfo{journal}{\emph{arXiv preprint arXiv:2306.06331}}
  (\bibinfo{year}{2023}).
\newblock


\bibitem[de{} Winter(2023)]%
        {de2023can}
\bibfield{author}{\bibinfo{person}{Joost~CF de{} Winter}.}
  \bibinfo{year}{2023}\natexlab{}.
\newblock \showarticletitle{Can ChatGPT pass high school exams on English
  language comprehension}.
\newblock \bibinfo{journal}{\emph{Researchgate. Preprint}}
  (\bibinfo{year}{2023}).
\newblock


\bibitem[Deng et~al\mbox{.}(2009)]%
        {deng2009imagenet}
\bibfield{author}{\bibinfo{person}{Jia Deng}, \bibinfo{person}{Wei Dong},
  \bibinfo{person}{Richard Socher}, \bibinfo{person}{Li-Jia Li},
  \bibinfo{person}{Kai Li}, {and} \bibinfo{person}{Li Fei-Fei}.}
  \bibinfo{year}{2009}\natexlab{}.
\newblock \showarticletitle{Imagenet: A large-scale hierarchical image
  database}. In \bibinfo{booktitle}{\emph{2009 IEEE conference on computer
  vision and pattern recognition}}. Ieee, \bibinfo{pages}{248--255}.
\newblock


\bibitem[Deroy et~al\mbox{.}(2023)]%
        {deroy2023ready}
\bibfield{author}{\bibinfo{person}{Aniket Deroy}, \bibinfo{person}{Kripabandhu
  Ghosh}, {and} \bibinfo{person}{Saptarshi Ghosh}.}
  \bibinfo{year}{2023}\natexlab{}.
\newblock \showarticletitle{How Ready are Pre-trained Abstractive Models and
  LLMs for Legal Case Judgement Summarization?}
\newblock \bibinfo{journal}{\emph{arXiv preprint arXiv:2306.01248}}
  (\bibinfo{year}{2023}).
\newblock


\bibitem[Deshpande et~al\mbox{.}(2023)]%
        {deshpande2023toxicity}
\bibfield{author}{\bibinfo{person}{Ameet Deshpande}, \bibinfo{person}{Vishvak
  Murahari}, \bibinfo{person}{Tanmay Rajpurohit}, \bibinfo{person}{Ashwin
  Kalyan}, {and} \bibinfo{person}{Karthik Narasimhan}.}
  \bibinfo{year}{2023}\natexlab{}.
\newblock \showarticletitle{Toxicity in chatgpt: Analyzing persona-assigned
  language models}.
\newblock \bibinfo{journal}{\emph{arXiv preprint arXiv:2304.05335}}
  (\bibinfo{year}{2023}).
\newblock


\bibitem[Devlin et~al\mbox{.}(2018)]%
        {devlin2018bert}
\bibfield{author}{\bibinfo{person}{Jacob Devlin}, \bibinfo{person}{Ming-Wei
  Chang}, \bibinfo{person}{Kenton Lee}, {and} \bibinfo{person}{Kristina
  Toutanova}.} \bibinfo{year}{2018}\natexlab{}.
\newblock \showarticletitle{Bert: Pre-training of deep bidirectional
  transformers for language understanding}.
\newblock \bibinfo{journal}{\emph{arXiv preprint arXiv:1810.04805}}
  (\bibinfo{year}{2018}).
\newblock


\bibitem[Dhamala et~al\mbox{.}(2021)]%
        {dhamala2021bold}
\bibfield{author}{\bibinfo{person}{Jwala Dhamala}, \bibinfo{person}{Tony Sun},
  \bibinfo{person}{Varun Kumar}, \bibinfo{person}{Satyapriya Krishna},
  \bibinfo{person}{Yada Pruksachatkun}, \bibinfo{person}{Kai-Wei Chang}, {and}
  \bibinfo{person}{Rahul Gupta}.} \bibinfo{year}{2021}\natexlab{}.
\newblock \showarticletitle{Bold: Dataset and metrics for measuring biases in
  open-ended language generation}. In \bibinfo{booktitle}{\emph{Proceedings of
  the 2021 ACM conference on fairness, accountability, and transparency}}.
  \bibinfo{pages}{862--872}.
\newblock


\bibitem[Dubois et~al\mbox{.}(2023)]%
        {dubois2023alpacafarm}
\bibfield{author}{\bibinfo{person}{Yann Dubois}, \bibinfo{person}{Xuechen Li},
  \bibinfo{person}{Rohan Taori}, \bibinfo{person}{Tianyi Zhang},
  \bibinfo{person}{Ishaan Gulrajani}, \bibinfo{person}{Jimmy Ba},
  \bibinfo{person}{Carlos Guestrin}, \bibinfo{person}{Percy Liang}, {and}
  \bibinfo{person}{Tatsunori~B Hashimoto}.} \bibinfo{year}{2023}\natexlab{}.
\newblock \showarticletitle{Alpacafarm: A simulation framework for methods that
  learn from human feedback}.
\newblock \bibinfo{journal}{\emph{arXiv preprint arXiv:2305.14387}}
  (\bibinfo{year}{2023}).
\newblock


\bibitem[Duong and Solomon(2023)]%
        {duong2023analysis}
\bibfield{author}{\bibinfo{person}{Dat Duong} {and} \bibinfo{person}{Benjamin~D
  Solomon}.} \bibinfo{year}{2023}\natexlab{}.
\newblock \showarticletitle{Analysis of large-language model versus human
  performance for genetics questions}.
\newblock \bibinfo{journal}{\emph{European Journal of Human Genetics}}
  (\bibinfo{year}{2023}), \bibinfo{pages}{1--3}.
\newblock


\bibitem[Fan et~al\mbox{.}(2023)]%
        {fan2023recommender}
\bibfield{author}{\bibinfo{person}{Wenqi Fan}, \bibinfo{person}{Zihuai Zhao},
  \bibinfo{person}{Jiatong Li}, \bibinfo{person}{Yunqing Liu},
  \bibinfo{person}{Xiaowei Mei}, \bibinfo{person}{Yiqi Wang},
  \bibinfo{person}{Jiliang Tang}, {and} \bibinfo{person}{Qing Li}.}
  \bibinfo{year}{2023}\natexlab{}.
\newblock \bibinfo{title}{Recommender Systems in the Era of Large Language
  Models (LLMs)}.
\newblock
\newblock
\showeprint[arxiv]{2307.02046}~[cs.IR]


\bibitem[Fansi~Tchango et~al\mbox{.}(2022)]%
        {fansi2022ddxplus}
\bibfield{author}{\bibinfo{person}{Arsene Fansi~Tchango},
  \bibinfo{person}{Rishab Goel}, \bibinfo{person}{Zhi Wen},
  \bibinfo{person}{Julien Martel}, {and} \bibinfo{person}{Joumana Ghosn}.}
  \bibinfo{year}{2022}\natexlab{}.
\newblock \showarticletitle{Ddxplus: A new dataset for automatic medical
  diagnosis}.
\newblock \bibinfo{journal}{\emph{Advances in Neural Information Processing
  Systems}}  \bibinfo{volume}{35} (\bibinfo{year}{2022}),
  \bibinfo{pages}{31306--31318}.
\newblock


\bibitem[Ferrara(2023)]%
        {ferrara2023should}
\bibfield{author}{\bibinfo{person}{Emilio Ferrara}.}
  \bibinfo{year}{2023}\natexlab{}.
\newblock \showarticletitle{Should chatgpt be biased? challenges and risks of
  bias in large language models}.
\newblock \bibinfo{journal}{\emph{arXiv preprint arXiv:2304.03738}}
  (\bibinfo{year}{2023}).
\newblock


\bibitem[Floridi and Chiriatti(2020)]%
        {floridi2020gpt}
\bibfield{author}{\bibinfo{person}{Luciano Floridi} {and}
  \bibinfo{person}{Massimo Chiriatti}.} \bibinfo{year}{2020}\natexlab{}.
\newblock \showarticletitle{GPT-3: Its nature, scope, limits, and
  consequences}.
\newblock \bibinfo{journal}{\emph{Minds and Machines}}  \bibinfo{volume}{30}
  (\bibinfo{year}{2020}), \bibinfo{pages}{681--694}.
\newblock


\bibitem[Frank(2023)]%
        {frank2023baby}
\bibfield{author}{\bibinfo{person}{Michael~C Frank}.}
  \bibinfo{year}{2023}\natexlab{}.
\newblock \showarticletitle{Baby steps in evaluating the capacities of large
  language models}.
\newblock \bibinfo{journal}{\emph{Nature Reviews Psychology}}
  (\bibinfo{year}{2023}), \bibinfo{pages}{1--2}.
\newblock


\bibitem[Frieder et~al\mbox{.}(2023)]%
        {frieder2023mathematical}
\bibfield{author}{\bibinfo{person}{Simon Frieder}, \bibinfo{person}{Luca
  Pinchetti}, \bibinfo{person}{Ryan-Rhys Griffiths}, \bibinfo{person}{Tommaso
  Salvatori}, \bibinfo{person}{Thomas Lukasiewicz},
  \bibinfo{person}{Philipp~Christian Petersen}, \bibinfo{person}{Alexis
  Chevalier}, {and} \bibinfo{person}{Julius Berner}.}
  \bibinfo{year}{2023}\natexlab{}.
\newblock \showarticletitle{Mathematical capabilities of chatgpt}.
\newblock \bibinfo{journal}{\emph{arXiv preprint arXiv:2301.13867}}
  (\bibinfo{year}{2023}).
\newblock


\bibitem[Fu et~al\mbox{.}(2023a)]%
        {fu2023mme}
\bibfield{author}{\bibinfo{person}{Chaoyou Fu}, \bibinfo{person}{Peixian Chen},
  \bibinfo{person}{Yunhang Shen}, \bibinfo{person}{Yulei Qin},
  \bibinfo{person}{Mengdan Zhang}, \bibinfo{person}{Xu Lin},
  \bibinfo{person}{Zhenyu Qiu}, \bibinfo{person}{Wei Lin},
  \bibinfo{person}{Jinrui Yang}, \bibinfo{person}{Xiawu Zheng},
  {et~al\mbox{.}}} \bibinfo{year}{2023}\natexlab{a}.
\newblock \showarticletitle{MME: A Comprehensive Evaluation Benchmark for
  Multimodal Large Language Models}.
\newblock \bibinfo{journal}{\emph{arXiv preprint arXiv:2306.13394}}
  (\bibinfo{year}{2023}).
\newblock


\bibitem[Fu et~al\mbox{.}(2023b)]%
        {fu2023chain}
\bibfield{author}{\bibinfo{person}{Yao Fu}, \bibinfo{person}{Litu Ou},
  \bibinfo{person}{Mingyu Chen}, \bibinfo{person}{Yuhao Wan},
  \bibinfo{person}{Hao Peng}, {and} \bibinfo{person}{Tushar Khot}.}
  \bibinfo{year}{2023}\natexlab{b}.
\newblock \showarticletitle{Chain-of-Thought Hub: A Continuous Effort to
  Measure Large Language Models' Reasoning Performance}.
\newblock \bibinfo{journal}{\emph{arXiv preprint arXiv:2305.17306}}
  (\bibinfo{year}{2023}).
\newblock


\bibitem[Fushiki(2011)]%
        {fushiki2011estimation}
\bibfield{author}{\bibinfo{person}{Tadayoshi Fushiki}.}
  \bibinfo{year}{2011}\natexlab{}.
\newblock \showarticletitle{Estimation of prediction error by using K-fold
  cross-validation}.
\newblock \bibinfo{journal}{\emph{Statistics and Computing}}
  \bibinfo{volume}{21} (\bibinfo{year}{2011}), \bibinfo{pages}{137--146}.
\newblock


\bibitem[Gallant et~al\mbox{.}(1990)]%
        {gallant1990perceptron}
\bibfield{author}{\bibinfo{person}{Stephen~I Gallant} {et~al\mbox{.}}}
  \bibinfo{year}{1990}\natexlab{}.
\newblock \showarticletitle{Perceptron-based learning algorithms}.
\newblock \bibinfo{journal}{\emph{IEEE Transactions on neural networks}}
  \bibinfo{volume}{1}, \bibinfo{number}{2} (\bibinfo{year}{1990}),
  \bibinfo{pages}{179--191}.
\newblock


\bibitem[Gao et~al\mbox{.}(2022)]%
        {gao2022adaptive}
\bibfield{author}{\bibinfo{person}{Irena Gao}, \bibinfo{person}{Gabriel
  Ilharco}, \bibinfo{person}{Scott Lundberg}, {and}
  \bibinfo{person}{Marco~Tulio Ribeiro}.} \bibinfo{year}{2022}\natexlab{}.
\newblock \showarticletitle{Adaptive Testing of Computer Vision Models}.
\newblock \bibinfo{journal}{\emph{arXiv preprint arXiv:2212.02774}}
  (\bibinfo{year}{2022}).
\newblock


\bibitem[Gao and Lin(2004)]%
        {gao2004introduction}
\bibfield{author}{\bibinfo{person}{Jianfeng Gao} {and}
  \bibinfo{person}{Chin-Yew Lin}.} \bibinfo{year}{2004}\natexlab{}.
\newblock \bibinfo{title}{Introduction to the special issue on statistical
  language modeling}.
\newblock , \bibinfo{numpages}{87--93}~pages.
\newblock


\bibitem[Gao et~al\mbox{.}(2020)]%
        {gao2020making}
\bibfield{author}{\bibinfo{person}{Tianyu Gao}, \bibinfo{person}{Adam Fisch},
  {and} \bibinfo{person}{Danqi Chen}.} \bibinfo{year}{2020}\natexlab{}.
\newblock \showarticletitle{Making pre-trained language models better few-shot
  learners}.
\newblock \bibinfo{journal}{\emph{arXiv preprint arXiv:2012.15723}}
  (\bibinfo{year}{2020}).
\newblock


\bibitem[Gehman et~al\mbox{.}(2020)]%
        {gehman2020realtoxicityprompts}
\bibfield{author}{\bibinfo{person}{Samuel Gehman}, \bibinfo{person}{Suchin
  Gururangan}, \bibinfo{person}{Maarten Sap}, \bibinfo{person}{Yejin Choi},
  {and} \bibinfo{person}{Noah~A Smith}.} \bibinfo{year}{2020}\natexlab{}.
\newblock \showarticletitle{RealToxicityPrompts: Evaluating Neural Toxic
  Degeneration in Language Models}. In \bibinfo{booktitle}{\emph{Findings of
  the Association for Computational Linguistics: EMNLP 2020}}.
  \bibinfo{pages}{3356--3369}.
\newblock


\bibitem[Geifman and El-Yaniv(2017)]%
        {geifman2017selective}
\bibfield{author}{\bibinfo{person}{Yonatan Geifman} {and} \bibinfo{person}{Ran
  El-Yaniv}.} \bibinfo{year}{2017}\natexlab{}.
\newblock \showarticletitle{Selective classification for deep neural networks}.
\newblock \bibinfo{journal}{\emph{Advances in neural information processing
  systems}}  \bibinfo{volume}{30} (\bibinfo{year}{2017}).
\newblock


\bibitem[Gekhman et~al\mbox{.}(2023)]%
        {gekhman2023trueteacher}
\bibfield{author}{\bibinfo{person}{Zorik Gekhman}, \bibinfo{person}{Jonathan
  Herzig}, \bibinfo{person}{Roee Aharoni}, \bibinfo{person}{Chen Elkind}, {and}
  \bibinfo{person}{Idan Szpektor}.} \bibinfo{year}{2023}\natexlab{}.
\newblock \showarticletitle{Trueteacher: Learning factual consistency
  evaluation with large language models}.
\newblock \bibinfo{journal}{\emph{arXiv preprint arXiv:2305.11171}}
  (\bibinfo{year}{2023}).
\newblock


\bibitem[Gendron et~al\mbox{.}(2023)]%
        {gendron2023large}
\bibfield{author}{\bibinfo{person}{Ga{\"e}l Gendron}, \bibinfo{person}{Qiming
  Bao}, \bibinfo{person}{Michael Witbrock}, {and} \bibinfo{person}{Gillian
  Dobbie}.} \bibinfo{year}{2023}\natexlab{}.
\newblock \showarticletitle{Large Language Models Are Not Abstract Reasoners}.
\newblock \bibinfo{journal}{\emph{arXiv preprint arXiv:2305.19555}}
  (\bibinfo{year}{2023}).
\newblock


\bibitem[Gilson et~al\mbox{.}(2023)]%
        {gilson2023does}
\bibfield{author}{\bibinfo{person}{Aidan Gilson}, \bibinfo{person}{Conrad~W
  Safranek}, \bibinfo{person}{Thomas Huang}, \bibinfo{person}{Vimig Socrates},
  \bibinfo{person}{Ling Chi}, \bibinfo{person}{Richard~Andrew Taylor},
  \bibinfo{person}{David Chartash}, {et~al\mbox{.}}}
  \bibinfo{year}{2023}\natexlab{}.
\newblock \showarticletitle{How does CHATGPT perform on the United States
  Medical Licensing Examination? the implications of large language models for
  medical education and knowledge assessment}.
\newblock \bibinfo{journal}{\emph{JMIR Medical Education}} \bibinfo{volume}{9},
  \bibinfo{number}{1} (\bibinfo{year}{2023}), \bibinfo{pages}{e45312}.
\newblock


\bibitem[Graham et~al\mbox{.}(2013)]%
        {graham2013moral}
\bibfield{author}{\bibinfo{person}{Jesse Graham}, \bibinfo{person}{Jonathan
  Haidt}, \bibinfo{person}{Sena Koleva}, \bibinfo{person}{Matt Motyl},
  \bibinfo{person}{Ravi Iyer}, \bibinfo{person}{Sean~P Wojcik}, {and}
  \bibinfo{person}{Peter~H Ditto}.} \bibinfo{year}{2013}\natexlab{}.
\newblock \showarticletitle{Moral foundations theory: The pragmatic validity of
  moral pluralism}.
\newblock In \bibinfo{booktitle}{\emph{Advances in experimental social
  psychology}}. Vol.~\bibinfo{volume}{47}. \bibinfo{publisher}{Elsevier},
  \bibinfo{pages}{55--130}.
\newblock


\bibitem[Gu et~al\mbox{.}(2023)]%
        {gu2023xiezhi}
\bibfield{author}{\bibinfo{person}{Zhouhong Gu}, \bibinfo{person}{Xiaoxuan
  Zhu}, \bibinfo{person}{Haoning Ye}, \bibinfo{person}{Lin Zhang},
  \bibinfo{person}{Jianchen Wang}, \bibinfo{person}{Sihang Jiang},
  \bibinfo{person}{Zhuozhi Xiong}, \bibinfo{person}{Zihan Li},
  \bibinfo{person}{Qianyu He}, \bibinfo{person}{Rui Xu}, {et~al\mbox{.}}}
  \bibinfo{year}{2023}\natexlab{}.
\newblock \showarticletitle{Xiezhi: An Ever-Updating Benchmark for Holistic
  Domain Knowledge Evaluation}.
\newblock \bibinfo{journal}{\emph{arXiv preprint arXiv:2306.05783}}
  (\bibinfo{year}{2023}).
\newblock


\bibitem[Guo et~al\mbox{.}(2017)]%
        {guo2017calibration}
\bibfield{author}{\bibinfo{person}{Chuan Guo}, \bibinfo{person}{Geoff Pleiss},
  \bibinfo{person}{Yu Sun}, {and} \bibinfo{person}{Kilian~Q Weinberger}.}
  \bibinfo{year}{2017}\natexlab{}.
\newblock \showarticletitle{On calibration of modern neural networks}. In
  \bibinfo{booktitle}{\emph{International conference on machine learning}}.
  PMLR, \bibinfo{pages}{1321--1330}.
\newblock


\bibitem[Guo et~al\mbox{.}(2023)]%
        {guo2023indeed}
\bibfield{author}{\bibinfo{person}{Taicheng Guo}, \bibinfo{person}{Kehan Guo},
  \bibinfo{person}{Zhengwen Liang}, \bibinfo{person}{Zhichun Guo},
  \bibinfo{person}{Nitesh~V Chawla}, \bibinfo{person}{Olaf Wiest},
  \bibinfo{person}{Xiangliang Zhang}, {et~al\mbox{.}}}
  \bibinfo{year}{2023}\natexlab{}.
\newblock \showarticletitle{What indeed can GPT models do in chemistry? A
  comprehensive benchmark on eight tasks}.
\newblock \bibinfo{journal}{\emph{arXiv preprint arXiv:2305.18365}}
  (\bibinfo{year}{2023}).
\newblock


\bibitem[Hagendorff and Fabi(2023)]%
        {hagendorff2023humanlike}
\bibfield{author}{\bibinfo{person}{Thilo Hagendorff} {and}
  \bibinfo{person}{Sarah Fabi}.} \bibinfo{year}{2023}\natexlab{}.
\newblock \bibinfo{title}{Human-Like Intuitive Behavior and Reasoning Biases
  Emerged in Language Models -- and Disappeared in GPT-4}.
\newblock
\newblock
\showeprint[arxiv]{2306.07622}~[cs.CL]


\bibitem[Hamidi and Roberts(2023)]%
        {hamidi2023evaluation}
\bibfield{author}{\bibinfo{person}{Alaleh Hamidi} {and} \bibinfo{person}{Kirk
  Roberts}.} \bibinfo{year}{2023}\natexlab{}.
\newblock \showarticletitle{Evaluation of AI Chatbots for Patient-Specific EHR
  Questions}.
\newblock \bibinfo{journal}{\emph{arXiv preprint arXiv:2306.02549}}
  (\bibinfo{year}{2023}).
\newblock


\bibitem[Hardt et~al\mbox{.}(2016)]%
        {hardt2016equality}
\bibfield{author}{\bibinfo{person}{Moritz Hardt}, \bibinfo{person}{Eric Price},
  {and} \bibinfo{person}{Nati Srebro}.} \bibinfo{year}{2016}\natexlab{}.
\newblock \showarticletitle{Equality of opportunity in supervised learning}.
\newblock \bibinfo{journal}{\emph{Advances in neural information processing
  systems}}  \bibinfo{volume}{29} (\bibinfo{year}{2016}).
\newblock


\bibitem[Hartmann et~al\mbox{.}(2023)]%
        {hartmann2023political}
\bibfield{author}{\bibinfo{person}{Jochen Hartmann}, \bibinfo{person}{Jasper
  Schwenzow}, {and} \bibinfo{person}{Maximilian Witte}.}
  \bibinfo{year}{2023}\natexlab{}.
\newblock \showarticletitle{The political ideology of conversational AI:
  Converging evidence on ChatGPT's pro-environmental, left-libertarian
  orientation}.
\newblock \bibinfo{journal}{\emph{arXiv preprint arXiv:2301.01768}}
  (\bibinfo{year}{2023}).
\newblock


\bibitem[He et~al\mbox{.}(2023)]%
        {he2023can}
\bibfield{author}{\bibinfo{person}{Qianyu He}, \bibinfo{person}{Jie Zeng},
  \bibinfo{person}{Wenhao Huang}, \bibinfo{person}{Lina Chen},
  \bibinfo{person}{Jin Xiao}, \bibinfo{person}{Qianxi He},
  \bibinfo{person}{Xunzhe Zhou}, \bibinfo{person}{Lida Chen},
  \bibinfo{person}{Xintao Wang}, \bibinfo{person}{Yuncheng Huang},
  {et~al\mbox{.}}} \bibinfo{year}{2023}\natexlab{}.
\newblock \showarticletitle{Can Large Language Models Understand Real-World
  Complex Instructions?}
\newblock \bibinfo{journal}{\emph{arXiv preprint arXiv:2309.09150}}
  (\bibinfo{year}{2023}).
\newblock


\bibitem[Hellas et~al\mbox{.}(2023)]%
        {hellas2023exploring}
\bibfield{author}{\bibinfo{person}{Arto Hellas}, \bibinfo{person}{Juho
  Leinonen}, \bibinfo{person}{Sami Sarsa}, \bibinfo{person}{Charles Koutcheme},
  \bibinfo{person}{Lilja Kujanp{\"a}{\"a}}, {and} \bibinfo{person}{Juha
  Sorva}.} \bibinfo{year}{2023}\natexlab{}.
\newblock \showarticletitle{Exploring the Responses of Large Language Models to
  Beginner Programmers' Help Requests}.
\newblock \bibinfo{journal}{\emph{arXiv preprint arXiv:2306.05715}}
  (\bibinfo{year}{2023}).
\newblock


\bibitem[Hendrycks et~al\mbox{.}(2021a)]%
        {hendrycks2021measuring-2}
\bibfield{author}{\bibinfo{person}{Dan Hendrycks}, \bibinfo{person}{Steven
  Basart}, \bibinfo{person}{Saurav Kadavath}, \bibinfo{person}{Mantas Mazeika},
  \bibinfo{person}{Akul Arora}, \bibinfo{person}{Ethan Guo},
  \bibinfo{person}{Collin Burns}, \bibinfo{person}{Samir Puranik},
  \bibinfo{person}{Horace He}, \bibinfo{person}{Dawn Song}, {et~al\mbox{.}}}
  \bibinfo{year}{2021}\natexlab{a}.
\newblock \showarticletitle{Measuring coding challenge competence with apps}.
\newblock \bibinfo{journal}{\emph{arXiv preprint arXiv:2105.09938}}
  (\bibinfo{year}{2021}).
\newblock


\bibitem[Hendrycks et~al\mbox{.}(2020a)]%
        {hendrycks2020aligning}
\bibfield{author}{\bibinfo{person}{Dan Hendrycks}, \bibinfo{person}{Collin
  Burns}, \bibinfo{person}{Steven Basart}, \bibinfo{person}{Andrew Critch},
  \bibinfo{person}{Jerry Li}, \bibinfo{person}{Dawn Song}, {and}
  \bibinfo{person}{Jacob Steinhardt}.} \bibinfo{year}{2020}\natexlab{a}.
\newblock \showarticletitle{Aligning ai with shared human values}.
\newblock \bibinfo{journal}{\emph{arXiv preprint arXiv:2008.02275}}
  (\bibinfo{year}{2020}).
\newblock


\bibitem[Hendrycks et~al\mbox{.}(2020b)]%
        {hendrycks2020measuring}
\bibfield{author}{\bibinfo{person}{Dan Hendrycks}, \bibinfo{person}{Collin
  Burns}, \bibinfo{person}{Steven Basart}, \bibinfo{person}{Andy Zou},
  \bibinfo{person}{Mantas Mazeika}, \bibinfo{person}{Dawn Song}, {and}
  \bibinfo{person}{Jacob Steinhardt}.} \bibinfo{year}{2020}\natexlab{b}.
\newblock \showarticletitle{Measuring massive multitask language
  understanding}.
\newblock \bibinfo{journal}{\emph{arXiv preprint arXiv:2009.03300}}
  (\bibinfo{year}{2020}).
\newblock


\bibitem[Hendrycks et~al\mbox{.}(2021b)]%
        {hendrycks2021cuad}
\bibfield{author}{\bibinfo{person}{Dan Hendrycks}, \bibinfo{person}{Collin
  Burns}, \bibinfo{person}{Anya Chen}, {and} \bibinfo{person}{Spencer Ball}.}
  \bibinfo{year}{2021}\natexlab{b}.
\newblock \showarticletitle{Cuad: An expert-annotated nlp dataset for legal
  contract review}.
\newblock \bibinfo{journal}{\emph{arXiv preprint arXiv:2103.06268}}
  (\bibinfo{year}{2021}).
\newblock


\bibitem[Hendrycks et~al\mbox{.}(2021c)]%
        {hendrycks2021measuring-1}
\bibfield{author}{\bibinfo{person}{Dan Hendrycks}, \bibinfo{person}{Collin
  Burns}, \bibinfo{person}{Saurav Kadavath}, \bibinfo{person}{Akul Arora},
  \bibinfo{person}{Steven Basart}, \bibinfo{person}{Eric Tang},
  \bibinfo{person}{Dawn Song}, {and} \bibinfo{person}{Jacob Steinhardt}.}
  \bibinfo{year}{2021}\natexlab{c}.
\newblock \showarticletitle{Measuring mathematical problem solving with the
  math dataset}.
\newblock \bibinfo{journal}{\emph{arXiv preprint arXiv:2103.03874}}
  (\bibinfo{year}{2021}).
\newblock


\bibitem[Holmes et~al\mbox{.}(2023)]%
        {holmes2023evaluating}
\bibfield{author}{\bibinfo{person}{Jason Holmes}, \bibinfo{person}{Zhengliang
  Liu}, \bibinfo{person}{Lian Zhang}, \bibinfo{person}{Yuzhen Ding},
  \bibinfo{person}{Terence~T Sio}, \bibinfo{person}{Lisa~A McGee},
  \bibinfo{person}{Jonathan~B Ashman}, \bibinfo{person}{Xiang Li},
  \bibinfo{person}{Tianming Liu}, \bibinfo{person}{Jiajian Shen},
  {et~al\mbox{.}}} \bibinfo{year}{2023}\natexlab{}.
\newblock \showarticletitle{Evaluating large language models on a
  highly-specialized topic, radiation oncology physics}.
\newblock \bibinfo{journal}{\emph{arXiv preprint arXiv:2304.01938}}
  (\bibinfo{year}{2023}).
\newblock


\bibitem[Honovich et~al\mbox{.}(2022)]%
        {honovich2022true}
\bibfield{author}{\bibinfo{person}{Or Honovich}, \bibinfo{person}{Roee
  Aharoni}, \bibinfo{person}{Jonathan Herzig}, \bibinfo{person}{Hagai
  Taitelbaum}, \bibinfo{person}{Doron Kukliansy}, \bibinfo{person}{Vered
  Cohen}, \bibinfo{person}{Thomas Scialom}, \bibinfo{person}{Idan Szpektor},
  \bibinfo{person}{Avinatan Hassidim}, {and} \bibinfo{person}{Yossi Matias}.}
  \bibinfo{year}{2022}\natexlab{}.
\newblock \showarticletitle{TRUE: Re-evaluating factual consistency
  evaluation}.
\newblock \bibinfo{journal}{\emph{arXiv preprint arXiv:2204.04991}}
  (\bibinfo{year}{2022}).
\newblock


\bibitem[Hou et~al\mbox{.}(2023)]%
        {hou2023choice}
\bibfield{author}{\bibinfo{person}{Zhaoyi~Joey Hou}, \bibinfo{person}{Li
  Zhang}, {and} \bibinfo{person}{Chris Callison-Burch}.}
  \bibinfo{year}{2023}\natexlab{}.
\newblock \showarticletitle{Choice-75: A Dataset on Decision Branching in
  Script Learning}.
\newblock \bibinfo{journal}{\emph{arXiv preprint arXiv:2309.11737}}
  (\bibinfo{year}{2023}).
\newblock


\bibitem[Huang et~al\mbox{.}(2023c)]%
        {huang2023emotionally}
\bibfield{author}{\bibinfo{person}{Jen{-}tse Huang}, \bibinfo{person}{Man~Ho
  Lam}, \bibinfo{person}{Eric~John Li}, \bibinfo{person}{Shujie Ren},
  \bibinfo{person}{Wenxuan Wang}, \bibinfo{person}{Wenxiang Jiao},
  \bibinfo{person}{Zhaopeng Tu}, {and} \bibinfo{person}{Michael~R. Lyu}.}
  \bibinfo{year}{2023}\natexlab{c}.
\newblock \showarticletitle{Emotionally Numb or Empathetic? Evaluating How
  {LLM}s Feel Using Emotion{B}ench}.
\newblock \bibinfo{journal}{\emph{arXiv preprint arXiv:2308.03656}}
  (\bibinfo{year}{2023}).
\newblock


\bibitem[Huang et~al\mbox{.}(2023b)]%
        {huang2023language}
\bibfield{author}{\bibinfo{person}{Shaohan Huang}, \bibinfo{person}{Li Dong},
  \bibinfo{person}{Wenhui Wang}, \bibinfo{person}{Yaru Hao},
  \bibinfo{person}{Saksham Singhal}, \bibinfo{person}{Shuming Ma},
  \bibinfo{person}{Tengchao Lv}, \bibinfo{person}{Lei Cui},
  \bibinfo{person}{Owais~Khan Mohammed}, \bibinfo{person}{Qiang Liu},
  {et~al\mbox{.}}} \bibinfo{year}{2023}\natexlab{b}.
\newblock \showarticletitle{Language is not all you need: Aligning perception
  with language models}.
\newblock \bibinfo{journal}{\emph{arXiv preprint arXiv:2302.14045}}
  (\bibinfo{year}{2023}).
\newblock


\bibitem[Huang et~al\mbox{.}(2023a)]%
        {huang2023c}
\bibfield{author}{\bibinfo{person}{Yuzhen Huang}, \bibinfo{person}{Yuzhuo Bai},
  \bibinfo{person}{Zhihao Zhu}, \bibinfo{person}{Junlei Zhang},
  \bibinfo{person}{Jinghan Zhang}, \bibinfo{person}{Tangjun Su},
  \bibinfo{person}{Junteng Liu}, \bibinfo{person}{Chuancheng Lv},
  \bibinfo{person}{Yikai Zhang}, \bibinfo{person}{Jiayi Lei}, {et~al\mbox{.}}}
  \bibinfo{year}{2023}\natexlab{a}.
\newblock \showarticletitle{C-eval: A multi-level multi-discipline chinese
  evaluation suite for foundation models}.
\newblock \bibinfo{journal}{\emph{arXiv preprint arXiv:2305.08322}}
  (\bibinfo{year}{2023}).
\newblock


\bibitem[Huang et~al\mbox{.}(2023d)]%
        {huang2023trustgpt}
\bibfield{author}{\bibinfo{person}{Yue Huang}, \bibinfo{person}{Qihui Zhang},
  \bibinfo{person}{Philip~S. Y}, {and} \bibinfo{person}{Lichao Sun}.}
  \bibinfo{year}{2023}\natexlab{d}.
\newblock \bibinfo{title}{TrustGPT: A Benchmark for Trustworthy and Responsible
  Large Language Models}.
\newblock
\newblock
\showeprint[arxiv]{2306.11507}~[cs.CL]


\bibitem[HuggingFace(2023)]%
        {leaderboard}
\bibfield{author}{\bibinfo{person}{HuggingFace}.}
  \bibinfo{year}{2023}\natexlab{}.
\newblock \bibinfo{title}{Open-source Large Language Models Leaderboard}.
\newblock
  \bibinfo{howpublished}{\url{https://huggingface.co/spaces/HuggingFaceH4/open_llm_leaderboard}}.
\newblock


\bibitem[Jahan et~al\mbox{.}(2023)]%
        {jahan2023evaluation}
\bibfield{author}{\bibinfo{person}{Israt Jahan},
  \bibinfo{person}{Md~Tahmid~Rahman Laskar}, \bibinfo{person}{Chun Peng}, {and}
  \bibinfo{person}{Jimmy Huang}.} \bibinfo{year}{2023}\natexlab{}.
\newblock \showarticletitle{Evaluation of ChatGPT on Biomedical Tasks: A
  Zero-Shot Comparison with Fine-Tuned Generative Transformers}.
\newblock \bibinfo{journal}{\emph{arXiv preprint arXiv:2306.04504}}
  (\bibinfo{year}{2023}).
\newblock


\bibitem[Jain et~al\mbox{.}(2023)]%
        {jain2023bring}
\bibfield{author}{\bibinfo{person}{Neel Jain}, \bibinfo{person}{Khalid
  Saifullah}, \bibinfo{person}{Yuxin Wen}, \bibinfo{person}{John Kirchenbauer},
  \bibinfo{person}{Manli Shu}, \bibinfo{person}{Aniruddha Saha},
  \bibinfo{person}{Micah Goldblum}, \bibinfo{person}{Jonas Geiping}, {and}
  \bibinfo{person}{Tom Goldstein}.} \bibinfo{year}{2023}\natexlab{}.
\newblock \showarticletitle{Bring Your Own Data! Self-Supervised Evaluation for
  Large Language Models}.
\newblock \bibinfo{journal}{\emph{arXiv preprint arXiv:2306.13651}}
  (\bibinfo{year}{2023}).
\newblock


\bibitem[Jansson et~al\mbox{.}(2021)]%
        {jansson2021online}
\bibfield{author}{\bibinfo{person}{Malin Jansson}, \bibinfo{person}{Stefan
  Hrastinski}, \bibinfo{person}{Stefan Stenbom}, {and} \bibinfo{person}{Fredrik
  Enoksson}.} \bibinfo{year}{2021}\natexlab{}.
\newblock \showarticletitle{Online question and answer sessions: How students
  support their own and other students' processes of inquiry in a text-based
  learning environment}.
\newblock \bibinfo{journal}{\emph{The Internet and Higher Education}}
  \bibinfo{volume}{51} (\bibinfo{year}{2021}), \bibinfo{pages}{100817}.
\newblock


\bibitem[Jentzsch and Kersting(2023)]%
        {jentzsch2023chatgpt}
\bibfield{author}{\bibinfo{person}{Sophie Jentzsch} {and}
  \bibinfo{person}{Kristian Kersting}.} \bibinfo{year}{2023}\natexlab{}.
\newblock \showarticletitle{ChatGPT is fun, but it is not funny! Humor is still
  challenging Large Language Models}.
\newblock \bibinfo{journal}{\emph{arXiv preprint arXiv:2306.04563}}
  (\bibinfo{year}{2023}).
\newblock


\bibitem[Ji et~al\mbox{.}(2023)]%
        {ji2023beavertails}
\bibfield{author}{\bibinfo{person}{Jiaming Ji}, \bibinfo{person}{Mickel Liu},
  \bibinfo{person}{Juntao Dai}, \bibinfo{person}{Xuehai Pan},
  \bibinfo{person}{Chi Zhang}, \bibinfo{person}{Ce Bian},
  \bibinfo{person}{Ruiyang Sun}, \bibinfo{person}{Yizhou Wang}, {and}
  \bibinfo{person}{Yaodong Yang}.} \bibinfo{year}{2023}\natexlab{}.
\newblock \showarticletitle{Beavertails: Towards improved safety alignment of
  llm via a human-preference dataset}.
\newblock \bibinfo{journal}{\emph{arXiv preprint arXiv:2307.04657}}
  (\bibinfo{year}{2023}).
\newblock


\bibitem[Jiang et~al\mbox{.}(2023)]%
        {jiang2023structgpt}
\bibfield{author}{\bibinfo{person}{Jinhao Jiang}, \bibinfo{person}{Kun Zhou},
  \bibinfo{person}{Zican Dong}, \bibinfo{person}{Keming Ye},
  \bibinfo{person}{Wayne~Xin Zhao}, {and} \bibinfo{person}{Ji-Rong Wen}.}
  \bibinfo{year}{2023}\natexlab{}.
\newblock \showarticletitle{Structgpt: A general framework for large language
  model to reason over structured data}.
\newblock \bibinfo{journal}{\emph{arXiv preprint arXiv:2305.09645}}
  (\bibinfo{year}{2023}).
\newblock


\bibitem[Johnson et~al\mbox{.}(2023)]%
        {johnson2023assessing}
\bibfield{author}{\bibinfo{person}{Douglas Johnson}, \bibinfo{person}{Rachel
  Goodman}, \bibinfo{person}{J Patrinely}, \bibinfo{person}{Cosby Stone},
  \bibinfo{person}{Eli Zimmerman}, \bibinfo{person}{Rebecca Donald},
  \bibinfo{person}{Sam Chang}, \bibinfo{person}{Sean Berkowitz},
  \bibinfo{person}{Avni Finn}, \bibinfo{person}{Eiman Jahangir},
  {et~al\mbox{.}}} \bibinfo{year}{2023}\natexlab{}.
\newblock \showarticletitle{Assessing the accuracy and reliability of
  AI-generated medical responses: an evaluation of the Chat-GPT model}.
\newblock  (\bibinfo{year}{2023}).
\newblock


\bibitem[Joshi et~al\mbox{.}(2017)]%
        {TQ}
\bibfield{author}{\bibinfo{person}{Mandar Joshi}, \bibinfo{person}{Eunsol
  Choi}, \bibinfo{person}{Daniel~S. Weld}, {and} \bibinfo{person}{Luke
  Zettlemoyer}.} \bibinfo{year}{2017}\natexlab{}.
\newblock \showarticletitle{TriviaQA: A Large Scale Distantly Supervised
  Challenge Dataset for Reading Comprehension}. In
  \bibinfo{booktitle}{\emph{Proceedings of the 55th Annual Meeting of the
  Association for Computational Linguistics}}. \bibinfo{publisher}{Association
  for Computational Linguistics}, \bibinfo{address}{Vancouver, Canada}.
\newblock


\bibitem[Kadavath et~al\mbox{.}(2022)]%
        {Kadavath2022LanguageM}
\bibfield{author}{\bibinfo{person}{Saurav Kadavath}, \bibinfo{person}{Tom
  Conerly}, \bibinfo{person}{Amanda Askell}, \bibinfo{person}{T.~J. Henighan},
  \bibinfo{person}{Dawn Drain}, \bibinfo{person}{Ethan Perez},
  \bibinfo{person}{Nicholas Schiefer}, \bibinfo{person}{Zachary Dodds},
  \bibinfo{person}{Nova DasSarma}, \bibinfo{person}{Eli Tran-Johnson},
  \bibinfo{person}{Scott Johnston}, \bibinfo{person}{Sheer El-Showk},
  \bibinfo{person}{Andy Jones}, \bibinfo{person}{Nelson Elhage},
  \bibinfo{person}{Tristan Hume}, \bibinfo{person}{Anna Chen},
  \bibinfo{person}{Yuntao Bai}, \bibinfo{person}{Sam Bowman},
  \bibinfo{person}{Stanislav Fort}, \bibinfo{person}{Deep Ganguli},
  \bibinfo{person}{Danny Hernandez}, \bibinfo{person}{Josh Jacobson},
  \bibinfo{person}{John Kernion}, \bibinfo{person}{Shauna Kravec},
  \bibinfo{person}{Liane Lovitt}, \bibinfo{person}{Kamal Ndousse},
  \bibinfo{person}{Catherine Olsson}, \bibinfo{person}{Sam Ringer},
  \bibinfo{person}{Dario Amodei}, \bibinfo{person}{Tom~B. Brown},
  \bibinfo{person}{Jack Clark}, \bibinfo{person}{Nicholas Joseph},
  \bibinfo{person}{Benjamin Mann}, \bibinfo{person}{Sam McCandlish},
  \bibinfo{person}{Christopher Olah}, {and} \bibinfo{person}{Jared Kaplan}.}
  \bibinfo{year}{2022}\natexlab{}.
\newblock \showarticletitle{Language Models (Mostly) Know What They Know}.
\newblock \bibinfo{journal}{\emph{ArXiv}}  \bibinfo{volume}{abs/2207.05221}
  (\bibinfo{year}{2022}).
\newblock


\bibitem[Karpas et~al\mbox{.}(2022)]%
        {karpas2022mrkl}
\bibfield{author}{\bibinfo{person}{Ehud Karpas}, \bibinfo{person}{Omri Abend},
  \bibinfo{person}{Yonatan Belinkov}, \bibinfo{person}{Barak Lenz},
  \bibinfo{person}{Opher Lieber}, \bibinfo{person}{Nir Ratner},
  \bibinfo{person}{Yoav Shoham}, \bibinfo{person}{Hofit Bata},
  \bibinfo{person}{Yoav Levine}, \bibinfo{person}{Kevin Leyton-Brown},
  {et~al\mbox{.}}} \bibinfo{year}{2022}\natexlab{}.
\newblock \showarticletitle{MRKL Systems: A modular, neuro-symbolic
  architecture that combines large language models, external knowledge sources
  and discrete reasoning}.
\newblock \bibinfo{journal}{\emph{arXiv preprint arXiv:2205.00445}}
  (\bibinfo{year}{2022}).
\newblock


\bibitem[Kasneci et~al\mbox{.}(2023)]%
        {kasneci2023chatgpt}
\bibfield{author}{\bibinfo{person}{Enkelejda Kasneci}, \bibinfo{person}{Kathrin
  Se{\ss}ler}, \bibinfo{person}{Stefan K{\"u}chemann}, \bibinfo{person}{Maria
  Bannert}, \bibinfo{person}{Daryna Dementieva}, \bibinfo{person}{Frank
  Fischer}, \bibinfo{person}{Urs Gasser}, \bibinfo{person}{Georg Groh},
  \bibinfo{person}{Stephan G{\"u}nnemann}, \bibinfo{person}{Eyke
  H{\"u}llermeier}, {et~al\mbox{.}}} \bibinfo{year}{2023}\natexlab{}.
\newblock \showarticletitle{ChatGPT for good? On opportunities and challenges
  of large language models for education}.
\newblock \bibinfo{journal}{\emph{Learning and Individual Differences}}
  \bibinfo{volume}{103} (\bibinfo{year}{2023}), \bibinfo{pages}{102274}.
\newblock


\bibitem[Khalfa(1994)]%
        {khalfa1994intelligence}
\bibfield{author}{\bibinfo{person}{Jean Khalfa}.}
  \bibinfo{year}{1994}\natexlab{}.
\newblock \showarticletitle{What is intelligence?}
\newblock  (\bibinfo{year}{1994}).
\newblock


\bibitem[Khan et~al\mbox{.}(2023)]%
        {khan2023covllm}
\bibfield{author}{\bibinfo{person}{Yousuf~A Khan}, \bibinfo{person}{Clarisse
  Hokia}, \bibinfo{person}{Jennifer Xu}, {and} \bibinfo{person}{Ben Ehlert}.}
  \bibinfo{year}{2023}\natexlab{}.
\newblock \showarticletitle{covLLM: Large Language Models for COVID-19
  Biomedical Literature}.
\newblock \bibinfo{journal}{\emph{arXiv preprint arXiv:2306.04926}}
  (\bibinfo{year}{2023}).
\newblock


\bibitem[Kiela et~al\mbox{.}(2021)]%
        {kiela2021dynabench}
\bibfield{author}{\bibinfo{person}{Douwe Kiela}, \bibinfo{person}{Max Bartolo},
  \bibinfo{person}{Yixin Nie}, \bibinfo{person}{Divyansh Kaushik},
  \bibinfo{person}{Atticus Geiger}, \bibinfo{person}{Zhengxuan Wu},
  \bibinfo{person}{Bertie Vidgen}, \bibinfo{person}{Grusha Prasad},
  \bibinfo{person}{Amanpreet Singh}, \bibinfo{person}{Pratik Ringshia},
  {et~al\mbox{.}}} \bibinfo{year}{2021}\natexlab{}.
\newblock \showarticletitle{Dynabench: Rethinking benchmarking in NLP}.
\newblock \bibinfo{journal}{\emph{arXiv preprint arXiv:2104.14337}}
  (\bibinfo{year}{2021}).
\newblock


\bibitem[Kohavi et~al\mbox{.}(1995)]%
        {kohavi1995study}
\bibfield{author}{\bibinfo{person}{Ron Kohavi} {et~al\mbox{.}}}
  \bibinfo{year}{1995}\natexlab{}.
\newblock \showarticletitle{A study of cross-validation and bootstrap for
  accuracy estimation and model selection}. In
  \bibinfo{booktitle}{\emph{Ijcai}}, Vol.~\bibinfo{volume}{14}. Montreal,
  Canada, \bibinfo{pages}{1137--1145}.
\newblock


\bibitem[Kombrink et~al\mbox{.}(2011)]%
        {kombrink2011recurrent}
\bibfield{author}{\bibinfo{person}{Stefan Kombrink}, \bibinfo{person}{Tomas
  Mikolov}, \bibinfo{person}{Martin Karafi{\'a}t}, {and}
  \bibinfo{person}{Luk{\'a}s Burget}.} \bibinfo{year}{2011}\natexlab{}.
\newblock \showarticletitle{Recurrent Neural Network Based Language Modeling in
  Meeting Recognition.}. In \bibinfo{booktitle}{\emph{Interspeech}},
  Vol.~\bibinfo{volume}{11}. \bibinfo{pages}{2877--2880}.
\newblock


\bibitem[Kung et~al\mbox{.}(2023)]%
        {kung2023performance}
\bibfield{author}{\bibinfo{person}{Tiffany~H Kung}, \bibinfo{person}{Morgan
  Cheatham}, \bibinfo{person}{Arielle Medenilla}, \bibinfo{person}{Czarina
  Sillos}, \bibinfo{person}{Lorie De~Leon}, \bibinfo{person}{Camille
  Elepa{\~n}o}, \bibinfo{person}{Maria Madriaga}, \bibinfo{person}{Rimel
  Aggabao}, \bibinfo{person}{Giezel Diaz-Candido}, \bibinfo{person}{James
  Maningo}, {et~al\mbox{.}}} \bibinfo{year}{2023}\natexlab{}.
\newblock \showarticletitle{Performance of ChatGPT on USMLE: Potential for
  AI-assisted medical education using large language models}.
\newblock \bibinfo{journal}{\emph{PLoS digital health}} \bibinfo{volume}{2},
  \bibinfo{number}{2} (\bibinfo{year}{2023}), \bibinfo{pages}{e0000198}.
\newblock


\bibitem[Kwiatkowski et~al\mbox{.}(2019)]%
        {NQ}
\bibfield{author}{\bibinfo{person}{Tom Kwiatkowski},
  \bibinfo{person}{Jennimaria Palomaki}, \bibinfo{person}{Olivia Redfield},
  \bibinfo{person}{Michael Collins}, \bibinfo{person}{Ankur Parikh},
  \bibinfo{person}{Chris Alberti}, \bibinfo{person}{Danielle Epstein},
  \bibinfo{person}{Illia Polosukhin}, \bibinfo{person}{Matthew Kelcey},
  \bibinfo{person}{Jacob Devlin}, \bibinfo{person}{Kenton Lee},
  \bibinfo{person}{Kristina~N. Toutanova}, \bibinfo{person}{Llion Jones},
  \bibinfo{person}{Ming-Wei Chang}, \bibinfo{person}{Andrew Dai},
  \bibinfo{person}{Jakob Uszkoreit}, \bibinfo{person}{Quoc Le}, {and}
  \bibinfo{person}{Slav Petrov}.} \bibinfo{year}{2019}\natexlab{}.
\newblock \showarticletitle{Natural Questions: a Benchmark for Question
  Answering Research}.
\newblock \bibinfo{journal}{\emph{Transactions of the Association of
  Computational Linguistics}} (\bibinfo{year}{2019}).
\newblock


\bibitem[Lahat et~al\mbox{.}(2023)]%
        {lahat2023evaluating}
\bibfield{author}{\bibinfo{person}{Adi Lahat}, \bibinfo{person}{Eyal Shachar},
  \bibinfo{person}{Benjamin Avidan}, \bibinfo{person}{Zina Shatz},
  \bibinfo{person}{Benjamin~S Glicksberg}, {and} \bibinfo{person}{Eyal Klang}.}
  \bibinfo{year}{2023}\natexlab{}.
\newblock \showarticletitle{Evaluating the use of large language model in
  identifying top research questions in gastroenterology}.
\newblock \bibinfo{journal}{\emph{Scientific reports}} \bibinfo{volume}{13},
  \bibinfo{number}{1} (\bibinfo{year}{2023}), \bibinfo{pages}{4164}.
\newblock


\bibitem[Lai et~al\mbox{.}(2023)]%
        {lai2023chatgpt}
\bibfield{author}{\bibinfo{person}{Viet~Dac Lai}, \bibinfo{person}{Nghia~Trung
  Ngo}, \bibinfo{person}{Amir Pouran~Ben Veyseh}, \bibinfo{person}{Hieu Man},
  \bibinfo{person}{Franck Dernoncourt}, \bibinfo{person}{Trung Bui}, {and}
  \bibinfo{person}{Thien~Huu Nguyen}.} \bibinfo{year}{2023}\natexlab{}.
\newblock \showarticletitle{ChatGPT Beyond English: Towards a Comprehensive
  Evaluation of Large Language Models in Multilingual Learning}.
\newblock \bibinfo{journal}{\emph{arXiv preprint arXiv:2304.05613}}
  (\bibinfo{year}{2023}).
\newblock


\bibitem[Lanzi and Loiacono(2023)]%
        {lanzi2023chatgpt}
\bibfield{author}{\bibinfo{person}{Pier~Luca Lanzi} {and}
  \bibinfo{person}{Daniele Loiacono}.} \bibinfo{year}{2023}\natexlab{}.
\newblock \showarticletitle{Chatgpt and other large language models as
  evolutionary engines for online interactive collaborative game design}.
\newblock \bibinfo{journal}{\emph{arXiv preprint arXiv:2303.02155}}
  (\bibinfo{year}{2023}).
\newblock


\bibitem[Laskar et~al\mbox{.}(2023)]%
        {laskar2023systematic}
\bibfield{author}{\bibinfo{person}{Md~Tahmid~Rahman Laskar},
  \bibinfo{person}{M~Saiful Bari}, \bibinfo{person}{Mizanur Rahman},
  \bibinfo{person}{Md~Amran~Hossen Bhuiyan}, \bibinfo{person}{Shafiq Joty},
  {and} \bibinfo{person}{Jimmy~Xiangji Huang}.}
  \bibinfo{year}{2023}\natexlab{}.
\newblock \showarticletitle{A Systematic Study and Comprehensive Evaluation of
  ChatGPT on Benchmark Datasets}.
\newblock \bibinfo{journal}{\emph{arXiv preprint arXiv:2305.18486}}
  (\bibinfo{year}{2023}).
\newblock


\bibitem[Le and Zhang(2023)]%
        {le2023evaluation}
\bibfield{author}{\bibinfo{person}{Van-Hoang Le} {and} \bibinfo{person}{Hongyu
  Zhang}.} \bibinfo{year}{2023}\natexlab{}.
\newblock \showarticletitle{An Evaluation of Log Parsing with ChatGPT}.
\newblock \bibinfo{journal}{\emph{arXiv preprint arXiv:2306.01590}}
  (\bibinfo{year}{2023}).
\newblock


\bibitem[LeCun et~al\mbox{.}(2015)]%
        {lecun2015deep}
\bibfield{author}{\bibinfo{person}{Yann LeCun}, \bibinfo{person}{Yoshua
  Bengio}, {and} \bibinfo{person}{Geoffrey Hinton}.}
  \bibinfo{year}{2015}\natexlab{}.
\newblock \showarticletitle{Deep learning}.
\newblock \bibinfo{journal}{\emph{nature}} \bibinfo{volume}{521},
  \bibinfo{number}{7553} (\bibinfo{year}{2015}), \bibinfo{pages}{436--444}.
\newblock


\bibitem[Lee et~al\mbox{.}(2023)]%
        {lee2023can}
\bibfield{author}{\bibinfo{person}{Noah Lee}, \bibinfo{person}{Na~Min An},
  {and} \bibinfo{person}{James Thorne}.} \bibinfo{year}{2023}\natexlab{}.
\newblock \showarticletitle{Can Large Language Models Infer and Disagree Like
  Humans?}
\newblock \bibinfo{journal}{\emph{arXiv preprint arXiv:2305.13788}}
  (\bibinfo{year}{2023}).
\newblock


\bibitem[Lewis et~al\mbox{.}(2019)]%
        {lewis2019bart}
\bibfield{author}{\bibinfo{person}{Mike Lewis}, \bibinfo{person}{Yinhan Liu},
  \bibinfo{person}{Naman Goyal}, \bibinfo{person}{Marjan Ghazvininejad},
  \bibinfo{person}{Abdelrahman Mohamed}, \bibinfo{person}{Omer Levy},
  \bibinfo{person}{Ves Stoyanov}, {and} \bibinfo{person}{Luke Zettlemoyer}.}
  \bibinfo{year}{2019}\natexlab{}.
\newblock \showarticletitle{Bart: Denoising sequence-to-sequence pre-training
  for natural language generation, translation, and comprehension}.
\newblock \bibinfo{journal}{\emph{arXiv preprint arXiv:1910.13461}}
  (\bibinfo{year}{2019}).
\newblock


\bibitem[Li et~al\mbox{.}(2023e)]%
        {li2023seed}
\bibfield{author}{\bibinfo{person}{Bohao Li}, \bibinfo{person}{Rui Wang},
  \bibinfo{person}{Guangzhi Wang}, \bibinfo{person}{Yuying Ge},
  \bibinfo{person}{Yixiao Ge}, {and} \bibinfo{person}{Ying Shan}.}
  \bibinfo{year}{2023}\natexlab{e}.
\newblock \showarticletitle{Seed-bench: Benchmarking multimodal llms with
  generative comprehension}.
\newblock \bibinfo{journal}{\emph{arXiv preprint arXiv:2307.16125}}
  (\bibinfo{year}{2023}).
\newblock


\bibitem[Li et~al\mbox{.}(2023g)]%
        {li2023cmmlu}
\bibfield{author}{\bibinfo{person}{Haonan Li}, \bibinfo{person}{Yixuan Zhang},
  \bibinfo{person}{Fajri Koto}, \bibinfo{person}{Yifei Yang},
  \bibinfo{person}{Hai Zhao}, \bibinfo{person}{Yeyun Gong},
  \bibinfo{person}{Nan Duan}, {and} \bibinfo{person}{Timothy Baldwin}.}
  \bibinfo{year}{2023}\natexlab{g}.
\newblock \showarticletitle{CMMLU: Measuring massive multitask language
  understanding in Chinese}.
\newblock \bibinfo{journal}{\emph{arXiv preprint arXiv:2306.09212}}
  (\bibinfo{year}{2023}).
\newblock


\bibitem[Li et~al\mbox{.}(2023d)]%
        {li2023apibank}
\bibfield{author}{\bibinfo{person}{Minghao Li}, \bibinfo{person}{Feifan Song},
  \bibinfo{person}{Bowen Yu}, \bibinfo{person}{Haiyang Yu},
  \bibinfo{person}{Zhoujun Li}, \bibinfo{person}{Fei Huang}, {and}
  \bibinfo{person}{Yongbin Li}.} \bibinfo{year}{2023}\natexlab{d}.
\newblock \bibinfo{title}{API-Bank: A Benchmark for Tool-Augmented LLMs}.
\newblock
\newblock
\showeprint[arxiv]{2304.08244}~[cs.CL]


\bibitem[Li et~al\mbox{.}(2023a)]%
        {li2023exploring}
\bibfield{author}{\bibinfo{person}{Ruyu Li}, \bibinfo{person}{Wenhao Deng},
  \bibinfo{person}{Yu Cheng}, \bibinfo{person}{Zheng Yuan},
  \bibinfo{person}{Jiaqi Zhang}, {and} \bibinfo{person}{Fajie Yuan}.}
  \bibinfo{year}{2023}\natexlab{a}.
\newblock \showarticletitle{Exploring the Upper Limits of Text-Based
  Collaborative Filtering Using Large Language Models: Discoveries and
  Insights}.
\newblock \bibinfo{journal}{\emph{arXiv preprint arXiv:2305.11700}}
  (\bibinfo{year}{2023}).
\newblock


\bibitem[Li et~al\mbox{.}(2023c)]%
        {li2023survey}
\bibfield{author}{\bibinfo{person}{Xinzhe Li}, \bibinfo{person}{Ming Liu},
  \bibinfo{person}{Shang Gao}, {and} \bibinfo{person}{Wray Buntine}.}
  \bibinfo{year}{2023}\natexlab{c}.
\newblock \bibinfo{title}{A Survey on Out-of-Distribution Evaluation of Neural
  NLP Models}.
\newblock
\newblock
\showeprint[arxiv]{2306.15261}~[cs.CL]


\bibitem[Li et~al\mbox{.}(2023f)]%
        {alpaca_eval}
\bibfield{author}{\bibinfo{person}{Xuechen Li}, \bibinfo{person}{Tianyi Zhang},
  \bibinfo{person}{Yann Dubois}, \bibinfo{person}{Rohan Taori},
  \bibinfo{person}{Ishaan Gulrajani}, \bibinfo{person}{Carlos Guestrin},
  \bibinfo{person}{Percy Liang}, {and} \bibinfo{person}{Tatsunori~B.
  Hashimoto}.} \bibinfo{year}{2023}\natexlab{f}.
\newblock \bibinfo{title}{AlpacaEval: An Automatic Evaluator of
  Instruction-following Models}.
\newblock
  \bibinfo{howpublished}{\url{https://github.com/tatsu-lab/alpaca_eval}}.
\newblock


\bibitem[Li et~al\mbox{.}(2023b)]%
        {li2023evaluating}
\bibfield{author}{\bibinfo{person}{Yifan Li}, \bibinfo{person}{Yifan Du},
  \bibinfo{person}{Kun Zhou}, \bibinfo{person}{Jinpeng Wang},
  \bibinfo{person}{Wayne~Xin Zhao}, {and} \bibinfo{person}{Ji-Rong Wen}.}
  \bibinfo{year}{2023}\natexlab{b}.
\newblock \showarticletitle{Evaluating object hallucination in large
  vision-language models}.
\newblock \bibinfo{journal}{\emph{arXiv preprint arXiv:2305.10355}}
  (\bibinfo{year}{2023}).
\newblock


\bibitem[Liang et~al\mbox{.}(2022)]%
        {liang2022holistic}
\bibfield{author}{\bibinfo{person}{Percy Liang}, \bibinfo{person}{Rishi
  Bommasani}, \bibinfo{person}{Tony Lee}, \bibinfo{person}{Dimitris Tsipras},
  \bibinfo{person}{Dilara Soylu}, \bibinfo{person}{Michihiro Yasunaga},
  \bibinfo{person}{Yian Zhang}, \bibinfo{person}{Deepak Narayanan},
  \bibinfo{person}{Yuhuai Wu}, \bibinfo{person}{Ananya Kumar}, {et~al\mbox{.}}}
  \bibinfo{year}{2022}\natexlab{}.
\newblock \showarticletitle{Holistic evaluation of language models}.
\newblock \bibinfo{journal}{\emph{arXiv preprint arXiv:2211.09110}}
  (\bibinfo{year}{2022}).
\newblock


\bibitem[Liang et~al\mbox{.}(2023a)]%
        {liang2023leveraging}
\bibfield{author}{\bibinfo{person}{Tian Liang}, \bibinfo{person}{Zhiwei He},
  \bibinfo{person}{Jen-tes Huang}, \bibinfo{person}{Wenxuan Wang},
  \bibinfo{person}{Wenxiang Jiao}, \bibinfo{person}{Rui Wang},
  \bibinfo{person}{Yujiu Yang}, \bibinfo{person}{Zhaopeng Tu},
  \bibinfo{person}{Shuming Shi}, {and} \bibinfo{person}{Xing Wang}.}
  \bibinfo{year}{2023}\natexlab{a}.
\newblock \showarticletitle{Leveraging Word Guessing Games to Assess the
  Intelligence of Large Language Models}.
\newblock \bibinfo{journal}{\emph{arXiv preprint arXiv:2310.20499}}
  (\bibinfo{year}{2023}).
\newblock


\bibitem[Liang et~al\mbox{.}(2023b)]%
        {liang2023uhgeval}
\bibfield{author}{\bibinfo{person}{Xun Liang}, \bibinfo{person}{Shichao Song},
  \bibinfo{person}{Simin Niu}, \bibinfo{person}{Zhiyu Li},
  \bibinfo{person}{Feiyu Xiong}, \bibinfo{person}{Bo Tang},
  \bibinfo{person}{Zhaohui Wy}, \bibinfo{person}{Dawei He},
  \bibinfo{person}{Peng Cheng}, \bibinfo{person}{Zhonghao Wang},
  {et~al\mbox{.}}} \bibinfo{year}{2023}\natexlab{b}.
\newblock \showarticletitle{UHGEval: Benchmarking the Hallucination of Chinese
  Large Language Models via Unconstrained Generation}.
\newblock \bibinfo{journal}{\emph{arXiv preprint arXiv:2311.15296}}
  (\bibinfo{year}{2023}).
\newblock


\bibitem[Li{\'e}vin et~al\mbox{.}(2022)]%
        {lievin2022can}
\bibfield{author}{\bibinfo{person}{Valentin Li{\'e}vin},
  \bibinfo{person}{Christoffer~Egeberg Hother}, {and} \bibinfo{person}{Ole
  Winther}.} \bibinfo{year}{2022}\natexlab{}.
\newblock \showarticletitle{Can large language models reason about medical
  questions?}
\newblock \bibinfo{journal}{\emph{arXiv preprint arXiv:2207.08143}}
  (\bibinfo{year}{2022}).
\newblock


\bibitem[Lin(2004)]%
        {lin-2004-rouge}
\bibfield{author}{\bibinfo{person}{Chin-Yew Lin}.}
  \bibinfo{year}{2004}\natexlab{}.
\newblock \showarticletitle{{ROUGE}: A Package for Automatic Evaluation of
  Summaries}. In \bibinfo{booktitle}{\emph{Text Summarization Branches Out}}.
  \bibinfo{publisher}{Association for Computational Linguistics},
  \bibinfo{address}{Barcelona, Spain}, \bibinfo{pages}{74--81}.
\newblock
\urldef\tempurl%
\url{https://aclanthology.org/W04-1013}
\showURL{%
\tempurl}


\bibitem[Lin et~al\mbox{.}(2021)]%
        {lin2021truthfulqa}
\bibfield{author}{\bibinfo{person}{Stephanie Lin}, \bibinfo{person}{Jacob
  Hilton}, {and} \bibinfo{person}{Owain Evans}.}
  \bibinfo{year}{2021}\natexlab{}.
\newblock \showarticletitle{Truthfulqa: Measuring how models mimic human
  falsehoods}.
\newblock \bibinfo{journal}{\emph{arXiv preprint arXiv:2109.07958}}
  (\bibinfo{year}{2021}).
\newblock


\bibitem[Lin et~al\mbox{.}(2014)]%
        {lin2014microsoft}
\bibfield{author}{\bibinfo{person}{Tsung-Yi Lin}, \bibinfo{person}{Michael
  Maire}, \bibinfo{person}{Serge Belongie}, \bibinfo{person}{James Hays},
  \bibinfo{person}{Pietro Perona}, \bibinfo{person}{Deva Ramanan},
  \bibinfo{person}{Piotr Doll{\'a}r}, {and} \bibinfo{person}{C~Lawrence
  Zitnick}.} \bibinfo{year}{2014}\natexlab{}.
\newblock \showarticletitle{Microsoft coco: Common objects in context}. In
  \bibinfo{booktitle}{\emph{Computer Vision--ECCV 2014: 13th European
  Conference, Zurich, Switzerland, September 6-12, 2014, Proceedings, Part V
  13}}. Springer, \bibinfo{pages}{740--755}.
\newblock


\bibitem[Lin and Chen(2023)]%
        {lin2023llm}
\bibfield{author}{\bibinfo{person}{Yen-Ting Lin} {and}
  \bibinfo{person}{Yun-Nung Chen}.} \bibinfo{year}{2023}\natexlab{}.
\newblock \showarticletitle{LLM-Eval: Unified Multi-Dimensional Automatic
  Evaluation for Open-Domain Conversations with Large Language Models}.
\newblock \bibinfo{journal}{\emph{arXiv preprint arXiv:2305.13711}}
  (\bibinfo{year}{2023}).
\newblock


\bibitem[Liu et~al\mbox{.}(2023c)]%
        {liu2023m3ke}
\bibfield{author}{\bibinfo{person}{Chuang Liu}, \bibinfo{person}{Renren Jin},
  \bibinfo{person}{Yuqi Ren}, \bibinfo{person}{Linhao Yu},
  \bibinfo{person}{Tianyu Dong}, \bibinfo{person}{Xiaohan Peng},
  \bibinfo{person}{Shuting Zhang}, \bibinfo{person}{Jianxiang Peng},
  \bibinfo{person}{Peiyi Zhang}, \bibinfo{person}{Qingqing Lyu},
  \bibinfo{person}{Xiaowen Su}, \bibinfo{person}{Qun Liu}, {and}
  \bibinfo{person}{Deyi Xiong}.} \bibinfo{year}{2023}\natexlab{c}.
\newblock \bibinfo{title}{M3KE: A Massive Multi-Level Multi-Subject Knowledge
  Evaluation Benchmark for Chinese Large Language Models}.
\newblock
\newblock
\showeprint[arxiv]{2305.10263}~[cs.CL]


\bibitem[Liu et~al\mbox{.}(2023d)]%
        {liu2023mitigating}
\bibfield{author}{\bibinfo{person}{Fuxiao Liu}, \bibinfo{person}{Kevin Lin},
  \bibinfo{person}{Linjie Li}, \bibinfo{person}{Jianfeng Wang},
  \bibinfo{person}{Yaser Yacoob}, {and} \bibinfo{person}{Lijuan Wang}.}
  \bibinfo{year}{2023}\natexlab{d}.
\newblock \bibinfo{title}{Mitigating Hallucination in Large Multi-Modal Models
  via Robust Instruction Tuning}.
\newblock
\newblock
\showeprint[arxiv]{2306.14565}~[cs.CV]


\bibitem[Liu et~al\mbox{.}(2023e)]%
        {liu2023evaluating}
\bibfield{author}{\bibinfo{person}{Hanmeng Liu}, \bibinfo{person}{Ruoxi Ning},
  \bibinfo{person}{Zhiyang Teng}, \bibinfo{person}{Jian Liu},
  \bibinfo{person}{Qiji Zhou}, {and} \bibinfo{person}{Yue Zhang}.}
  \bibinfo{year}{2023}\natexlab{e}.
\newblock \bibinfo{title}{Evaluating the Logical Reasoning Ability of ChatGPT
  and GPT-4}.
\newblock
\newblock
\showeprint[arxiv]{2304.03439}~[cs.CL]


\bibitem[Liu et~al\mbox{.}(2023f)]%
        {liu2023your}
\bibfield{author}{\bibinfo{person}{Jiawei Liu}, \bibinfo{person}{Chunqiu~Steven
  Xia}, \bibinfo{person}{Yuyao Wang}, {and} \bibinfo{person}{Lingming Zhang}.}
  \bibinfo{year}{2023}\natexlab{f}.
\newblock \showarticletitle{Is your code generated by chatgpt really correct?
  rigorous evaluation of large language models for code generation}.
\newblock \bibinfo{journal}{\emph{arXiv preprint arXiv:2305.01210}}
  (\bibinfo{year}{2023}).
\newblock


\bibitem[Liu et~al\mbox{.}(2023a)]%
        {liu2023mmbench}
\bibfield{author}{\bibinfo{person}{Yuan Liu}, \bibinfo{person}{Haodong Duan},
  \bibinfo{person}{Yuanhan Zhang}, \bibinfo{person}{Bo Li},
  \bibinfo{person}{Songyang Zhang}, \bibinfo{person}{Wangbo Zhao},
  \bibinfo{person}{Yike Yuan}, \bibinfo{person}{Jiaqi Wang},
  \bibinfo{person}{Conghui He}, \bibinfo{person}{Ziwei Liu},
  \bibinfo{person}{Kai Chen}, {and} \bibinfo{person}{Dahua Lin}.}
  \bibinfo{year}{2023}\natexlab{a}.
\newblock \bibinfo{title}{MMBench: Is Your Multi-modal Model an All-around
  Player?}
\newblock
\newblock
\showeprint[arxiv]{2307.06281}~[cs.CV]


\bibitem[Liu et~al\mbox{.}(2023b)]%
        {liu2023summary}
\bibfield{author}{\bibinfo{person}{Yiheng Liu}, \bibinfo{person}{Tianle Han},
  \bibinfo{person}{Siyuan Ma}, \bibinfo{person}{Jiayue Zhang},
  \bibinfo{person}{Yuanyuan Yang}, \bibinfo{person}{Jiaming Tian},
  \bibinfo{person}{Hao He}, \bibinfo{person}{Antong Li},
  \bibinfo{person}{Mengshen He}, \bibinfo{person}{Zhengliang Liu},
  {et~al\mbox{.}}} \bibinfo{year}{2023}\natexlab{b}.
\newblock \showarticletitle{Summary of chatgpt/gpt-4 research and perspective
  towards the future of large language models}.
\newblock \bibinfo{journal}{\emph{arXiv preprint arXiv:2304.01852}}
  (\bibinfo{year}{2023}).
\newblock


\bibitem[LMSYS(2023)]%
        {chatbotarena}
\bibfield{author}{\bibinfo{person}{LMSYS}.} \bibinfo{year}{2023}\natexlab{}.
\newblock \bibinfo{title}{Chatbot Arena: Benchmarking LLMs in the Wild with Elo
  Ratings}.
\newblock \bibinfo{howpublished}{\url{https://lmsys.org}}.
\newblock


\bibitem[Lopez-Lira and Tang(2023)]%
        {lopez2023can}
\bibfield{author}{\bibinfo{person}{Alejandro Lopez-Lira} {and}
  \bibinfo{person}{Yuehua Tang}.} \bibinfo{year}{2023}\natexlab{}.
\newblock \showarticletitle{Can chatgpt forecast stock price movements? Return
  predictability and large language models}.
\newblock \bibinfo{journal}{\emph{arXiv preprint arXiv:2304.07619}}
  (\bibinfo{year}{2023}).
\newblock


\bibitem[Lyu et~al\mbox{.}(2023b)]%
        {lyu2023new}
\bibfield{author}{\bibinfo{person}{Chenyang Lyu}, \bibinfo{person}{Jitao Xu},
  {and} \bibinfo{person}{Longyue Wang}.} \bibinfo{year}{2023}\natexlab{b}.
\newblock \showarticletitle{New trends in machine translation using large
  language models: Case examples with chatgpt}.
\newblock \bibinfo{journal}{\emph{arXiv preprint arXiv:2305.01181}}
  (\bibinfo{year}{2023}).
\newblock


\bibitem[Lyu et~al\mbox{.}(2023a)]%
        {lyu2023translating}
\bibfield{author}{\bibinfo{person}{Qing Lyu}, \bibinfo{person}{Josh Tan},
  \bibinfo{person}{Mike~E Zapadka}, \bibinfo{person}{Janardhana Ponnatapuram},
  \bibinfo{person}{Chuang Niu}, \bibinfo{person}{Ge Wang}, {and}
  \bibinfo{person}{Christopher~T Whitlow}.} \bibinfo{year}{2023}\natexlab{a}.
\newblock \showarticletitle{Translating radiology reports into plain language
  using chatgpt and gpt-4 with prompt learning: Promising results, limitations,
  and potential}.
\newblock \bibinfo{journal}{\emph{arXiv preprint arXiv:2303.09038}}
  (\bibinfo{year}{2023}).
\newblock


\bibitem[Ma et~al\mbox{.}(2021)]%
        {ma2021dynaboard}
\bibfield{author}{\bibinfo{person}{Zhiyi Ma}, \bibinfo{person}{Kawin
  Ethayarajh}, \bibinfo{person}{Tristan Thrush}, \bibinfo{person}{Somya Jain},
  \bibinfo{person}{Ledell Wu}, \bibinfo{person}{Robin Jia},
  \bibinfo{person}{Christopher Potts}, \bibinfo{person}{Adina Williams}, {and}
  \bibinfo{person}{Douwe Kiela}.} \bibinfo{year}{2021}\natexlab{}.
\newblock \showarticletitle{Dynaboard: An evaluation-as-a-service platform for
  holistic next-generation benchmarking}.
\newblock \bibinfo{journal}{\emph{Advances in Neural Information Processing
  Systems}}  \bibinfo{volume}{34} (\bibinfo{year}{2021}),
  \bibinfo{pages}{10351--10367}.
\newblock


\bibitem[Manakul et~al\mbox{.}(2023a)]%
        {manakul2023selfcheckgpt}
\bibfield{author}{\bibinfo{person}{Potsawee Manakul}, \bibinfo{person}{Adian
  Liusie}, {and} \bibinfo{person}{Mark~JF Gales}.}
  \bibinfo{year}{2023}\natexlab{a}.
\newblock \showarticletitle{Selfcheckgpt: Zero-resource black-box hallucination
  detection for generative large language models}.
\newblock \bibinfo{journal}{\emph{arXiv preprint arXiv:2303.08896}}
  (\bibinfo{year}{2023}).
\newblock


\bibitem[Manakul et~al\mbox{.}(2023b)]%
        {manakul2023mqag}
\bibfield{author}{\bibinfo{person}{Potsawee Manakul}, \bibinfo{person}{Adian
  Liusie}, {and} \bibinfo{person}{Mark J.~F. Gales}.}
  \bibinfo{year}{2023}\natexlab{b}.
\newblock \bibinfo{title}{MQAG: Multiple-choice Question Answering and
  Generation for Assessing Information Consistency in Summarization}.
\newblock
\newblock
\showeprint[arxiv]{2301.12307}~[cs.CL]


\bibitem[Margatina et~al\mbox{.}(2023)]%
        {margatina2023dynamic}
\bibfield{author}{\bibinfo{person}{Katerina Margatina}, \bibinfo{person}{Shuai
  Wang}, \bibinfo{person}{Yogarshi Vyas}, \bibinfo{person}{Neha~Anna John},
  \bibinfo{person}{Yassine Benajiba}, {and} \bibinfo{person}{Miguel
  Ballesteros}.} \bibinfo{year}{2023}\natexlab{}.
\newblock \showarticletitle{Dynamic benchmarking of masked language models on
  temporal concept drift with multiple views}.
\newblock \bibinfo{journal}{\emph{arXiv preprint arXiv:2302.12297}}
  (\bibinfo{year}{2023}).
\newblock


\bibitem[McCarthy(2007)]%
        {mccarthy2007artificial}
\bibfield{author}{\bibinfo{person}{John McCarthy}.}
  \bibinfo{year}{2007}\natexlab{}.
\newblock \showarticletitle{What is artificial intelligence}.
\newblock  (\bibinfo{year}{2007}).
\newblock


\bibitem[Microsoft(2023)]%
        {BingChat}
\bibfield{author}{\bibinfo{person}{Microsoft}.}
  \bibinfo{year}{2023}\natexlab{}.
\newblock \showarticletitle{Bing Chat}.
\newblock \bibinfo{journal}{\emph{https://www.bing.com/new}}
  (\bibinfo{year}{2023}).
\newblock


\bibitem[Min et~al\mbox{.}(2023)]%
        {min2023factscore}
\bibfield{author}{\bibinfo{person}{Sewon Min}, \bibinfo{person}{Kalpesh
  Krishna}, \bibinfo{person}{Xinxi Lyu}, \bibinfo{person}{Mike Lewis},
  \bibinfo{person}{Wen-tau Yih}, \bibinfo{person}{Pang~Wei Koh},
  \bibinfo{person}{Mohit Iyyer}, \bibinfo{person}{Luke Zettlemoyer}, {and}
  \bibinfo{person}{Hannaneh Hajishirzi}.} \bibinfo{year}{2023}\natexlab{}.
\newblock \showarticletitle{FActScore: Fine-grained Atomic Evaluation of
  Factual Precision in Long Form Text Generation}.
\newblock \bibinfo{journal}{\emph{arXiv preprint arXiv:2305.14251}}
  (\bibinfo{year}{2023}).
\newblock


\bibitem[Nay et~al\mbox{.}(2023)]%
        {nay2023large}
\bibfield{author}{\bibinfo{person}{John~J Nay}, \bibinfo{person}{David
  Karamardian}, \bibinfo{person}{Sarah~B Lawsky}, \bibinfo{person}{Wenting
  Tao}, \bibinfo{person}{Meghana Bhat}, \bibinfo{person}{Raghav Jain},
  \bibinfo{person}{Aaron~Travis Lee}, \bibinfo{person}{Jonathan~H Choi}, {and}
  \bibinfo{person}{Jungo Kasai}.} \bibinfo{year}{2023}\natexlab{}.
\newblock \showarticletitle{Large Language Models as Tax Attorneys: A Case
  Study in Legal Capabilities Emergence}.
\newblock \bibinfo{journal}{\emph{arXiv preprint arXiv:2306.07075}}
  (\bibinfo{year}{2023}).
\newblock


\bibitem[Nie et~al\mbox{.}(2019)]%
        {nie2019adversarial}
\bibfield{author}{\bibinfo{person}{Yixin Nie}, \bibinfo{person}{Adina
  Williams}, \bibinfo{person}{Emily Dinan}, \bibinfo{person}{Mohit Bansal},
  \bibinfo{person}{Jason Weston}, {and} \bibinfo{person}{Douwe Kiela}.}
  \bibinfo{year}{2019}\natexlab{}.
\newblock \showarticletitle{Adversarial NLI: A new benchmark for natural
  language understanding}.
\newblock \bibinfo{journal}{\emph{arXiv preprint arXiv:1910.14599}}
  (\bibinfo{year}{2019}).
\newblock


\bibitem[Nijkamp et~al\mbox{.}(2022)]%
        {nijkamp2022codegen}
\bibfield{author}{\bibinfo{person}{Erik Nijkamp}, \bibinfo{person}{Bo Pang},
  \bibinfo{person}{Hiroaki Hayashi}, \bibinfo{person}{Lifu Tu},
  \bibinfo{person}{Huan Wang}, \bibinfo{person}{Yingbo Zhou},
  \bibinfo{person}{Silvio Savarese}, {and} \bibinfo{person}{Caiming Xiong}.}
  \bibinfo{year}{2022}\natexlab{}.
\newblock \showarticletitle{Codegen: An open large language model for code with
  multi-turn program synthesis}.
\newblock \bibinfo{journal}{\emph{arXiv preprint arXiv:2203.13474}}
  (\bibinfo{year}{2022}).
\newblock


\bibitem[Novikova et~al\mbox{.}(2017)]%
        {novikova2017we}
\bibfield{author}{\bibinfo{person}{Jekaterina Novikova},
  \bibinfo{person}{Ond{\v{r}}ej Du{\v{s}}ek}, \bibinfo{person}{Amanda~Cercas
  Curry}, {and} \bibinfo{person}{Verena Rieser}.}
  \bibinfo{year}{2017}\natexlab{}.
\newblock \showarticletitle{Why we need new evaluation metrics for NLG}.
\newblock \bibinfo{journal}{\emph{arXiv preprint arXiv:1707.06875}}
  (\bibinfo{year}{2017}).
\newblock


\bibitem[Oh et~al\mbox{.}(2023)]%
        {oh2023chatgpt}
\bibfield{author}{\bibinfo{person}{Namkee Oh}, \bibinfo{person}{Gyu-Seong
  Choi}, {and} \bibinfo{person}{Woo~Yong Lee}.}
  \bibinfo{year}{2023}\natexlab{}.
\newblock \showarticletitle{ChatGPT goes to the operating room: evaluating
  GPT-4 performance and its potential in surgical education and training in the
  era of large language models}.
\newblock \bibinfo{journal}{\emph{Annals of Surgical Treatment and Research}}
  \bibinfo{volume}{104}, \bibinfo{number}{5} (\bibinfo{year}{2023}),
  \bibinfo{pages}{269}.
\newblock


\bibitem[Olney(2023)]%
        {olney2023generating}
\bibfield{author}{\bibinfo{person}{Andrew~M Olney}.}
  \bibinfo{year}{2023}\natexlab{}.
\newblock \showarticletitle{Generating multiple choice questions from a
  textbook: Llms match human performance on most metrics}. In
  \bibinfo{booktitle}{\emph{AIED Workshops}}.
\newblock


\bibitem[OpenAI(2023a)]%
        {chatgpt}
\bibfield{author}{\bibinfo{person}{OpenAI}.} \bibinfo{year}{2023}\natexlab{a}.
\newblock \bibinfo{howpublished}{\url{https://chat.openai.com.chat}}.
\newblock


\bibitem[OpenAI(2023b)]%
        {openai2023gpt4}
\bibfield{author}{\bibinfo{person}{OpenAI}.} \bibinfo{year}{2023}\natexlab{b}.
\newblock \bibinfo{title}{GPT-4 Technical Report}.
\newblock
\newblock
\showeprint[arxiv]{2303.08774}~[cs.CL]


\bibitem[Orr{\`u} et~al\mbox{.}(2023)]%
        {orru2023human}
\bibfield{author}{\bibinfo{person}{Graziella Orr{\`u}}, \bibinfo{person}{Andrea
  Piarulli}, \bibinfo{person}{Ciro Conversano}, {and} \bibinfo{person}{Angelo
  Gemignani}.} \bibinfo{year}{2023}\natexlab{}.
\newblock \showarticletitle{Human-like problem-solving abilities in large
  language models using ChatGPT}.
\newblock \bibinfo{journal}{\emph{Frontiers in Artificial Intelligence}}
  \bibinfo{volume}{6} (\bibinfo{year}{2023}).
\newblock


\bibitem[Ott et~al\mbox{.}(2023)]%
        {ott2023thoughtsource}
\bibfield{author}{\bibinfo{person}{Simon Ott}, \bibinfo{person}{Konstantin
  Hebenstreit}, \bibinfo{person}{Valentin Li{\'e}vin},
  \bibinfo{person}{Christoffer~Egeberg Hother}, \bibinfo{person}{Milad Moradi},
  \bibinfo{person}{Maximilian Mayrhauser}, \bibinfo{person}{Robert Praas},
  \bibinfo{person}{Ole Winther}, {and} \bibinfo{person}{Matthias Samwald}.}
  \bibinfo{year}{2023}\natexlab{}.
\newblock \showarticletitle{ThoughtSource: A central hub for large language
  model reasoning data}.
\newblock \bibinfo{journal}{\emph{arXiv preprint arXiv:2301.11596}}
  (\bibinfo{year}{2023}).
\newblock


\bibitem[Ouyang et~al\mbox{.}(2022)]%
        {ouyang2022training}
\bibfield{author}{\bibinfo{person}{Long Ouyang}, \bibinfo{person}{Jeffrey Wu},
  \bibinfo{person}{Xu Jiang}, \bibinfo{person}{Diogo Almeida},
  \bibinfo{person}{Carroll Wainwright}, \bibinfo{person}{Pamela Mishkin},
  \bibinfo{person}{Chong Zhang}, \bibinfo{person}{Sandhini Agarwal},
  \bibinfo{person}{Katarina Slama}, \bibinfo{person}{Alex Ray},
  {et~al\mbox{.}}} \bibinfo{year}{2022}\natexlab{}.
\newblock \showarticletitle{Training language models to follow instructions
  with human feedback}.
\newblock \bibinfo{journal}{\emph{Advances in Neural Information Processing
  Systems}}  \bibinfo{volume}{35} (\bibinfo{year}{2022}),
  \bibinfo{pages}{27730--27744}.
\newblock


\bibitem[Pallagani et~al\mbox{.}(2023)]%
        {pallagani2023understanding}
\bibfield{author}{\bibinfo{person}{Vishal Pallagani}, \bibinfo{person}{Bharath
  Muppasani}, \bibinfo{person}{Keerthiram Murugesan},
  \bibinfo{person}{Francesca Rossi}, \bibinfo{person}{Biplav Srivastava},
  \bibinfo{person}{Lior Horesh}, \bibinfo{person}{Francesco Fabiano}, {and}
  \bibinfo{person}{Andrea Loreggia}.} \bibinfo{year}{2023}\natexlab{}.
\newblock \showarticletitle{Understanding the Capabilities of Large Language
  Models for Automated Planning}.
\newblock \bibinfo{journal}{\emph{arXiv preprint arXiv:2305.16151}}
  (\bibinfo{year}{2023}).
\newblock


\bibitem[Pan et~al\mbox{.}(2023)]%
        {pan2023unifying}
\bibfield{author}{\bibinfo{person}{Shirui Pan}, \bibinfo{person}{Linhao Luo},
  \bibinfo{person}{Yufei Wang}, \bibinfo{person}{Chen Chen},
  \bibinfo{person}{Jiapu Wang}, {and} \bibinfo{person}{Xindong Wu}.}
  \bibinfo{year}{2023}\natexlab{}.
\newblock \bibinfo{title}{Unifying Large Language Models and Knowledge Graphs:
  A Roadmap}.
\newblock
\newblock
\showeprint[arxiv]{2306.08302}~[cs.CL]


\bibitem[Parisi et~al\mbox{.}(2022)]%
        {parisi2022talm}
\bibfield{author}{\bibinfo{person}{Aaron Parisi}, \bibinfo{person}{Yao Zhao},
  {and} \bibinfo{person}{Noah Fiedel}.} \bibinfo{year}{2022}\natexlab{}.
\newblock \showarticletitle{Talm: Tool augmented language models}.
\newblock \bibinfo{journal}{\emph{arXiv preprint arXiv:2205.12255}}
  (\bibinfo{year}{2022}).
\newblock


\bibitem[Parrish et~al\mbox{.}(2022)]%
        {parrish2022bbq}
\bibfield{author}{\bibinfo{person}{Alicia Parrish}, \bibinfo{person}{Angelica
  Chen}, \bibinfo{person}{Nikita Nangia}, \bibinfo{person}{Vishakh Padmakumar},
  \bibinfo{person}{Jason Phang}, \bibinfo{person}{Jana Thompson},
  \bibinfo{person}{Phu~Mon Htut}, {and} \bibinfo{person}{Samuel Bowman}.}
  \bibinfo{year}{2022}\natexlab{}.
\newblock \showarticletitle{BBQ: A hand-built bias benchmark for question
  answering}. In \bibinfo{booktitle}{\emph{Findings of the Association for
  Computational Linguistics: ACL 2022}}. \bibinfo{pages}{2086--2105}.
\newblock


\bibitem[Pe{\~n}a et~al\mbox{.}(2023)]%
        {pena2023leveraging}
\bibfield{author}{\bibinfo{person}{Alejandro Pe{\~n}a},
  \bibinfo{person}{Aythami Morales}, \bibinfo{person}{Julian Fierrez},
  \bibinfo{person}{Ignacio Serna}, \bibinfo{person}{Javier Ortega-Garcia},
  \bibinfo{person}{I{\~n}igo Puente}, \bibinfo{person}{Jorge Cordova}, {and}
  \bibinfo{person}{Gonzalo Cordova}.} \bibinfo{year}{2023}\natexlab{}.
\newblock \showarticletitle{Leveraging Large Language Models for Topic
  Classification in the Domain of Public Affairs}.
\newblock \bibinfo{journal}{\emph{arXiv preprint arXiv:2306.02864}}
  (\bibinfo{year}{2023}).
\newblock


\bibitem[Peng et~al\mbox{.}(1997)]%
        {peng1997validity}
\bibfield{author}{\bibinfo{person}{Kaiping Peng}, \bibinfo{person}{Richard~E
  Nisbett}, {and} \bibinfo{person}{Nancy~YC Wong}.}
  \bibinfo{year}{1997}\natexlab{}.
\newblock \showarticletitle{Validity problems comparing values across cultures
  and possible solutions.}
\newblock \bibinfo{journal}{\emph{Psychological methods}} \bibinfo{volume}{2},
  \bibinfo{number}{4} (\bibinfo{year}{1997}), \bibinfo{pages}{329}.
\newblock


\bibitem[Pezeshkpour(2023)]%
        {pezeshkpour2023measuring}
\bibfield{author}{\bibinfo{person}{Pouya Pezeshkpour}.}
  \bibinfo{year}{2023}\natexlab{}.
\newblock \showarticletitle{Measuring and Modifying Factual Knowledge in Large
  Language Models}.
\newblock \bibinfo{journal}{\emph{arXiv preprint arXiv:2306.06264}}
  (\bibinfo{year}{2023}).
\newblock


\bibitem[Phang et~al\mbox{.}(2021)]%
        {phang2021adversarially}
\bibfield{author}{\bibinfo{person}{Jason Phang}, \bibinfo{person}{Angelica
  Chen}, \bibinfo{person}{William Huang}, {and} \bibinfo{person}{Samuel~R
  Bowman}.} \bibinfo{year}{2021}\natexlab{}.
\newblock \showarticletitle{Adversarially constructed evaluation sets are more
  challenging, but may not be fair}.
\newblock \bibinfo{journal}{\emph{arXiv preprint arXiv:2111.08181}}
  (\bibinfo{year}{2021}).
\newblock


\bibitem[Pu and Demberg(2023)]%
        {pu2023chatgpt}
\bibfield{author}{\bibinfo{person}{Dongqi Pu} {and} \bibinfo{person}{Vera
  Demberg}.} \bibinfo{year}{2023}\natexlab{}.
\newblock \bibinfo{title}{ChatGPT vs Human-authored Text: Insights into
  Controllable Text Summarization and Sentence Style Transfer}.
\newblock
\newblock
\showeprint[arxiv]{2306.07799}~[cs.CL]


\bibitem[Qin et~al\mbox{.}(2023c)]%
        {qin2023chatgpt}
\bibfield{author}{\bibinfo{person}{Chengwei Qin}, \bibinfo{person}{Aston
  Zhang}, \bibinfo{person}{Zhuosheng Zhang}, \bibinfo{person}{Jiaao Chen},
  \bibinfo{person}{Michihiro Yasunaga}, {and} \bibinfo{person}{Diyi Yang}.}
  \bibinfo{year}{2023}\natexlab{c}.
\newblock \showarticletitle{Is ChatGPT a general-purpose natural language
  processing task solver?}
\newblock \bibinfo{journal}{\emph{arXiv preprint arXiv:2302.06476}}
  (\bibinfo{year}{2023}).
\newblock


\bibitem[Qin et~al\mbox{.}(2023a)]%
        {qin2023tool}
\bibfield{author}{\bibinfo{person}{Yujia Qin}, \bibinfo{person}{Shengding Hu},
  \bibinfo{person}{Yankai Lin}, \bibinfo{person}{Weize Chen},
  \bibinfo{person}{Ning Ding}, \bibinfo{person}{Ganqu Cui},
  \bibinfo{person}{Zheni Zeng}, \bibinfo{person}{Yufei Huang},
  \bibinfo{person}{Chaojun Xiao}, \bibinfo{person}{Chi Han},
  \bibinfo{person}{Yi~Ren Fung}, \bibinfo{person}{Yusheng Su},
  \bibinfo{person}{Huadong Wang}, \bibinfo{person}{Cheng Qian},
  \bibinfo{person}{Runchu Tian}, \bibinfo{person}{Kunlun Zhu},
  \bibinfo{person}{Shihao Liang}, \bibinfo{person}{Xingyu Shen},
  \bibinfo{person}{Bokai Xu}, \bibinfo{person}{Zhen Zhang},
  \bibinfo{person}{Yining Ye}, \bibinfo{person}{Bowen Li},
  \bibinfo{person}{Ziwei Tang}, \bibinfo{person}{Jing Yi},
  \bibinfo{person}{Yuzhang Zhu}, \bibinfo{person}{Zhenning Dai},
  \bibinfo{person}{Lan Yan}, \bibinfo{person}{Xin Cong}, \bibinfo{person}{Yaxi
  Lu}, \bibinfo{person}{Weilin Zhao}, \bibinfo{person}{Yuxiang Huang},
  \bibinfo{person}{Junxi Yan}, \bibinfo{person}{Xu Han}, \bibinfo{person}{Xian
  Sun}, \bibinfo{person}{Dahai Li}, \bibinfo{person}{Jason Phang},
  \bibinfo{person}{Cheng Yang}, \bibinfo{person}{Tongshuang Wu},
  \bibinfo{person}{Heng Ji}, \bibinfo{person}{Zhiyuan Liu}, {and}
  \bibinfo{person}{Maosong Sun}.} \bibinfo{year}{2023}\natexlab{a}.
\newblock \bibinfo{title}{Tool Learning with Foundation Models}.
\newblock
\newblock
\showeprint[arxiv]{2304.08354}~[cs.CL]


\bibitem[Qin et~al\mbox{.}(2023b)]%
        {qin2023toolllm}
\bibfield{author}{\bibinfo{person}{Yujia Qin}, \bibinfo{person}{Shihao Liang},
  \bibinfo{person}{Yining Ye}, \bibinfo{person}{Kunlun Zhu},
  \bibinfo{person}{Lan Yan}, \bibinfo{person}{Yaxi Lu}, \bibinfo{person}{Yankai
  Lin}, \bibinfo{person}{Xin Cong}, \bibinfo{person}{Xiangru Tang},
  \bibinfo{person}{Bill Qian}, \bibinfo{person}{Sihan Zhao},
  \bibinfo{person}{Runchu Tian}, \bibinfo{person}{Ruobing Xie},
  \bibinfo{person}{Jie Zhou}, \bibinfo{person}{Mark Gerstein},
  \bibinfo{person}{Dahai Li}, \bibinfo{person}{Zhiyuan Liu}, {and}
  \bibinfo{person}{Maosong Sun}.} \bibinfo{year}{2023}\natexlab{b}.
\newblock \bibinfo{title}{ToolLLM: Facilitating Large Language Models to Master
  16000+ Real-world APIs}.
\newblock
\newblock
\showeprint[arxiv]{2307.16789}~[cs.AI]


\bibitem[Radford et~al\mbox{.}(2018)]%
        {radford2018improving}
\bibfield{author}{\bibinfo{person}{Alec Radford}, \bibinfo{person}{Karthik
  Narasimhan}, \bibinfo{person}{Tim Salimans}, \bibinfo{person}{Ilya
  Sutskever}, {et~al\mbox{.}}} \bibinfo{year}{2018}\natexlab{}.
\newblock \showarticletitle{Improving language understanding by generative
  pre-training}.
\newblock  (\bibinfo{year}{2018}).
\newblock


\bibitem[Rawte et~al\mbox{.}(2023)]%
        {rawte2023survey}
\bibfield{author}{\bibinfo{person}{Vipula Rawte}, \bibinfo{person}{Amit Sheth},
  {and} \bibinfo{person}{Amitava Das}.} \bibinfo{year}{2023}\natexlab{}.
\newblock \showarticletitle{A Survey of Hallucination in Large Foundation
  Models}.
\newblock \bibinfo{journal}{\emph{arXiv preprint arXiv:2309.05922}}
  (\bibinfo{year}{2023}).
\newblock


\bibitem[Ribeiro and Lundberg(2022)]%
        {ribeiro2022adaptive}
\bibfield{author}{\bibinfo{person}{Marco~Tulio Ribeiro} {and}
  \bibinfo{person}{Scott Lundberg}.} \bibinfo{year}{2022}\natexlab{}.
\newblock \showarticletitle{Adaptive testing and debugging of nlp models}. In
  \bibinfo{booktitle}{\emph{Proceedings of the 60th Annual Meeting of the
  Association for Computational Linguistics (Volume 1: Long Papers)}}.
  \bibinfo{pages}{3253--3267}.
\newblock


\bibitem[Ribeiro et~al\mbox{.}(2020)]%
        {ribeiro2020beyond}
\bibfield{author}{\bibinfo{person}{Marco~Tulio Ribeiro},
  \bibinfo{person}{Tongshuang Wu}, \bibinfo{person}{Carlos Guestrin}, {and}
  \bibinfo{person}{Sameer Singh}.} \bibinfo{year}{2020}\natexlab{}.
\newblock \showarticletitle{Beyond accuracy: Behavioral testing of NLP models
  with CheckList}.
\newblock \bibinfo{journal}{\emph{arXiv preprint arXiv:2005.04118}}
  (\bibinfo{year}{2020}).
\newblock


\bibitem[Riccardi and Desai(2023)]%
        {riccardi2023two}
\bibfield{author}{\bibinfo{person}{Nicholas Riccardi} {and}
  \bibinfo{person}{Rutvik~H Desai}.} \bibinfo{year}{2023}\natexlab{}.
\newblock \showarticletitle{The Two Word Test: A Semantic Benchmark for Large
  Language Models}.
\newblock \bibinfo{journal}{\emph{arXiv preprint arXiv:2306.04610}}
  (\bibinfo{year}{2023}).
\newblock


\bibitem[Rutinowski et~al\mbox{.}(2023)]%
        {rutinowski2023self}
\bibfield{author}{\bibinfo{person}{J{\'e}r{\^o}me Rutinowski},
  \bibinfo{person}{Sven Franke}, \bibinfo{person}{Jan Endendyk},
  \bibinfo{person}{Ina Dormuth}, {and} \bibinfo{person}{Markus Pauly}.}
  \bibinfo{year}{2023}\natexlab{}.
\newblock \showarticletitle{The Self-Perception and Political Biases of
  ChatGPT}.
\newblock \bibinfo{journal}{\emph{arXiv preprint arXiv:2304.07333}}
  (\bibinfo{year}{2023}).
\newblock


\bibitem[Safdari et~al\mbox{.}(2023)]%
        {safdari2023personality}
\bibfield{author}{\bibinfo{person}{Mustafa Safdari}, \bibinfo{person}{Greg
  Serapio-Garc{\'\i}a}, \bibinfo{person}{Cl{\'e}ment Crepy},
  \bibinfo{person}{Stephen Fitz}, \bibinfo{person}{Peter Romero},
  \bibinfo{person}{Luning Sun}, \bibinfo{person}{Marwa Abdulhai},
  \bibinfo{person}{Aleksandra Faust}, {and} \bibinfo{person}{Maja
  Matari{\'c}}.} \bibinfo{year}{2023}\natexlab{}.
\newblock \showarticletitle{Personality Traits in Large Language Models}.
\newblock \bibinfo{journal}{\emph{arXiv preprint arXiv:2307.00184}}
  (\bibinfo{year}{2023}).
\newblock


\bibitem[Samaan et~al\mbox{.}(2023)]%
        {samaan2023assessing}
\bibfield{author}{\bibinfo{person}{Jamil~S Samaan}, \bibinfo{person}{Yee~Hui
  Yeo}, \bibinfo{person}{Nithya Rajeev}, \bibinfo{person}{Lauren Hawley},
  \bibinfo{person}{Stuart Abel}, \bibinfo{person}{Wee~Han Ng},
  \bibinfo{person}{Nitin Srinivasan}, \bibinfo{person}{Justin Park},
  \bibinfo{person}{Miguel Burch}, \bibinfo{person}{Rabindra Watson},
  {et~al\mbox{.}}} \bibinfo{year}{2023}\natexlab{}.
\newblock \showarticletitle{Assessing the accuracy of responses by the language
  model ChatGPT to questions regarding bariatric surgery}.
\newblock \bibinfo{journal}{\emph{Obesity Surgery}} (\bibinfo{year}{2023}),
  \bibinfo{pages}{1--7}.
\newblock


\bibitem[Saparov et~al\mbox{.}(2023)]%
        {saparov2023testing}
\bibfield{author}{\bibinfo{person}{Abulhair Saparov},
  \bibinfo{person}{Richard~Yuanzhe Pang}, \bibinfo{person}{Vishakh Padmakumar},
  \bibinfo{person}{Nitish Joshi}, \bibinfo{person}{Seyed~Mehran Kazemi},
  \bibinfo{person}{Najoung Kim}, {and} \bibinfo{person}{He He}.}
  \bibinfo{year}{2023}\natexlab{}.
\newblock \showarticletitle{Testing the General Deductive Reasoning Capacity of
  Large Language Models Using OOD Examples}.
\newblock \bibinfo{journal}{\emph{arXiv preprint arXiv:2305.15269}}
  (\bibinfo{year}{2023}).
\newblock


\bibitem[Sawada et~al\mbox{.}(2023)]%
        {sawada2023arb}
\bibfield{author}{\bibinfo{person}{Tomohiro Sawada}, \bibinfo{person}{Daniel
  Paleka}, \bibinfo{person}{Alexander Havrilla}, \bibinfo{person}{Pranav
  Tadepalli}, \bibinfo{person}{Paula Vidas}, \bibinfo{person}{Alexander
  Kranias}, \bibinfo{person}{John~J. Nay}, \bibinfo{person}{Kshitij Gupta},
  {and} \bibinfo{person}{Aran Komatsuzaki}.} \bibinfo{year}{2023}\natexlab{}.
\newblock \bibinfo{title}{ARB: Advanced Reasoning Benchmark for Large Language
  Models}.
\newblock
\newblock
\showeprint[arxiv]{2307.13692}~[cs.CL]


\bibitem[Schick et~al\mbox{.}(2023)]%
        {schick2023toolformer}
\bibfield{author}{\bibinfo{person}{Timo Schick}, \bibinfo{person}{Jane
  Dwivedi-Yu}, \bibinfo{person}{Roberto Dess{\`\i}}, \bibinfo{person}{Roberta
  Raileanu}, \bibinfo{person}{Maria Lomeli}, \bibinfo{person}{Luke
  Zettlemoyer}, \bibinfo{person}{Nicola Cancedda}, {and}
  \bibinfo{person}{Thomas Scialom}.} \bibinfo{year}{2023}\natexlab{}.
\newblock \showarticletitle{Toolformer: Language models can teach themselves to
  use tools}.
\newblock \bibinfo{journal}{\emph{arXiv preprint arXiv:2302.04761}}
  (\bibinfo{year}{2023}).
\newblock


\bibitem[Sharma et~al\mbox{.}(2023)]%
        {sharma2023performance}
\bibfield{author}{\bibinfo{person}{Prabin Sharma}, \bibinfo{person}{Kisan
  Thapa}, \bibinfo{person}{Prastab Dhakal}, \bibinfo{person}{Mala~Deep
  Upadhaya}, \bibinfo{person}{Santosh Adhikari}, {and}
  \bibinfo{person}{Salik~Ram Khanal}.} \bibinfo{year}{2023}\natexlab{}.
\newblock \showarticletitle{Performance of ChatGPT on USMLE: Unlocking the
  Potential of Large Language Models for AI-Assisted Medical Education}.
\newblock \bibinfo{journal}{\emph{arXiv preprint arXiv:2307.00112}}
  (\bibinfo{year}{2023}).
\newblock


\bibitem[Shen et~al\mbox{.}(2023)]%
        {shen2023hugginggpt}
\bibfield{author}{\bibinfo{person}{Yongliang Shen}, \bibinfo{person}{Kaitao
  Song}, \bibinfo{person}{Xu Tan}, \bibinfo{person}{Dongsheng Li},
  \bibinfo{person}{Weiming Lu}, {and} \bibinfo{person}{Yueting Zhuang}.}
  \bibinfo{year}{2023}\natexlab{}.
\newblock \showarticletitle{Hugginggpt: Solving ai tasks with chatgpt and its
  friends in huggingface}.
\newblock \bibinfo{journal}{\emph{arXiv preprint arXiv:2303.17580}}
  (\bibinfo{year}{2023}).
\newblock


\bibitem[Sheng et~al\mbox{.}(2021)]%
        {sheng2021societal}
\bibfield{author}{\bibinfo{person}{Emily Sheng}, \bibinfo{person}{Kai-Wei
  Chang}, \bibinfo{person}{Prem Natarajan}, {and} \bibinfo{person}{Nanyun
  Peng}.} \bibinfo{year}{2021}\natexlab{}.
\newblock \showarticletitle{Societal Biases in Language Generation: Progress
  and Challenges}. In \bibinfo{booktitle}{\emph{Proceedings of the 59th Annual
  Meeting of the Association for Computational Linguistics and the 11th
  International Joint Conference on Natural Language Processing (Volume 1: Long
  Papers)}}. \bibinfo{pages}{4275--4293}.
\newblock


\bibitem[Simmons(2022)]%
        {simmons2022moral}
\bibfield{author}{\bibinfo{person}{Gabriel Simmons}.}
  \bibinfo{year}{2022}\natexlab{}.
\newblock \showarticletitle{Moral mimicry: Large language models produce moral
  rationalizations tailored to political identity}.
\newblock \bibinfo{journal}{\emph{arXiv preprint arXiv:2209.12106}}
  (\bibinfo{year}{2022}).
\newblock


\bibitem[Singhal et~al\mbox{.}(2022)]%
        {singhal2022large}
\bibfield{author}{\bibinfo{person}{Karan Singhal}, \bibinfo{person}{Shekoofeh
  Azizi}, \bibinfo{person}{Tao Tu}, \bibinfo{person}{S~Sara Mahdavi},
  \bibinfo{person}{Jason Wei}, \bibinfo{person}{Hyung~Won Chung},
  \bibinfo{person}{Nathan Scales}, \bibinfo{person}{Ajay Tanwani},
  \bibinfo{person}{Heather Cole-Lewis}, \bibinfo{person}{Stephen Pfohl},
  {et~al\mbox{.}}} \bibinfo{year}{2022}\natexlab{}.
\newblock \showarticletitle{Large Language Models Encode Clinical Knowledge}.
\newblock \bibinfo{journal}{\emph{arXiv preprint arXiv:2212.13138}}
  (\bibinfo{year}{2022}).
\newblock


\bibitem[Singhal et~al\mbox{.}(2023)]%
        {singhal2023large}
\bibfield{author}{\bibinfo{person}{Karan Singhal}, \bibinfo{person}{Shekoofeh
  Azizi}, \bibinfo{person}{Tao Tu}, \bibinfo{person}{S~Sara Mahdavi},
  \bibinfo{person}{Jason Wei}, \bibinfo{person}{Hyung~Won Chung},
  \bibinfo{person}{Nathan Scales}, \bibinfo{person}{Ajay Tanwani},
  \bibinfo{person}{Heather Cole-Lewis}, \bibinfo{person}{Stephen Pfohl},
  {et~al\mbox{.}}} \bibinfo{year}{2023}\natexlab{}.
\newblock \showarticletitle{Large language models encode clinical knowledge}.
\newblock \bibinfo{journal}{\emph{Nature}} \bibinfo{volume}{620},
  \bibinfo{number}{7972} (\bibinfo{year}{2023}), \bibinfo{pages}{172--180}.
\newblock


\bibitem[Smith et~al\mbox{.}(2022)]%
        {smith2022using}
\bibfield{author}{\bibinfo{person}{Shaden Smith}, \bibinfo{person}{Mostofa
  Patwary}, \bibinfo{person}{Brandon Norick}, \bibinfo{person}{Patrick
  LeGresley}, \bibinfo{person}{Samyam Rajbhandari}, \bibinfo{person}{Jared
  Casper}, \bibinfo{person}{Zhun Liu}, \bibinfo{person}{Shrimai Prabhumoye},
  \bibinfo{person}{George Zerveas}, \bibinfo{person}{Vijay Korthikanti},
  {et~al\mbox{.}}} \bibinfo{year}{2022}\natexlab{}.
\newblock \showarticletitle{Using deepspeed and megatron to train
  megatron-turing nlg 530b, a large-scale generative language model}.
\newblock \bibinfo{journal}{\emph{arXiv preprint arXiv:2201.11990}}
  (\bibinfo{year}{2022}).
\newblock


\bibitem[Song et~al\mbox{.}(2023)]%
        {song2023have}
\bibfield{author}{\bibinfo{person}{Xiaoyang Song}, \bibinfo{person}{Akshat
  Gupta}, \bibinfo{person}{Kiyan Mohebbizadeh}, \bibinfo{person}{Shujie Hu},
  {and} \bibinfo{person}{Anant Singh}.} \bibinfo{year}{2023}\natexlab{}.
\newblock \showarticletitle{Have Large Language Models Developed a
  Personality?: Applicability of Self-Assessment Tests in Measuring Personality
  in LLMs}.
\newblock \bibinfo{journal}{\emph{arXiv preprint arXiv:2305.14693}}
  (\bibinfo{year}{2023}).
\newblock


\bibitem[Sridhara et~al\mbox{.}(2023)]%
        {sridhara2023chatgpt}
\bibfield{author}{\bibinfo{person}{Giriprasad Sridhara},
  \bibinfo{person}{Sourav Mazumdar}, {et~al\mbox{.}}}
  \bibinfo{year}{2023}\natexlab{}.
\newblock \showarticletitle{ChatGPT: A Study on its Utility for Ubiquitous
  Software Engineering Tasks}.
\newblock \bibinfo{journal}{\emph{arXiv preprint arXiv:2305.16837}}
  (\bibinfo{year}{2023}).
\newblock


\bibitem[Srivastava et~al\mbox{.}(2022)]%
        {srivastava2022beyond}
\bibfield{author}{\bibinfo{person}{Aarohi Srivastava}, \bibinfo{person}{Abhinav
  Rastogi}, \bibinfo{person}{Abhishek Rao}, \bibinfo{person}{Abu Awal~Md
  Shoeb}, \bibinfo{person}{Abubakar Abid}, \bibinfo{person}{Adam Fisch},
  \bibinfo{person}{Adam~R Brown}, \bibinfo{person}{Adam Santoro},
  \bibinfo{person}{Aditya Gupta}, \bibinfo{person}{Adri{\`a} Garriga-Alonso},
  {et~al\mbox{.}}} \bibinfo{year}{2022}\natexlab{}.
\newblock \showarticletitle{Beyond the imitation game: Quantifying and
  extrapolating the capabilities of language models}.
\newblock \bibinfo{journal}{\emph{arXiv preprint arXiv:2206.04615}}
  (\bibinfo{year}{2022}).
\newblock


\bibitem[Sun et~al\mbox{.}(2023)]%
        {sun2023chatgpt}
\bibfield{author}{\bibinfo{person}{Weiwei Sun}, \bibinfo{person}{Lingyong Yan},
  \bibinfo{person}{Xinyu Ma}, \bibinfo{person}{Pengjie Ren},
  \bibinfo{person}{Dawei Yin}, {and} \bibinfo{person}{Zhaochun Ren}.}
  \bibinfo{year}{2023}\natexlab{}.
\newblock \showarticletitle{Is ChatGPT Good at Search? Investigating Large
  Language Models as Re-Ranking Agent}.
\newblock \bibinfo{journal}{\emph{arXiv preprint arXiv:2304.09542}}
  (\bibinfo{year}{2023}).
\newblock


\bibitem[Tao et~al\mbox{.}(2023)]%
        {tao2023eveval}
\bibfield{author}{\bibinfo{person}{Zhengwei Tao}, \bibinfo{person}{Zhi Jin},
  \bibinfo{person}{Xiaoying Bai}, \bibinfo{person}{Haiyan Zhao},
  \bibinfo{person}{Yanlin Feng}, \bibinfo{person}{Jia Li}, {and}
  \bibinfo{person}{Wenpeng Hu}.} \bibinfo{year}{2023}\natexlab{}.
\newblock \showarticletitle{EvEval: A Comprehensive Evaluation of Event
  Semantics for Large Language Models}.
\newblock \bibinfo{journal}{\emph{arXiv preprint arXiv:2305.15268}}
  (\bibinfo{year}{2023}).
\newblock


\bibitem[Thakur et~al\mbox{.}(2021)]%
        {thakur2021beir}
\bibfield{author}{\bibinfo{person}{Nandan Thakur}, \bibinfo{person}{Nils
  Reimers}, \bibinfo{person}{Andreas R{\"u}ckl{\'e}}, \bibinfo{person}{Abhishek
  Srivastava}, {and} \bibinfo{person}{Iryna Gurevych}.}
  \bibinfo{year}{2021}\natexlab{}.
\newblock \showarticletitle{Beir: A heterogenous benchmark for zero-shot
  evaluation of information retrieval models}.
\newblock \bibinfo{journal}{\emph{arXiv preprint arXiv:2104.08663}}
  (\bibinfo{year}{2021}).
\newblock


\bibitem[Thirunavukarasu et~al\mbox{.}(2023)]%
        {thirunavukarasu2023trialling}
\bibfield{author}{\bibinfo{person}{Arun~James Thirunavukarasu},
  \bibinfo{person}{Refaat Hassan}, \bibinfo{person}{Shathar Mahmood},
  \bibinfo{person}{Rohan Sanghera}, \bibinfo{person}{Kara Barzangi},
  \bibinfo{person}{Mohanned El~Mukashfi}, {and} \bibinfo{person}{Sachin Shah}.}
  \bibinfo{year}{2023}\natexlab{}.
\newblock \showarticletitle{Trialling a large language model (ChatGPT) in
  general practice with the Applied Knowledge Test: observational study
  demonstrating opportunities and limitations in primary care}.
\newblock \bibinfo{journal}{\emph{JMIR Medical Education}} \bibinfo{volume}{9},
  \bibinfo{number}{1} (\bibinfo{year}{2023}), \bibinfo{pages}{e46599}.
\newblock


\bibitem[Thoppilan et~al\mbox{.}(2022)]%
        {thoppilan2022lamda}
\bibfield{author}{\bibinfo{person}{Romal Thoppilan}, \bibinfo{person}{Daniel
  De~Freitas}, \bibinfo{person}{Jamie Hall}, \bibinfo{person}{Noam Shazeer},
  \bibinfo{person}{Apoorv Kulshreshtha}, \bibinfo{person}{Heng-Tze Cheng},
  \bibinfo{person}{Alicia Jin}, \bibinfo{person}{Taylor Bos},
  \bibinfo{person}{Leslie Baker}, \bibinfo{person}{Yu Du}, {et~al\mbox{.}}}
  \bibinfo{year}{2022}\natexlab{}.
\newblock \showarticletitle{Lamda: Language models for dialog applications}.
\newblock \bibinfo{journal}{\emph{arXiv preprint arXiv:2201.08239}}
  (\bibinfo{year}{2022}).
\newblock


\bibitem[Thrush et~al\mbox{.}(2022)]%
        {thrush2022dynatask}
\bibfield{author}{\bibinfo{person}{Tristan Thrush}, \bibinfo{person}{Kushal
  Tirumala}, \bibinfo{person}{Anmol Gupta}, \bibinfo{person}{Max Bartolo},
  \bibinfo{person}{Pedro Rodriguez}, \bibinfo{person}{Tariq Kane},
  \bibinfo{person}{William~Gaviria Rojas}, \bibinfo{person}{Peter Mattson},
  \bibinfo{person}{Adina Williams}, {and} \bibinfo{person}{Douwe Kiela}.}
  \bibinfo{year}{2022}\natexlab{}.
\newblock \showarticletitle{Dynatask: A framework for creating dynamic AI
  benchmark tasks}.
\newblock \bibinfo{journal}{\emph{arXiv preprint arXiv:2204.01906}}
  (\bibinfo{year}{2022}).
\newblock


\bibitem[Tian et~al\mbox{.}(2023)]%
        {tian2023just}
\bibfield{author}{\bibinfo{person}{Katherine Tian}, \bibinfo{person}{Eric
  Mitchell}, \bibinfo{person}{Allan Zhou}, \bibinfo{person}{Archit Sharma},
  \bibinfo{person}{Rafael Rafailov}, \bibinfo{person}{Huaxiu Yao},
  \bibinfo{person}{Chelsea Finn}, {and} \bibinfo{person}{Christopher~D
  Manning}.} \bibinfo{year}{2023}\natexlab{}.
\newblock \showarticletitle{Just ask for calibration: Strategies for eliciting
  calibrated confidence scores from language models fine-tuned with human
  feedback}.
\newblock \bibinfo{journal}{\emph{arXiv preprint arXiv:2305.14975}}
  (\bibinfo{year}{2023}).
\newblock


\bibitem[Tian et~al\mbox{.}(2018)]%
        {tian2018deeptest}
\bibfield{author}{\bibinfo{person}{Yuchi Tian}, \bibinfo{person}{Kexin Pei},
  \bibinfo{person}{Suman Jana}, {and} \bibinfo{person}{Baishakhi Ray}.}
  \bibinfo{year}{2018}\natexlab{}.
\newblock \showarticletitle{Deeptest: Automated testing of
  deep-neural-network-driven autonomous cars}. In
  \bibinfo{booktitle}{\emph{Proceedings of the 40th international conference on
  software engineering}}. \bibinfo{pages}{303--314}.
\newblock


\bibitem[ToolBench(2023)]%
        {toolbench}
\bibfield{author}{\bibinfo{person}{ToolBench}.}
  \bibinfo{year}{2023}\natexlab{}.
\newblock \bibinfo{title}{Open-source tools learning benchmarks}.
\newblock \bibinfo{howpublished}{\url{https://github.com/sambanova/toolbench}}.
\newblock


\bibitem[Touvron et~al\mbox{.}(2023)]%
        {touvron2023llama}
\bibfield{author}{\bibinfo{person}{Hugo Touvron}, \bibinfo{person}{Thibaut
  Lavril}, \bibinfo{person}{Gautier Izacard}, \bibinfo{person}{Xavier
  Martinet}, \bibinfo{person}{Marie-Anne Lachaux},
  \bibinfo{person}{Timoth{\'e}e Lacroix}, \bibinfo{person}{Baptiste
  Rozi{\`e}re}, \bibinfo{person}{Naman Goyal}, \bibinfo{person}{Eric Hambro},
  \bibinfo{person}{Faisal Azhar}, {et~al\mbox{.}}}
  \bibinfo{year}{2023}\natexlab{}.
\newblock \showarticletitle{Llama: Open and efficient foundation language
  models}.
\newblock \bibinfo{journal}{\emph{arXiv preprint arXiv:2302.13971}}
  (\bibinfo{year}{2023}).
\newblock


\bibitem[Turing(2009)]%
        {turing2009computing}
\bibfield{author}{\bibinfo{person}{Alan~M Turing}.}
  \bibinfo{year}{2009}\natexlab{}.
\newblock \bibinfo{booktitle}{\emph{Computing machinery and intelligence}}.
\newblock \bibinfo{publisher}{Springer}.
\newblock


\bibitem[Valmeekam et~al\mbox{.}(2023)]%
        {valmeekam2023planning}
\bibfield{author}{\bibinfo{person}{Karthik Valmeekam}, \bibinfo{person}{Matthew
  Marquez}, \bibinfo{person}{Sarath Sreedharan}, {and}
  \bibinfo{person}{Subbarao Kambhampati}.} \bibinfo{year}{2023}\natexlab{}.
\newblock \showarticletitle{On the Planning Abilities of Large Language
  Models--A Critical Investigation}.
\newblock \bibinfo{journal}{\emph{arXiv preprint arXiv:2305.15771}}
  (\bibinfo{year}{2023}).
\newblock


\bibitem[Valmeekam et~al\mbox{.}(2022)]%
        {valmeekam2022large}
\bibfield{author}{\bibinfo{person}{Karthik Valmeekam}, \bibinfo{person}{Alberto
  Olmo}, \bibinfo{person}{Sarath Sreedharan}, {and} \bibinfo{person}{Subbarao
  Kambhampati}.} \bibinfo{year}{2022}\natexlab{}.
\newblock \showarticletitle{Large Language Models Still Can't Plan (A Benchmark
  for LLMs on Planning and Reasoning about Change)}.
\newblock \bibinfo{journal}{\emph{arXiv preprint arXiv:2206.10498}}
  (\bibinfo{year}{2022}).
\newblock


\bibitem[Van Der~Lee et~al\mbox{.}(2019)]%
        {van2019best}
\bibfield{author}{\bibinfo{person}{Chris Van Der~Lee}, \bibinfo{person}{Albert
  Gatt}, \bibinfo{person}{Emiel Van~Miltenburg}, \bibinfo{person}{Sander
  Wubben}, {and} \bibinfo{person}{Emiel Krahmer}.}
  \bibinfo{year}{2019}\natexlab{}.
\newblock \showarticletitle{Best practices for the human evaluation of
  automatically generated text}. In \bibinfo{booktitle}{\emph{Proceedings of
  the 12th International Conference on Natural Language Generation}}.
  \bibinfo{pages}{355--368}.
\newblock


\bibitem[Vaswani et~al\mbox{.}(2017)]%
        {vaswani2017attention}
\bibfield{author}{\bibinfo{person}{Ashish Vaswani}, \bibinfo{person}{Noam
  Shazeer}, \bibinfo{person}{Niki Parmar}, \bibinfo{person}{Jakob Uszkoreit},
  \bibinfo{person}{Llion Jones}, \bibinfo{person}{Aidan~N Gomez},
  \bibinfo{person}{{\L}ukasz Kaiser}, {and} \bibinfo{person}{Illia
  Polosukhin}.} \bibinfo{year}{2017}\natexlab{}.
\newblock \showarticletitle{Attention is all you need}.
\newblock \bibinfo{journal}{\emph{Advances in neural information processing
  systems}}  \bibinfo{volume}{30} (\bibinfo{year}{2017}).
\newblock


\bibitem[Vu et~al\mbox{.}(2023)]%
        {vu2023freshllms}
\bibfield{author}{\bibinfo{person}{Tu Vu}, \bibinfo{person}{Mohit Iyyer},
  \bibinfo{person}{Xuezhi Wang}, \bibinfo{person}{Noah Constant},
  \bibinfo{person}{Jerry Wei}, \bibinfo{person}{Jason Wei},
  \bibinfo{person}{Chris Tar}, \bibinfo{person}{Yun-Hsuan Sung},
  \bibinfo{person}{Denny Zhou}, \bibinfo{person}{Quoc Le}, {and}
  \bibinfo{person}{Thang Luong}.} \bibinfo{year}{2023}\natexlab{}.
\newblock \bibinfo{title}{FreshLLMs: Refreshing Large Language Models with
  Search Engine Augmentation}.
\newblock
\newblock
\showeprint[arxiv]{2310.03214}~[cs.CL]


\bibitem[Wang et~al\mbox{.}(2019)]%
        {wang2019superglue}
\bibfield{author}{\bibinfo{person}{Alex Wang}, \bibinfo{person}{Yada
  Pruksachatkun}, \bibinfo{person}{Nikita Nangia}, \bibinfo{person}{Amanpreet
  Singh}, \bibinfo{person}{Julian Michael}, \bibinfo{person}{Felix Hill},
  \bibinfo{person}{Omer Levy}, {and} \bibinfo{person}{Samuel Bowman}.}
  \bibinfo{year}{2019}\natexlab{}.
\newblock \showarticletitle{Superglue: A stickier benchmark for general-purpose
  language understanding systems}.
\newblock \bibinfo{journal}{\emph{Advances in neural information processing
  systems}}  \bibinfo{volume}{32} (\bibinfo{year}{2019}).
\newblock


\bibitem[Wang et~al\mbox{.}(2018)]%
        {wang2018glue}
\bibfield{author}{\bibinfo{person}{Alex Wang}, \bibinfo{person}{Amanpreet
  Singh}, \bibinfo{person}{Julian Michael}, \bibinfo{person}{Felix Hill},
  \bibinfo{person}{Omer Levy}, {and} \bibinfo{person}{Samuel~R Bowman}.}
  \bibinfo{year}{2018}\natexlab{}.
\newblock \showarticletitle{GLUE: A multi-task benchmark and analysis platform
  for natural language understanding}.
\newblock \bibinfo{journal}{\emph{arXiv preprint arXiv:1804.07461}}
  (\bibinfo{year}{2018}).
\newblock


\bibitem[Wang et~al\mbox{.}(2023a)]%
        {wang2023decodingtrust}
\bibfield{author}{\bibinfo{person}{Boxin Wang}, \bibinfo{person}{Weixin Chen},
  \bibinfo{person}{Hengzhi Pei}, \bibinfo{person}{Chulin Xie},
  \bibinfo{person}{Mintong Kang}, \bibinfo{person}{Chenhui Zhang},
  \bibinfo{person}{Chejian Xu}, \bibinfo{person}{Zidi Xiong},
  \bibinfo{person}{Ritik Dutta}, \bibinfo{person}{Rylan Schaeffer},
  \bibinfo{person}{Sang~T. Truong}, \bibinfo{person}{Simran Arora},
  \bibinfo{person}{Mantas Mazeika}, \bibinfo{person}{Dan Hendrycks},
  \bibinfo{person}{Zinan Lin}, \bibinfo{person}{Yu Cheng},
  \bibinfo{person}{Sanmi Koyejo}, \bibinfo{person}{Dawn Song}, {and}
  \bibinfo{person}{Bo Li}.} \bibinfo{year}{2023}\natexlab{a}.
\newblock \bibinfo{title}{DecodingTrust: A Comprehensive Assessment of
  Trustworthiness in GPT Models}.
\newblock
\newblock
\showeprint[arxiv]{2306.11698}~[cs.CL]


\bibitem[Wang and Komatsuzaki(2021)]%
        {wang2021gpt}
\bibfield{author}{\bibinfo{person}{Ben Wang} {and} \bibinfo{person}{Aran
  Komatsuzaki}.} \bibinfo{year}{2021}\natexlab{}.
\newblock \bibinfo{title}{GPT-J-6B: A 6 billion parameter autoregressive
  language model}.
\newblock
\newblock


\bibitem[Wang et~al\mbox{.}(2021b)]%
        {wang2021adversarial}
\bibfield{author}{\bibinfo{person}{Boxin Wang}, \bibinfo{person}{Chejian Xu},
  \bibinfo{person}{Shuohang Wang}, \bibinfo{person}{Zhe Gan},
  \bibinfo{person}{Yu Cheng}, \bibinfo{person}{Jianfeng Gao},
  \bibinfo{person}{Ahmed~Hassan Awadallah}, {and} \bibinfo{person}{Bo Li}.}
  \bibinfo{year}{2021}\natexlab{b}.
\newblock \showarticletitle{Adversarial glue: A multi-task benchmark for
  robustness evaluation of language models}.
\newblock \bibinfo{journal}{\emph{arXiv preprint arXiv:2111.02840}}
  (\bibinfo{year}{2021}).
\newblock


\bibitem[Wang et~al\mbox{.}(2023c)]%
        {wang2023evaluating}
\bibfield{author}{\bibinfo{person}{Cunxiang Wang}, \bibinfo{person}{Sirui
  Cheng}, \bibinfo{person}{Zhikun Xu}, \bibinfo{person}{Bowen Ding},
  \bibinfo{person}{Yidong Wang}, {and} \bibinfo{person}{Yue Zhang}.}
  \bibinfo{year}{2023}\natexlab{c}.
\newblock \showarticletitle{Evaluating open question answering evaluation}.
\newblock \bibinfo{journal}{\emph{arXiv preprint arXiv:2305.12421}}
  (\bibinfo{year}{2023}).
\newblock


\bibitem[Wang et~al\mbox{.}(2023j)]%
        {wang2023chainofthought}
\bibfield{author}{\bibinfo{person}{Hongru Wang}, \bibinfo{person}{Rui Wang},
  \bibinfo{person}{Fei Mi}, \bibinfo{person}{Zezhong Wang},
  \bibinfo{person}{Ruifeng Xu}, {and} \bibinfo{person}{Kam-Fai Wong}.}
  \bibinfo{year}{2023}\natexlab{j}.
\newblock \bibinfo{title}{Chain-of-thought prompting for responding to in-depth
  dialogue questions with LLM}.
\newblock
\newblock
\showeprint[arxiv]{2305.11792}~[cs.CL]


\bibitem[Wang et~al\mbox{.}(2023d)]%
        {wang2023robustness}
\bibfield{author}{\bibinfo{person}{Jindong Wang}, \bibinfo{person}{Xixu Hu},
  \bibinfo{person}{Wenxin Hou}, \bibinfo{person}{Hao Chen},
  \bibinfo{person}{Runkai Zheng}, \bibinfo{person}{Yidong Wang},
  \bibinfo{person}{Linyi Yang}, \bibinfo{person}{Haojun Huang},
  \bibinfo{person}{Wei Ye}, \bibinfo{person}{Xiubo Geng}, {et~al\mbox{.}}}
  \bibinfo{year}{2023}\natexlab{d}.
\newblock \showarticletitle{On the robustness of chatgpt: An adversarial and
  out-of-distribution perspective}. In \bibinfo{booktitle}{\emph{ICLR workshop
  on Trustworthy and Reliable Large-Scale Machine Learning Models}}.
\newblock


\bibitem[Wang et~al\mbox{.}(2022)]%
        {wang2022generalizing}
\bibfield{author}{\bibinfo{person}{Jindong Wang}, \bibinfo{person}{Cuiling
  Lan}, \bibinfo{person}{Chang Liu}, \bibinfo{person}{Yidong Ouyang},
  \bibinfo{person}{Tao Qin}, \bibinfo{person}{Wang Lu},
  \bibinfo{person}{Yiqiang Chen}, \bibinfo{person}{Wenjun Zeng}, {and}
  \bibinfo{person}{Philip Yu}.} \bibinfo{year}{2022}\natexlab{}.
\newblock \showarticletitle{Generalizing to unseen domains: A survey on domain
  generalization}.
\newblock \bibinfo{journal}{\emph{IEEE Transactions on Knowledge and Data
  Engineering}} (\bibinfo{year}{2022}).
\newblock


\bibitem[Wang et~al\mbox{.}(2023h)]%
        {wang2023document}
\bibfield{author}{\bibinfo{person}{Longyue Wang}, \bibinfo{person}{Chenyang
  Lyu}, \bibinfo{person}{Tianbo Ji}, \bibinfo{person}{Zhirui Zhang},
  \bibinfo{person}{Dian Yu}, \bibinfo{person}{Shuming Shi}, {and}
  \bibinfo{person}{Zhaopeng Tu}.} \bibinfo{year}{2023}\natexlab{h}.
\newblock \showarticletitle{Document-level machine translation with large
  language models}.
\newblock \bibinfo{journal}{\emph{arXiv preprint arXiv:2304.02210}}
  (\bibinfo{year}{2023}).
\newblock


\bibitem[Wang et~al\mbox{.}(2023e)]%
        {wang2023large}
\bibfield{author}{\bibinfo{person}{Peiyi Wang}, \bibinfo{person}{Lei Li},
  \bibinfo{person}{Liang Chen}, \bibinfo{person}{Dawei Zhu},
  \bibinfo{person}{Binghuai Lin}, \bibinfo{person}{Yunbo Cao},
  \bibinfo{person}{Qi Liu}, \bibinfo{person}{Tianyu Liu}, {and}
  \bibinfo{person}{Zhifang Sui}.} \bibinfo{year}{2023}\natexlab{e}.
\newblock \showarticletitle{Large language models are not fair evaluators}.
\newblock \bibinfo{journal}{\emph{arXiv preprint arXiv:2305.17926}}
  (\bibinfo{year}{2023}).
\newblock


\bibitem[Wang and Demszky(2023)]%
        {wang2023chatgpt}
\bibfield{author}{\bibinfo{person}{Rose~E Wang} {and} \bibinfo{person}{Dorottya
  Demszky}.} \bibinfo{year}{2023}\natexlab{}.
\newblock \showarticletitle{Is ChatGPT a Good Teacher Coach? Measuring
  Zero-Shot Performance For Scoring and Providing Actionable Insights on
  Classroom Instruction}.
\newblock \bibinfo{journal}{\emph{arXiv preprint arXiv:2306.03090}}
  (\bibinfo{year}{2023}).
\newblock


\bibitem[Wang et~al\mbox{.}(2023b)]%
        {wang2023cmb}
\bibfield{author}{\bibinfo{person}{Xidong Wang}, \bibinfo{person}{Guiming~Hardy
  Chen}, \bibinfo{person}{Dingjie Song}, \bibinfo{person}{Zhiyi Zhang},
  \bibinfo{person}{Zhihong Chen}, \bibinfo{person}{Qingying Xiao},
  \bibinfo{person}{Feng Jiang}, \bibinfo{person}{Jianquan Li},
  \bibinfo{person}{Xiang Wan}, \bibinfo{person}{Benyou Wang}, {et~al\mbox{.}}}
  \bibinfo{year}{2023}\natexlab{b}.
\newblock \showarticletitle{CMB: A Comprehensive Medical Benchmark in Chinese}.
\newblock \bibinfo{journal}{\emph{arXiv preprint arXiv:2308.08833}}
  (\bibinfo{year}{2023}).
\newblock


\bibitem[Wang et~al\mbox{.}(2023g)]%
        {wang2023emotional}
\bibfield{author}{\bibinfo{person}{Xuena Wang}, \bibinfo{person}{Xueting Li},
  \bibinfo{person}{Zi Yin}, \bibinfo{person}{Yue Wu}, {and}
  \bibinfo{person}{Liu Jia}.} \bibinfo{year}{2023}\natexlab{g}.
\newblock \bibinfo{title}{Emotional Intelligence of Large Language Models}.
\newblock
\newblock
\showeprint[arxiv]{2307.09042}~[cs.AI]


\bibitem[Wang et~al\mbox{.}(2023i)]%
        {wang2023mint}
\bibfield{author}{\bibinfo{person}{Xingyao Wang}, \bibinfo{person}{Zihan Wang},
  \bibinfo{person}{Jiateng Liu}, \bibinfo{person}{Yangyi Chen},
  \bibinfo{person}{Lifan Yuan}, \bibinfo{person}{Hao Peng}, {and}
  \bibinfo{person}{Heng Ji}.} \bibinfo{year}{2023}\natexlab{i}.
\newblock \showarticletitle{MINT: Evaluating LLMs in Multi-turn Interaction
  with Tools and Language Feedback}.
\newblock \bibinfo{journal}{\emph{arXiv preprint arXiv:2309.10691}}
  (\bibinfo{year}{2023}).
\newblock


\bibitem[Wang et~al\mbox{.}(2021a)]%
        {wang2021codet5}
\bibfield{author}{\bibinfo{person}{Yue Wang}, \bibinfo{person}{Weishi Wang},
  \bibinfo{person}{Shafiq Joty}, {and} \bibinfo{person}{Steven~CH Hoi}.}
  \bibinfo{year}{2021}\natexlab{a}.
\newblock \showarticletitle{Codet5: Identifier-aware unified pre-trained
  encoder-decoder models for code understanding and generation}.
\newblock \bibinfo{journal}{\emph{arXiv preprint arXiv:2109.00859}}
  (\bibinfo{year}{2021}).
\newblock


\bibitem[Wang et~al\mbox{.}(2023l)]%
        {wang2023exploring}
\bibfield{author}{\bibinfo{person}{Yidong Wang}, \bibinfo{person}{Zhuohao Yu},
  \bibinfo{person}{Jindong Wang}, \bibinfo{person}{Qiang Heng},
  \bibinfo{person}{Hao Chen}, \bibinfo{person}{Wei Ye}, \bibinfo{person}{Rui
  Xie}, \bibinfo{person}{Xing Xie}, {and} \bibinfo{person}{Shikun Zhang}.}
  \bibinfo{year}{2023}\natexlab{l}.
\newblock \showarticletitle{Exploring Vision-Language Models for Imbalanced
  Learning}.
\newblock \bibinfo{journal}{\emph{arXiv preprint arXiv:2304.01457}}
  (\bibinfo{year}{2023}).
\newblock


\bibitem[Wang et~al\mbox{.}(2023m)]%
        {wang2023pandalm}
\bibfield{author}{\bibinfo{person}{Yidong Wang}, \bibinfo{person}{Zhuohao Yu},
  \bibinfo{person}{Zhengran Zeng}, \bibinfo{person}{Linyi Yang},
  \bibinfo{person}{Cunxiang Wang}, \bibinfo{person}{Hao Chen},
  \bibinfo{person}{Chaoya Jiang}, \bibinfo{person}{Rui Xie},
  \bibinfo{person}{Jindong Wang}, \bibinfo{person}{Xing Xie}, {et~al\mbox{.}}}
  \bibinfo{year}{2023}\natexlab{m}.
\newblock \showarticletitle{PandaLM: An Automatic Evaluation Benchmark for LLM
  Instruction Tuning Optimization}.
\newblock \bibinfo{journal}{\emph{arXiv preprint arXiv:2306.05087}}
  (\bibinfo{year}{2023}).
\newblock


\bibitem[Wang et~al\mbox{.}(2023f)]%
        {wang2023can}
\bibfield{author}{\bibinfo{person}{Zhuo Wang}, \bibinfo{person}{Rongzhen Li},
  \bibinfo{person}{Bowen Dong}, \bibinfo{person}{Jie Wang},
  \bibinfo{person}{Xiuxing Li}, \bibinfo{person}{Ning Liu},
  \bibinfo{person}{Chenhui Mao}, \bibinfo{person}{Wei Zhang},
  \bibinfo{person}{Liling Dong}, \bibinfo{person}{Jing Gao}, {et~al\mbox{.}}}
  \bibinfo{year}{2023}\natexlab{f}.
\newblock \showarticletitle{Can LLMs like GPT-4 outperform traditional AI tools
  in dementia diagnosis? Maybe, but not today}.
\newblock \bibinfo{journal}{\emph{arXiv preprint arXiv:2306.01499}}
  (\bibinfo{year}{2023}).
\newblock


\bibitem[Wang et~al\mbox{.}(2023k)]%
        {wang2023chatgpt1}
\bibfield{author}{\bibinfo{person}{Zengzhi Wang}, \bibinfo{person}{Qiming Xie},
  \bibinfo{person}{Zixiang Ding}, \bibinfo{person}{Yi Feng}, {and}
  \bibinfo{person}{Rui Xia}.} \bibinfo{year}{2023}\natexlab{k}.
\newblock \bibinfo{title}{Is ChatGPT a Good Sentiment Analyzer? A Preliminary
  Study}.
\newblock
\newblock
\showeprint[arxiv]{2304.04339}~[cs.CL]


\bibitem[Wei et~al\mbox{.}(2022a)]%
        {wei2022emergent}
\bibfield{author}{\bibinfo{person}{Jason Wei}, \bibinfo{person}{Yi Tay},
  \bibinfo{person}{Rishi Bommasani}, \bibinfo{person}{Colin Raffel},
  \bibinfo{person}{Barret Zoph}, \bibinfo{person}{Sebastian Borgeaud},
  \bibinfo{person}{Dani Yogatama}, \bibinfo{person}{Maarten Bosma},
  \bibinfo{person}{Denny Zhou}, \bibinfo{person}{Donald Metzler},
  {et~al\mbox{.}}} \bibinfo{year}{2022}\natexlab{a}.
\newblock \showarticletitle{Emergent abilities of large language models}.
\newblock \bibinfo{journal}{\emph{arXiv preprint arXiv:2206.07682}}
  (\bibinfo{year}{2022}).
\newblock


\bibitem[Wei et~al\mbox{.}(2022b)]%
        {Wei2022EmergentAO}
\bibfield{author}{\bibinfo{person}{Jason Wei}, \bibinfo{person}{Yi Tay},
  \bibinfo{person}{Rishi Bommasani}, \bibinfo{person}{Colin Raffel},
  \bibinfo{person}{Barret Zoph}, \bibinfo{person}{Sebastian Borgeaud},
  \bibinfo{person}{Dani Yogatama}, \bibinfo{person}{Maarten Bosma},
  \bibinfo{person}{Denny Zhou}, \bibinfo{person}{Donald Metzler},
  \bibinfo{person}{Ed~Huai hsin Chi}, \bibinfo{person}{Tatsunori Hashimoto},
  \bibinfo{person}{Oriol Vinyals}, \bibinfo{person}{Percy Liang},
  \bibinfo{person}{Jeff Dean}, {and} \bibinfo{person}{William Fedus}.}
  \bibinfo{year}{2022}\natexlab{b}.
\newblock \showarticletitle{Emergent Abilities of Large Language Models}.
\newblock \bibinfo{journal}{\emph{Trans. Mach. Learn. Res.}}
  \bibinfo{volume}{2022} (\bibinfo{year}{2022}).
\newblock


\bibitem[Wei et~al\mbox{.}(2023)]%
        {wei2023cmath}
\bibfield{author}{\bibinfo{person}{Tianwen Wei}, \bibinfo{person}{Jian Luan},
  \bibinfo{person}{Wei Liu}, \bibinfo{person}{Shuang Dong}, {and}
  \bibinfo{person}{Bin Wang}.} \bibinfo{year}{2023}\natexlab{}.
\newblock \bibinfo{title}{CMATH: Can Your Language Model Pass Chinese
  Elementary School Math Test?}
\newblock
\newblock
\showeprint[arxiv]{2306.16636}~[cs.CL]


\bibitem[White et~al\mbox{.}(2023)]%
        {white2023prompt}
\bibfield{author}{\bibinfo{person}{Jules White}, \bibinfo{person}{Quchen Fu},
  \bibinfo{person}{Sam Hays}, \bibinfo{person}{Michael Sandborn},
  \bibinfo{person}{Carlos Olea}, \bibinfo{person}{Henry Gilbert},
  \bibinfo{person}{Ashraf Elnashar}, \bibinfo{person}{Jesse Spencer-Smith},
  {and} \bibinfo{person}{Douglas~C Schmidt}.} \bibinfo{year}{2023}\natexlab{}.
\newblock \showarticletitle{A prompt pattern catalog to enhance prompt
  engineering with chatgpt}.
\newblock \bibinfo{journal}{\emph{arXiv preprint arXiv:2302.11382}}
  (\bibinfo{year}{2023}).
\newblock


\bibitem[Wong(2015)]%
        {wong2015performance}
\bibfield{author}{\bibinfo{person}{Tzu-Tsung Wong}.}
  \bibinfo{year}{2015}\natexlab{}.
\newblock \showarticletitle{Performance evaluation of classification algorithms
  by k-fold and leave-one-out cross validation}.
\newblock \bibinfo{journal}{\emph{Pattern Recognition}} \bibinfo{volume}{48},
  \bibinfo{number}{9} (\bibinfo{year}{2015}), \bibinfo{pages}{2839--2846}.
\newblock


\bibitem[Wu et~al\mbox{.}(2023c)]%
        {wu2023large}
\bibfield{author}{\bibinfo{person}{Patrick~Y Wu}, \bibinfo{person}{Joshua~A
  Tucker}, \bibinfo{person}{Jonathan Nagler}, {and} \bibinfo{person}{Solomon
  Messing}.} \bibinfo{year}{2023}\natexlab{c}.
\newblock \showarticletitle{Large Language Models Can Be Used to Estimate the
  Ideologies of Politicians in a Zero-Shot Learning Setting}.
\newblock \bibinfo{journal}{\emph{arXiv preprint arXiv:2303.12057}}
  (\bibinfo{year}{2023}).
\newblock


\bibitem[Wu et~al\mbox{.}(2023a)]%
        {wu2023empirical}
\bibfield{author}{\bibinfo{person}{Yiran Wu}, \bibinfo{person}{Feiran Jia},
  \bibinfo{person}{Shaokun Zhang}, \bibinfo{person}{Qingyun Wu},
  \bibinfo{person}{Hangyu Li}, \bibinfo{person}{Erkang Zhu},
  \bibinfo{person}{Yue Wang}, \bibinfo{person}{Yin~Tat Lee},
  \bibinfo{person}{Richard Peng}, {and} \bibinfo{person}{Chi Wang}.}
  \bibinfo{year}{2023}\natexlab{a}.
\newblock \showarticletitle{An Empirical Study on Challenging Math Problem
  Solving with GPT-4}.
\newblock \bibinfo{journal}{\emph{arXiv preprint arXiv:2306.01337}}
  (\bibinfo{year}{2023}).
\newblock


\bibitem[Wu et~al\mbox{.}(2022)]%
        {wu2022autoformalization}
\bibfield{author}{\bibinfo{person}{Yuhuai Wu}, \bibinfo{person}{Albert~Qiaochu
  Jiang}, \bibinfo{person}{Wenda Li}, \bibinfo{person}{Markus Rabe},
  \bibinfo{person}{Charles Staats}, \bibinfo{person}{Mateja Jamnik}, {and}
  \bibinfo{person}{Christian Szegedy}.} \bibinfo{year}{2022}\natexlab{}.
\newblock \showarticletitle{Autoformalization with large language models}.
\newblock \bibinfo{journal}{\emph{Advances in Neural Information Processing
  Systems}}  \bibinfo{volume}{35} (\bibinfo{year}{2022}),
  \bibinfo{pages}{32353--32368}.
\newblock


\bibitem[Wu et~al\mbox{.}(2023b)]%
        {wu2023reasoning}
\bibfield{author}{\bibinfo{person}{Zhaofeng Wu}, \bibinfo{person}{Linlu Qiu},
  \bibinfo{person}{Alexis Ross}, \bibinfo{person}{Ekin Aky{\"u}rek},
  \bibinfo{person}{Boyuan Chen}, \bibinfo{person}{Bailin Wang},
  \bibinfo{person}{Najoung Kim}, \bibinfo{person}{Jacob Andreas}, {and}
  \bibinfo{person}{Yoon Kim}.} \bibinfo{year}{2023}\natexlab{b}.
\newblock \showarticletitle{Reasoning or Reciting? Exploring the Capabilities
  and Limitations of Language Models Through Counterfactual Tasks}.
\newblock \bibinfo{journal}{\emph{arXiv preprint arXiv:2307.02477}}
  (\bibinfo{year}{2023}).
\newblock


\bibitem[Xie et~al\mbox{.}(2023)]%
        {xie2023ask}
\bibfield{author}{\bibinfo{person}{Qiming Xie}, \bibinfo{person}{Zengzhi Wang},
  \bibinfo{person}{Yi Feng}, {and} \bibinfo{person}{Rui Xia}.}
  \bibinfo{year}{2023}\natexlab{}.
\newblock \bibinfo{title}{Ask Again, Then Fail: Large Language Models'
  Vacillations in Judgement}.
\newblock
\newblock
\showeprint[arxiv]{2310.02174}~[cs.CL]


\bibitem[Xu et~al\mbox{.}(2023b)]%
        {xu2023large}
\bibfield{author}{\bibinfo{person}{Fangzhi Xu}, \bibinfo{person}{Qika Lin},
  \bibinfo{person}{Jiawei Han}, \bibinfo{person}{Tianzhe Zhao},
  \bibinfo{person}{Jun Liu}, {and} \bibinfo{person}{Erik Cambria}.}
  \bibinfo{year}{2023}\natexlab{b}.
\newblock \showarticletitle{Are Large Language Models Really Good Logical
  Reasoners? A Comprehensive Evaluation From Deductive, Inductive and Abductive
  Views}.
\newblock \bibinfo{journal}{\emph{arXiv preprint arXiv:2306.09841}}
  (\bibinfo{year}{2023}).
\newblock


\bibitem[Xu et~al\mbox{.}(2023c)]%
        {xu2023cvalues}
\bibfield{author}{\bibinfo{person}{Guohai Xu}, \bibinfo{person}{Jiayi Liu},
  \bibinfo{person}{Ming Yan}, \bibinfo{person}{Haotian Xu},
  \bibinfo{person}{Jinghui Si}, \bibinfo{person}{Zhuoran Zhou},
  \bibinfo{person}{Peng Yi}, \bibinfo{person}{Xing Gao}, \bibinfo{person}{Jitao
  Sang}, \bibinfo{person}{Rong Zhang}, \bibinfo{person}{Ji Zhang},
  \bibinfo{person}{Chao Peng}, \bibinfo{person}{Fei Huang}, {and}
  \bibinfo{person}{Jingren Zhou}.} \bibinfo{year}{2023}\natexlab{c}.
\newblock \bibinfo{title}{CValues: Measuring the Values of Chinese Large
  Language Models from Safety to Responsibility}.
\newblock
\newblock
\showeprint[arxiv]{2307.09705}~[cs.CL]


\bibitem[Xu et~al\mbox{.}(2023d)]%
        {xu2023lvlmehub}
\bibfield{author}{\bibinfo{person}{Peng Xu}, \bibinfo{person}{Wenqi Shao},
  \bibinfo{person}{Kaipeng Zhang}, \bibinfo{person}{Peng Gao},
  \bibinfo{person}{Shuo Liu}, \bibinfo{person}{Meng Lei},
  \bibinfo{person}{Fanqing Meng}, \bibinfo{person}{Siyuan Huang},
  \bibinfo{person}{Yu Qiao}, {and} \bibinfo{person}{Ping Luo}.}
  \bibinfo{year}{2023}\natexlab{d}.
\newblock \bibinfo{title}{LVLM-eHub: A Comprehensive Evaluation Benchmark for
  Large Vision-Language Models}.
\newblock
\newblock
\showeprint[arxiv]{2306.09265}~[cs.CV]


\bibitem[Xu et~al\mbox{.}(2023a)]%
        {xu2023chatgpt}
\bibfield{author}{\bibinfo{person}{Ruiyun Xu}, \bibinfo{person}{Yue Feng},
  {and} \bibinfo{person}{Hailiang Chen}.} \bibinfo{year}{2023}\natexlab{a}.
\newblock \showarticletitle{ChatGPT vs. Google: A Comparative Study of Search
  Performance and User Experience}.
\newblock \bibinfo{journal}{\emph{arXiv preprint arXiv:2307.01135}}
  (\bibinfo{year}{2023}).
\newblock


\bibitem[Yang and Menczer(2023)]%
        {yang2023large}
\bibfield{author}{\bibinfo{person}{Kai-Cheng Yang} {and}
  \bibinfo{person}{Filippo Menczer}.} \bibinfo{year}{2023}\natexlab{}.
\newblock \showarticletitle{Large language models can rate news outlet
  credibility}.
\newblock \bibinfo{journal}{\emph{arXiv preprint arXiv:2304.00228}}
  (\bibinfo{year}{2023}).
\newblock


\bibitem[Yang et~al\mbox{.}(2022)]%
        {yang2022glue}
\bibfield{author}{\bibinfo{person}{Linyi Yang}, \bibinfo{person}{Shuibai
  Zhang}, \bibinfo{person}{Libo Qin}, \bibinfo{person}{Yafu Li},
  \bibinfo{person}{Yidong Wang}, \bibinfo{person}{Hanmeng Liu},
  \bibinfo{person}{Jindong Wang}, \bibinfo{person}{Xing Xie}, {and}
  \bibinfo{person}{Yue Zhang}.} \bibinfo{year}{2022}\natexlab{}.
\newblock \showarticletitle{Glue-x: Evaluating natural language understanding
  models from an out-of-distribution generalization perspective}.
\newblock \bibinfo{journal}{\emph{arXiv preprint arXiv:2211.08073}}
  (\bibinfo{year}{2022}).
\newblock


\bibitem[Yin et~al\mbox{.}(2023)]%
        {yin2023lamm}
\bibfield{author}{\bibinfo{person}{Zhenfei Yin}, \bibinfo{person}{Jiong Wang},
  \bibinfo{person}{Jianjian Cao}, \bibinfo{person}{Zhelun Shi},
  \bibinfo{person}{Dingning Liu}, \bibinfo{person}{Mukai Li},
  \bibinfo{person}{Lu Sheng}, \bibinfo{person}{Lei Bai},
  \bibinfo{person}{Xiaoshui Huang}, \bibinfo{person}{Zhiyong Wang},
  {et~al\mbox{.}}} \bibinfo{year}{2023}\natexlab{}.
\newblock \showarticletitle{LAMM: Language-Assisted Multi-Modal
  Instruction-Tuning Dataset, Framework, and Benchmark}.
\newblock \bibinfo{journal}{\emph{arXiv preprint arXiv:2306.06687}}
  (\bibinfo{year}{2023}).
\newblock


\bibitem[Yu et~al\mbox{.}(2023b)]%
        {yu2023kola}
\bibfield{author}{\bibinfo{person}{Jifan Yu}, \bibinfo{person}{Xiaozhi Wang},
  \bibinfo{person}{Shangqing Tu}, \bibinfo{person}{Shulin Cao},
  \bibinfo{person}{Daniel Zhang-Li}, \bibinfo{person}{Xin Lv},
  \bibinfo{person}{Hao Peng}, \bibinfo{person}{Zijun Yao},
  \bibinfo{person}{Xiaohan Zhang}, \bibinfo{person}{Hanming Li},
  {et~al\mbox{.}}} \bibinfo{year}{2023}\natexlab{b}.
\newblock \showarticletitle{KoLA: Carefully Benchmarking World Knowledge of
  Large Language Models}.
\newblock \bibinfo{journal}{\emph{arXiv preprint arXiv:2306.09296}}
  (\bibinfo{year}{2023}).
\newblock


\bibitem[Yu et~al\mbox{.}(2023a)]%
        {yu2023metamath}
\bibfield{author}{\bibinfo{person}{Longhui Yu}, \bibinfo{person}{Weisen Jiang},
  \bibinfo{person}{Han Shi}, \bibinfo{person}{Jincheng Yu},
  \bibinfo{person}{Zhengying Liu}, \bibinfo{person}{Yu Zhang},
  \bibinfo{person}{James~T Kwok}, \bibinfo{person}{Zhenguo Li},
  \bibinfo{person}{Adrian Weller}, {and} \bibinfo{person}{Weiyang Liu}.}
  \bibinfo{year}{2023}\natexlab{a}.
\newblock \showarticletitle{MetaMath: Bootstrap Your Own Mathematical Questions
  for Large Language Models}.
\newblock \bibinfo{journal}{\emph{arXiv preprint arXiv:2309.12284}}
  (\bibinfo{year}{2023}).
\newblock


\bibitem[Yu et~al\mbox{.}(2023c)]%
        {yu2023mm}
\bibfield{author}{\bibinfo{person}{Weihao Yu}, \bibinfo{person}{Zhengyuan
  Yang}, \bibinfo{person}{Linjie Li}, \bibinfo{person}{Jianfeng Wang},
  \bibinfo{person}{Kevin Lin}, \bibinfo{person}{Zicheng Liu},
  \bibinfo{person}{Xinchao Wang}, {and} \bibinfo{person}{Lijuan Wang}.}
  \bibinfo{year}{2023}\natexlab{c}.
\newblock \showarticletitle{Mm-vet: Evaluating large multimodal models for
  integrated capabilities}.
\newblock \bibinfo{journal}{\emph{arXiv preprint arXiv:2308.02490}}
  (\bibinfo{year}{2023}).
\newblock


\bibitem[Yuan et~al\mbox{.}(2023a)]%
        {yuan2023revisiting}
\bibfield{author}{\bibinfo{person}{Lifan Yuan}, \bibinfo{person}{Yangyi Chen},
  \bibinfo{person}{Ganqu Cui}, \bibinfo{person}{Hongcheng Gao},
  \bibinfo{person}{Fangyuan Zou}, \bibinfo{person}{Xingyi Cheng},
  \bibinfo{person}{Heng Ji}, \bibinfo{person}{Zhiyuan Liu}, {and}
  \bibinfo{person}{Maosong Sun}.} \bibinfo{year}{2023}\natexlab{a}.
\newblock \bibinfo{title}{Revisiting Out-of-distribution Robustness in NLP:
  Benchmark, Analysis, and LLMs Evaluations}.
\newblock
\newblock
\showeprint[arxiv]{2306.04618}~[cs.CL]


\bibitem[Yuan et~al\mbox{.}(2023b)]%
        {yuan2023recommender}
\bibfield{author}{\bibinfo{person}{Zheng Yuan}, \bibinfo{person}{Fajie Yuan},
  \bibinfo{person}{Yu Song}, \bibinfo{person}{Youhua Li},
  \bibinfo{person}{Junchen Fu}, \bibinfo{person}{Fei Yang},
  \bibinfo{person}{Yunzhu Pan}, {and} \bibinfo{person}{Yongxin Ni}.}
  \bibinfo{year}{2023}\natexlab{b}.
\newblock \bibinfo{title}{Where to Go Next for Recommender Systems? ID- vs.
  Modality-based Recommender Models Revisited}.
\newblock
\newblock
\showeprint[arxiv]{2303.13835}~[cs.IR]


\bibitem[Yuan et~al\mbox{.}(2023c)]%
        {yuan2023well}
\bibfield{author}{\bibinfo{person}{Zheng Yuan}, \bibinfo{person}{Hongyi Yuan},
  \bibinfo{person}{Chuanqi Tan}, \bibinfo{person}{Wei Wang}, {and}
  \bibinfo{person}{Songfang Huang}.} \bibinfo{year}{2023}\natexlab{c}.
\newblock \showarticletitle{How well do Large Language Models perform in
  Arithmetic tasks?}
\newblock \bibinfo{journal}{\emph{arXiv preprint arXiv:2304.02015}}
  (\bibinfo{year}{2023}).
\newblock


\bibitem[Zemel et~al\mbox{.}(2013)]%
        {zemel2013learning}
\bibfield{author}{\bibinfo{person}{Rich Zemel}, \bibinfo{person}{Yu Wu},
  \bibinfo{person}{Kevin Swersky}, \bibinfo{person}{Toni Pitassi}, {and}
  \bibinfo{person}{Cynthia Dwork}.} \bibinfo{year}{2013}\natexlab{}.
\newblock \showarticletitle{Learning fair representations}. In
  \bibinfo{booktitle}{\emph{International conference on machine learning}}.
  PMLR, \bibinfo{pages}{325--333}.
\newblock


\bibitem[Zeng et~al\mbox{.}(2022)]%
        {zeng2022glm}
\bibfield{author}{\bibinfo{person}{Aohan Zeng}, \bibinfo{person}{Xiao Liu},
  \bibinfo{person}{Zhengxiao Du}, \bibinfo{person}{Zihan Wang},
  \bibinfo{person}{Hanyu Lai}, \bibinfo{person}{Ming Ding},
  \bibinfo{person}{Zhuoyi Yang}, \bibinfo{person}{Yifan Xu},
  \bibinfo{person}{Wendi Zheng}, \bibinfo{person}{Xiao Xia}, {et~al\mbox{.}}}
  \bibinfo{year}{2022}\natexlab{}.
\newblock \showarticletitle{Glm-130b: An open bilingual pre-trained model}.
\newblock \bibinfo{journal}{\emph{arXiv preprint arXiv:2210.02414}}
  (\bibinfo{year}{2022}).
\newblock


\bibitem[Zhang et~al\mbox{.}(2023h)]%
        {zhang2023evaluating1}
\bibfield{author}{\bibinfo{person}{Beichen Zhang}, \bibinfo{person}{Kun Zhou},
  \bibinfo{person}{Xilin Wei}, \bibinfo{person}{Wayne~Xin Zhao},
  \bibinfo{person}{Jing Sha}, \bibinfo{person}{Shijin Wang}, {and}
  \bibinfo{person}{Ji-Rong Wen}.} \bibinfo{year}{2023}\natexlab{h}.
\newblock \showarticletitle{Evaluating and Improving Tool-Augmented
  Computation-Intensive Math Reasoning}.
\newblock \bibinfo{journal}{\emph{arXiv preprint arXiv:2306.02408}}
  (\bibinfo{year}{2023}).
\newblock


\bibitem[Zhang et~al\mbox{.}(2023i)]%
        {zhang2023evaluating}
\bibfield{author}{\bibinfo{person}{Beichen Zhang}, \bibinfo{person}{Kun Zhou},
  \bibinfo{person}{Xilin Wei}, \bibinfo{person}{Wayne~Xin Zhao},
  \bibinfo{person}{Jing Sha}, \bibinfo{person}{Shijin Wang}, {and}
  \bibinfo{person}{Ji-Rong Wen}.} \bibinfo{year}{2023}\natexlab{i}.
\newblock \showarticletitle{Evaluating and Improving Tool-Augmented
  Computation-Intensive Math Reasoning}.
\newblock \bibinfo{journal}{\emph{arXiv preprint arXiv:2306.02408}}
  (\bibinfo{year}{2023}).
\newblock


\bibitem[Zhang et~al\mbox{.}(2023b)]%
        {zhang2023chatgpt}
\bibfield{author}{\bibinfo{person}{Jizhi Zhang}, \bibinfo{person}{Keqin Bao},
  \bibinfo{person}{Yang Zhang}, \bibinfo{person}{Wenjie Wang},
  \bibinfo{person}{Fuli Feng}, {and} \bibinfo{person}{Xiangnan He}.}
  \bibinfo{year}{2023}\natexlab{b}.
\newblock \showarticletitle{Is ChatGPT Fair for Recommendation? Evaluating
  Fairness in Large Language Model Recommendation}.
\newblock \bibinfo{journal}{\emph{arXiv preprint arXiv:2305.07609}}
  (\bibinfo{year}{2023}).
\newblock


\bibitem[Zhang et~al\mbox{.}(2022)]%
        {zhang2022opt}
\bibfield{author}{\bibinfo{person}{Susan Zhang}, \bibinfo{person}{Stephen
  Roller}, \bibinfo{person}{Naman Goyal}, \bibinfo{person}{Mikel Artetxe},
  \bibinfo{person}{Moya Chen}, \bibinfo{person}{Shuohui Chen},
  \bibinfo{person}{Christopher Dewan}, \bibinfo{person}{Mona Diab},
  \bibinfo{person}{Xian Li}, \bibinfo{person}{Xi~Victoria Lin},
  {et~al\mbox{.}}} \bibinfo{year}{2022}\natexlab{}.
\newblock \showarticletitle{Opt: Open pre-trained transformer language models}.
\newblock \bibinfo{journal}{\emph{arXiv preprint arXiv:2205.01068}}
  (\bibinfo{year}{2022}).
\newblock


\bibitem[Zhang et~al\mbox{.}(2023d)]%
        {zhang2023exploring}
\bibfield{author}{\bibinfo{person}{Sarah~J Zhang}, \bibinfo{person}{Samuel
  Florin}, \bibinfo{person}{Ariel~N Lee}, \bibinfo{person}{Eamon Niknafs},
  \bibinfo{person}{Andrei Marginean}, \bibinfo{person}{Annie Wang},
  \bibinfo{person}{Keith Tyser}, \bibinfo{person}{Zad Chin},
  \bibinfo{person}{Yann Hicke}, \bibinfo{person}{Nikhil Singh},
  {et~al\mbox{.}}} \bibinfo{year}{2023}\natexlab{d}.
\newblock \showarticletitle{Exploring the MIT Mathematics and EECS Curriculum
  Using Large Language Models}.
\newblock \bibinfo{journal}{\emph{arXiv preprint arXiv:2306.08997}}
  (\bibinfo{year}{2023}).
\newblock


\bibitem[Zhang et~al\mbox{.}(2019)]%
        {zhang2019bertscore}
\bibfield{author}{\bibinfo{person}{Tianyi Zhang}, \bibinfo{person}{Varsha
  Kishore}, \bibinfo{person}{Felix Wu}, \bibinfo{person}{Kilian~Q Weinberger},
  {and} \bibinfo{person}{Yoav Artzi}.} \bibinfo{year}{2019}\natexlab{}.
\newblock \showarticletitle{Bertscore: Evaluating text generation with bert}.
\newblock \bibinfo{journal}{\emph{arXiv preprint arXiv:1904.09675}}
  (\bibinfo{year}{2019}).
\newblock


\bibitem[Zhang et~al\mbox{.}(2023a)]%
        {zhang2023m3exam}
\bibfield{author}{\bibinfo{person}{Wenxuan Zhang},
  \bibinfo{person}{Sharifah~Mahani Aljunied}, \bibinfo{person}{Chang Gao},
  \bibinfo{person}{Yew~Ken Chia}, {and} \bibinfo{person}{Lidong Bing}.}
  \bibinfo{year}{2023}\natexlab{a}.
\newblock \showarticletitle{M3Exam: A Multilingual, Multimodal, Multilevel
  Benchmark for Examining Large Language Models}.
\newblock \bibinfo{journal}{\emph{arXiv preprint arXiv:2306.05179}}
  (\bibinfo{year}{2023}).
\newblock


\bibitem[Zhang et~al\mbox{.}(2023c)]%
        {zhang2023sentiment}
\bibfield{author}{\bibinfo{person}{Wenxuan Zhang}, \bibinfo{person}{Yue Deng},
  \bibinfo{person}{Bing Liu}, \bibinfo{person}{Sinno~Jialin Pan}, {and}
  \bibinfo{person}{Lidong Bing}.} \bibinfo{year}{2023}\natexlab{c}.
\newblock \showarticletitle{Sentiment Analysis in the Era of Large Language
  Models: A Reality Check}.
\newblock \bibinfo{journal}{\emph{arXiv preprint arXiv:2305.15005}}
  (\bibinfo{year}{2023}).
\newblock


\bibitem[Zhang et~al\mbox{.}(2023g)]%
        {zhang2023wider}
\bibfield{author}{\bibinfo{person}{Xinghua Zhang}, \bibinfo{person}{Bowen Yu},
  \bibinfo{person}{Haiyang Yu}, \bibinfo{person}{Yangyu Lv},
  \bibinfo{person}{Tingwen Liu}, \bibinfo{person}{Fei Huang},
  \bibinfo{person}{Hongbo Xu}, {and} \bibinfo{person}{Yongbin Li}.}
  \bibinfo{year}{2023}\natexlab{g}.
\newblock \showarticletitle{Wider and deeper llm networks are fairer llm
  evaluators}.
\newblock \bibinfo{journal}{\emph{arXiv preprint arXiv:2308.01862}}
  (\bibinfo{year}{2023}).
\newblock


\bibitem[Zhang et~al\mbox{.}(2023f)]%
        {zhang2023sirens}
\bibfield{author}{\bibinfo{person}{Yue Zhang}, \bibinfo{person}{Yafu Li},
  \bibinfo{person}{Leyang Cui}, \bibinfo{person}{Deng Cai},
  \bibinfo{person}{Lemao Liu}, \bibinfo{person}{Tingchen Fu},
  \bibinfo{person}{Xinting Huang}, \bibinfo{person}{Enbo Zhao},
  \bibinfo{person}{Yu Zhang}, \bibinfo{person}{Yulong Chen},
  \bibinfo{person}{Longyue Wang}, \bibinfo{person}{Anh~Tuan Luu},
  \bibinfo{person}{Wei Bi}, \bibinfo{person}{Freda Shi}, {and}
  \bibinfo{person}{Shuming Shi}.} \bibinfo{year}{2023}\natexlab{f}.
\newblock \bibinfo{title}{Siren's Song in the AI Ocean: A Survey on
  Hallucination in Large Language Models}.
\newblock
\newblock
\showeprint[arxiv]{2309.01219}~[cs.CL]


\bibitem[Zhang et~al\mbox{.}(2023e)]%
        {zhang2023safetybench}
\bibfield{author}{\bibinfo{person}{Zhexin Zhang}, \bibinfo{person}{Leqi Lei},
  \bibinfo{person}{Lindong Wu}, \bibinfo{person}{Rui Sun},
  \bibinfo{person}{Yongkang Huang}, \bibinfo{person}{Chong Long},
  \bibinfo{person}{Xiao Liu}, \bibinfo{person}{Xuanyu Lei},
  \bibinfo{person}{Jie Tang}, {and} \bibinfo{person}{Minlie Huang}.}
  \bibinfo{year}{2023}\natexlab{e}.
\newblock \showarticletitle{SafetyBench: Evaluating the Safety of Large
  Language Models with Multiple Choice Questions}.
\newblock \bibinfo{journal}{\emph{arXiv preprint arXiv:2309.07045}}
  (\bibinfo{year}{2023}).
\newblock


\bibitem[Zhao et~al\mbox{.}(2023a)]%
        {zhao2023mmicl}
\bibfield{author}{\bibinfo{person}{Haozhe Zhao}, \bibinfo{person}{Zefan Cai},
  \bibinfo{person}{Shuzheng Si}, \bibinfo{person}{Xiaojian Ma},
  \bibinfo{person}{Kaikai An}, \bibinfo{person}{Liang Chen},
  \bibinfo{person}{Zixuan Liu}, \bibinfo{person}{Sheng Wang},
  \bibinfo{person}{Wenjuan Han}, {and} \bibinfo{person}{Baobao Chang}.}
  \bibinfo{year}{2023}\natexlab{a}.
\newblock \showarticletitle{MMICL: Empowering Vision-language Model with
  Multi-Modal In-Context Learning}.
\newblock \bibinfo{journal}{\emph{arXiv preprint arXiv:2309.07915}}
  (\bibinfo{year}{2023}).
\newblock


\bibitem[Zhao et~al\mbox{.}(2023b)]%
        {zhao2023chbias}
\bibfield{author}{\bibinfo{person}{Jiaxu Zhao}, \bibinfo{person}{Meng Fang},
  \bibinfo{person}{Zijing Shi}, \bibinfo{person}{Yitong Li},
  \bibinfo{person}{Ling Chen}, {and} \bibinfo{person}{Mykola Pechenizkiy}.}
  \bibinfo{year}{2023}\natexlab{b}.
\newblock \bibinfo{title}{CHBias: Bias Evaluation and Mitigation of Chinese
  Conversational Language Models}.
\newblock
\newblock
\showeprint[arxiv]{2305.11262}~[cs.CL]


\bibitem[Zhao et~al\mbox{.}(2023d)]%
        {zhao2023survey}
\bibfield{author}{\bibinfo{person}{Wayne~Xin Zhao}, \bibinfo{person}{Kun Zhou},
  \bibinfo{person}{Junyi Li}, \bibinfo{person}{Tianyi Tang},
  \bibinfo{person}{Xiaolei Wang}, \bibinfo{person}{Yupeng Hou},
  \bibinfo{person}{Yingqian Min}, \bibinfo{person}{Beichen Zhang},
  \bibinfo{person}{Junjie Zhang}, \bibinfo{person}{Zican Dong},
  {et~al\mbox{.}}} \bibinfo{year}{2023}\natexlab{d}.
\newblock \showarticletitle{A survey of large language models}.
\newblock \bibinfo{journal}{\emph{arXiv preprint arXiv:2303.18223}}
  (\bibinfo{year}{2023}).
\newblock


\bibitem[Zhao et~al\mbox{.}(2023c)]%
        {zhao2023evaluating}
\bibfield{author}{\bibinfo{person}{Yunqing Zhao}, \bibinfo{person}{Tianyu
  Pang}, \bibinfo{person}{Chao Du}, \bibinfo{person}{Xiao Yang},
  \bibinfo{person}{Chongxuan Li}, \bibinfo{person}{Ngai-Man Cheung}, {and}
  \bibinfo{person}{Min Lin}.} \bibinfo{year}{2023}\natexlab{c}.
\newblock \showarticletitle{On Evaluating Adversarial Robustness of Large
  Vision-Language Models}.
\newblock \bibinfo{journal}{\emph{arXiv preprint arXiv:2305.16934}}
  (\bibinfo{year}{2023}).
\newblock


\bibitem[Zheng et~al\mbox{.}(2023a)]%
        {zheng2023lmsys}
\bibfield{author}{\bibinfo{person}{Lianmin Zheng}, \bibinfo{person}{Wei-Lin
  Chiang}, \bibinfo{person}{Ying Sheng}, \bibinfo{person}{Tianle Li},
  \bibinfo{person}{Siyuan Zhuang}, \bibinfo{person}{Zhanghao Wu},
  \bibinfo{person}{Yonghao Zhuang}, \bibinfo{person}{Zhuohan Li},
  \bibinfo{person}{Zi Lin}, \bibinfo{person}{Eric Xing}, {et~al\mbox{.}}}
  \bibinfo{year}{2023}\natexlab{a}.
\newblock \showarticletitle{LMSYS-Chat-1M: A Large-Scale Real-World LLM
  Conversation Dataset}.
\newblock \bibinfo{journal}{\emph{arXiv preprint arXiv:2309.11998}}
  (\bibinfo{year}{2023}).
\newblock


\bibitem[Zheng et~al\mbox{.}(2023b)]%
        {zheng2023judging}
\bibfield{author}{\bibinfo{person}{Lianmin Zheng}, \bibinfo{person}{Wei-Lin
  Chiang}, \bibinfo{person}{Ying Sheng}, \bibinfo{person}{Siyuan Zhuang},
  \bibinfo{person}{Zhanghao Wu}, \bibinfo{person}{Yonghao Zhuang},
  \bibinfo{person}{Zi Lin}, \bibinfo{person}{Zhuohan Li},
  \bibinfo{person}{Dacheng Li}, \bibinfo{person}{Eric.~P Xing},
  \bibinfo{person}{Hao Zhang}, \bibinfo{person}{Joseph~E. Gonzalez}, {and}
  \bibinfo{person}{Ion Stoica}.} \bibinfo{year}{2023}\natexlab{b}.
\newblock \bibinfo{title}{Judging LLM-as-a-judge with MT-Bench and Chatbot
  Arena}.
\newblock
\newblock
\showeprint[arxiv]{2306.05685}~[cs.CL]


\bibitem[Zhong et~al\mbox{.}(2022)]%
        {zhong2022towards}
\bibfield{author}{\bibinfo{person}{Ming Zhong}, \bibinfo{person}{Yang Liu},
  \bibinfo{person}{Da Yin}, \bibinfo{person}{Yuning Mao},
  \bibinfo{person}{Yizhu Jiao}, \bibinfo{person}{Pengfei Liu},
  \bibinfo{person}{Chenguang Zhu}, \bibinfo{person}{Heng Ji}, {and}
  \bibinfo{person}{Jiawei Han}.} \bibinfo{year}{2022}\natexlab{}.
\newblock \showarticletitle{Towards a unified multi-dimensional evaluator for
  text generation}.
\newblock \bibinfo{journal}{\emph{arXiv preprint arXiv:2210.07197}}
  (\bibinfo{year}{2022}).
\newblock


\bibitem[Zhong et~al\mbox{.}(2023)]%
        {zhong2023agieval}
\bibfield{author}{\bibinfo{person}{Wanjun Zhong}, \bibinfo{person}{Ruixiang
  Cui}, \bibinfo{person}{Yiduo Guo}, \bibinfo{person}{Yaobo Liang},
  \bibinfo{person}{Shuai Lu}, \bibinfo{person}{Yanlin Wang},
  \bibinfo{person}{Amin Saied}, \bibinfo{person}{Weizhu Chen}, {and}
  \bibinfo{person}{Nan Duan}.} \bibinfo{year}{2023}\natexlab{}.
\newblock \showarticletitle{Agieval: A human-centric benchmark for evaluating
  foundation models}.
\newblock \bibinfo{journal}{\emph{arXiv preprint arXiv:2304.06364}}
  (\bibinfo{year}{2023}).
\newblock


\bibitem[Zhou et~al\mbox{.}(2022)]%
        {zhou2022large}
\bibfield{author}{\bibinfo{person}{Yongchao Zhou}, \bibinfo{person}{Andrei~Ioan
  Muresanu}, \bibinfo{person}{Ziwen Han}, \bibinfo{person}{Keiran Paster},
  \bibinfo{person}{Silviu Pitis}, \bibinfo{person}{Harris Chan}, {and}
  \bibinfo{person}{Jimmy Ba}.} \bibinfo{year}{2022}\natexlab{}.
\newblock \showarticletitle{Large language models are human-level prompt
  engineers}.
\newblock \bibinfo{journal}{\emph{arXiv preprint arXiv:2211.01910}}
  (\bibinfo{year}{2022}).
\newblock


\bibitem[Zhu et~al\mbox{.}(2023)]%
        {zhu2023promptbench}
\bibfield{author}{\bibinfo{person}{Kaijie Zhu}, \bibinfo{person}{Jindong Wang},
  \bibinfo{person}{Jiaheng Zhou}, \bibinfo{person}{Zichen Wang},
  \bibinfo{person}{Hao Chen}, \bibinfo{person}{Yidong Wang},
  \bibinfo{person}{Linyi Yang}, \bibinfo{person}{Wei Ye},
  \bibinfo{person}{Neil~Zhenqiang Gong}, \bibinfo{person}{Yue Zhang},
  {et~al\mbox{.}}} \bibinfo{year}{2023}\natexlab{}.
\newblock \showarticletitle{PromptBench: Towards Evaluating the Robustness of
  Large Language Models on Adversarial Prompts}.
\newblock \bibinfo{journal}{\emph{arXiv preprint arXiv:2306.04528}}
  (\bibinfo{year}{2023}).
\newblock


\bibitem[Zhuang et~al\mbox{.}(2023)]%
        {zhuang2023efficiently}
\bibfield{author}{\bibinfo{person}{Yan Zhuang}, \bibinfo{person}{Qi Liu},
  \bibinfo{person}{Yuting Ning}, \bibinfo{person}{Weizhe Huang},
  \bibinfo{person}{Rui Lv}, \bibinfo{person}{Zhenya Huang},
  \bibinfo{person}{Guanhao Zhao}, \bibinfo{person}{Zheng Zhang},
  \bibinfo{person}{Qingyang Mao}, \bibinfo{person}{Shijin Wang},
  {et~al\mbox{.}}} \bibinfo{year}{2023}\natexlab{}.
\newblock \showarticletitle{Efficiently Measuring the Cognitive Ability of
  LLMs: An Adaptive Testing Perspective}.
\newblock \bibinfo{journal}{\emph{arXiv preprint arXiv:2306.10512}}
  (\bibinfo{year}{2023}).
\newblock


\bibitem[Zhuo et~al\mbox{.}(2023a)]%
        {zhuo2023exploring}
\bibfield{author}{\bibinfo{person}{Terry~Yue Zhuo}, \bibinfo{person}{Yujin
  Huang}, \bibinfo{person}{Chunyang Chen}, {and} \bibinfo{person}{Zhenchang
  Xing}.} \bibinfo{year}{2023}\natexlab{a}.
\newblock \showarticletitle{Exploring ai ethics of chatgpt: A diagnostic
  analysis}.
\newblock \bibinfo{journal}{\emph{arXiv preprint arXiv:2301.12867}}
  (\bibinfo{year}{2023}).
\newblock


\bibitem[Zhuo et~al\mbox{.}(2023b)]%
        {zhuo2023robustness}
\bibfield{author}{\bibinfo{person}{Terry~Yue Zhuo}, \bibinfo{person}{Zhuang
  Li}, \bibinfo{person}{Yujin Huang}, \bibinfo{person}{Yuan-Fang Li},
  \bibinfo{person}{Weiqing Wang}, \bibinfo{person}{Gholamreza Haffari}, {and}
  \bibinfo{person}{Fatemeh Shiri}.} \bibinfo{year}{2023}\natexlab{b}.
\newblock \showarticletitle{On Robustness of Prompt-based Semantic Parsing with
  Large Pre-trained Language Model: An Empirical Study on Codex}.
\newblock \bibinfo{journal}{\emph{arXiv preprint arXiv:2301.12868}}
  (\bibinfo{year}{2023}).
\newblock


\bibitem[Ziegler et~al\mbox{.}(2019)]%
        {ziegler2019fine}
\bibfield{author}{\bibinfo{person}{Daniel~M Ziegler}, \bibinfo{person}{Nisan
  Stiennon}, \bibinfo{person}{Jeffrey Wu}, \bibinfo{person}{Tom~B Brown},
  \bibinfo{person}{Alec Radford}, \bibinfo{person}{Dario Amodei},
  \bibinfo{person}{Paul Christiano}, {and} \bibinfo{person}{Geoffrey Irving}.}
  \bibinfo{year}{2019}\natexlab{}.
\newblock \showarticletitle{Fine-tuning language models from human
  preferences}.
\newblock \bibinfo{journal}{\emph{arXiv preprint arXiv:1909.08593}}
  (\bibinfo{year}{2019}).
\newblock


\bibitem[Ziems et~al\mbox{.}(2023)]%
        {ziems2023can}
\bibfield{author}{\bibinfo{person}{Caleb Ziems}, \bibinfo{person}{William
  Held}, \bibinfo{person}{Omar Shaikh}, \bibinfo{person}{Jiaao Chen},
  \bibinfo{person}{Zhehao Zhang}, {and} \bibinfo{person}{Diyi Yang}.}
  \bibinfo{year}{2023}\natexlab{}.
\newblock \showarticletitle{Can Large Language Models Transform Computational
  Social Science?}
\newblock \bibinfo{journal}{\emph{arXiv preprint arXiv:2305.03514}}
  (\bibinfo{year}{2023}).
\newblock


\end{thebibliography}
